\documentclass[review,number,sort&compress]{article}                     
\usepackage{arxiv}
\usepackage{hyperref}       
\usepackage{url}            
\usepackage[doublespacing]{setspace}
\usepackage{framed,multirow}
\usepackage{graphicx,subfig}
\graphicspath{{./}{../}{./images/}}
\DeclareGraphicsExtensions{.pdf,.tif,.jpg}
\usepackage{amsmath,amssymb}
\usepackage{algorithmic,algorithm,latexsym}
\usepackage{array,multirow}
\usepackage{fixltx2e}
\usepackage{url}
\usepackage{dblfloatfix}
\newcommand{\argmin}{\mathop{\mathrm{argmin}}}
\newcommand*{\thead}[1]{\multicolumn{1}{c}{\bfseries #1}}
\newcommand*{\theadb}[1]{{\begin{center}\scriptsize\bfseries #1\end{center}}}
\newcommand*{\theadm}[2]{\begin{tabular}[c]{@{}l@{}}{\scriptsize\bfseries #1}\\ {\scriptsize\bfseries #2}\end{tabular}}
\newcommand*{\tgraph}[1]{\includegraphics[width=\linewidth,clip,keepaspectratio]{#1}}
\begin{document}
    \title{Parzen Window Approximation on Riemannian Manifold
    }
    \author{\href{https://orcid.org/0000-0002-6491-8967}{\includegraphics[scale=0.06]{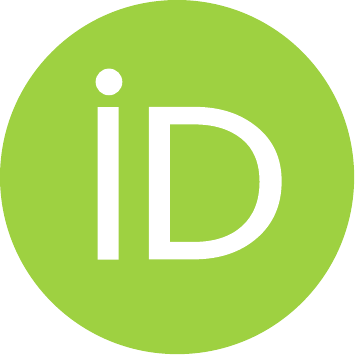}\hspace{1mm}Abhishek}~and~Shekhar~Verma\\
        Indian Institute of Information Technology, Allahabad, Uttar Pradesh-211015, India}
    \maketitle
    \begin{abstract}
        In graph motivated learning, label propagation largely depends on data affinity represented as edges between connected data points. The affinity assignment implicitly assumes even distribution of data on the manifold. This assumption may not hold and may lead to inaccurate metric assignment due to drift towards high-density regions. The drift affected heat kernel based affinity with a globally fixed Parzen window either discards genuine neighbors or forces distant data points to become a member of the neighborhood. This yields a biased affinity matrix. In this paper, the bias due to uneven data sampling on the Riemannian manifold is catered to by a variable Parzen window determined as a function of neighborhood size, ambient dimension, flatness range, etc. Additionally, affinity adjustment is used which offsets the effect of uneven sampling responsible for the bias. An affinity metric which takes into consideration the irregular sampling effect to yield accurate label propagation is proposed. Extensive experiments on synthetic and real-world data sets confirm that the proposed method increases the classification accuracy significantly and outperforms existing Parzen window estimators in graph Laplacian manifold regularization methods.
    \end{abstract}
    \keywords{Parzen window\and data affinity\and graph Laplacian regularization\and manifold regularization}
\section{Introduction}
\label{sec:introduction}
    Manifold learning \cite{manifold_learning02} and manifold regularization \cite{manifold_regularization} techniques work on the assumption that every $ (x_{i})_{i=1}^{n}\in X $\footnote{Refer Table \ref{table:symbols} for all mathematical symbols used.} data point on manifold is generated from $ x_{i}=f(\tau_{i})+\eta_{i} $ where, $ x_{i} $ actually lies in low dimensional space $ \tau_{i}\in\mathbb{R}^{d} $. Noise and redundancy $ \eta_{i} $ embeds it in a high dimensional smooth Riemann manifold $ \mathcal{M}\in\mathbb{R}^{D} $ where, $ D $ is the given ambient dimension, $ d $ is unknown intrinsic dimension and $ d\ll D $. The aim of manifold learning is to learn the manifold and discover the embedding in the intrinsic dimensions. Similarly, manifold regularization based semi-supervised learning (SSL) \cite{mit_i2ml,Bastanlar2014} exploits this intrinsic dimensional embedding as an intrinsic space regularization term under the Riemannian manifold assumption. Graph Laplacian \cite{laplacian01,laplacian02,iL2PA} approximates the Laplace-Beltrami operators of $ \mathcal{M} $ by measuring the divergence of the function gradient at every data point. An undirected weighted graph $ G=(X,W) $ is created over given data points $ X $ as its vertices and $ W $ is the affinity matrix containing a non-zero value at $ a_{ij}\in W $ if $ x_{i} $ and $ x_{j} $ are connected. The affinity is calculated by putting a heat kernel function over the distance metric which is high for spatially near data points and decays exponentially with distance. Assume $ u(x_{i}, \epsilon) $ be the heat distribution at distance $ \epsilon $ on the manifold $ \mathcal{M} $, initially it will be $ u(x_{i}, 0)=\varphi(x_{i}) $. At distance $ \epsilon>0 $, its value is given by $ u(x_{i}, \epsilon)=\int_{\mathcal{M}}\kappa(x_{i}, x_{j})\varphi(x_{j}) $. The heat kernel $ \kappa $ on the tangent plane is given by \cite{heatKernelequations}
    \[ \kappa(x_{i}, x_{j})=(4\pi \epsilon)^{-\frac{n}{2}}\exp\bigg(-\frac{\parallel x_{i}-x_{j}\parallel^{2}_{2}}{2\epsilon^{2}}\bigg)(\Phi(x_{i}, x_{j})+O(\epsilon)) \]
    $ \Phi $ is a smooth function with $ \Phi(x_{i},x_{i})=1 $ and $ \epsilon $ is the Parzen window or kernel bandwidth. When $ x_{i} $ and $ x_{j} $ are close on $ \mathcal{M} $ and $ \epsilon $ is small then
    \[ \kappa(x_{i}, x_{j})\approx (4\pi \epsilon)^{-\frac{n}{2}}\exp\bigg(-\frac{\parallel x_{i}-x_{j}\parallel^{2}_{2}}{2\epsilon^{2}}\bigg) \]
    $ \epsilon $ is the only hyper-parameter which drives the point-wise convergence of graph Laplacian \cite{ConvergenceOfLaplacianEigenmapsbelkin2007} to its respective Laplace-Beltrami operator. It is assumed that the given point cloud has been evenly sampled on $ \mathcal{M} $ i.e. the density around each data point remains similar and hence, a small $ \epsilon $ enforces smoothness over the function \cite{xiao2009graph,zhang2008graph}. However, this cannot be ensured in real-world data as the linear region around every data point $ x_{i} $ varies. Due to this, the neighborhoods do not maintain the same density.

    An ideal $ \epsilon $ can be obtained from \cite{convergenceSinger,joncas2017improved} $ \epsilon^{2}=B(\mathcal{M})n^{-\frac{1}{3+d/2}} $, where, $ B(\mathcal{M}) $ is a function defined on the geometrical properties (dimensions, curvature and volume) of the underlying manifold. $ \epsilon $ remains proportional to the injectivity radius of $ \mathcal{M} $ i.e. the maximum distance to which manifold is linear when density remains constant. However, as both $ B(\mathcal{M}) $ and $ d $ depend on the unknown data manifold, they cannot be calculated. Additionally, $ \epsilon $ suffers from affinity drift towards low energy (high-density) regions \cite{nadler2006diffusion,driftLafon} thus, tuning $ \epsilon $ poses a challenge in computing true graph Laplacian of $ \mathcal{M} $.

    In this paper, we propose a variable Parzen window (VPW) estimator with three affinity adjustment factors to cater to unevenly sampled data points on $ \mathcal{M} $. The problem of designing a neighborhood supporting the globally fixed Parzen window (FPW) is changed to fitting local Parzen window on the available neighborhood. An accurate Parzen window minimizes the effect of $ \eta_{i} $ and hence, solves the problem of affinity drift. Due to unevenness, the neighborhoods of two connected data points exhibit properties of two different distributions which makes the problem severe. This is countered by employing additional affinity adjustment methods which utilize the local geometrical properties of the respective neighborhoods, thereby balancing the final affinity.
\begin{table*}[!h]
    \caption{Symbols and their description}
    \label{table:symbols}
    \centering
    \begin{singlespace}\scriptsize
    \begin{tabular}{cp{4cm}||cp{4cm}}
        \hline\hline
        \thead{Symbol}  & \thead{Description}                                   &       \thead{Symbol}       & \thead{Description}                                    \\ \hline
        $ \mathcal{M} $ & Riemannian manifold                                   &      $ \sigma^{2} $      & variance                                     \\
        $ g $      & Riemannian metric                                     &        $ {e}_{ij} $        & Euclidean distance between $ x_{i} $ and $ x_{j} $     \\
        $ D $      & ambient dimension                                     &           $ C $            & normalizing constant                                   \\
        $ d $      & intrinsic dimension                                   &      $ \phi(\cdot) $       & window function                                        \\
        $ x_{i} $    & one data point                                        &        $ \hat{P} $         & probability density estimation                         \\
        $ X $      & set of data points                                    &           $ R $            & a small region                                         \\
        $ n $      & total number of data points                           &           $ l $            & number of data points inside $ R $                     \\
        $ m $      & total number of labeled data points                   &           $ q $            & probability of $ l $ points being inside $ R $         \\
        $ f $      & data generating function                              &           $ E $            & expectation                                            \\
        $ \tau $     & original data on intrinsic dimension                  &          $ \mu $           & volume of manifold centered at $ x $                   \\
        $ \eta $     & unknown noise and redundancy                          &          $ \rho $          & flat region proportional to injectivty radius          \\
        $ G $      & graph on $ X $                                        &      $ \mathcal{V} $       & unit ball volume                                       \\
        $ a_{ij} $    & affinity between $ x_{i} $ and $ x_{j} $              &     $ \epsilon_{ij} $      & variable Parzen window between $ x_{i} $ and $ x_{j} $ \\
        $ W $      & affinity matrix                                       &    $ \varepsilon_{ij} $    & known factors between $ x_{i} $ and $ x_{j} $          \\
        $ N(x_{i}) $   & set of neighbors in a local neighborhood of $ x_{i} $ & $ \bar{\varepsilon}_{ij} $ & unknown factors between $ x_{i} $ and $ x_{j} $        \\
        $ |\cdot| $   & number of elements in the set                         &         $ b_{ij} $         & mean neighborhood distance between $ x_{i} $ and $ x_{j} $          \\
        $ u $      & heat distribution                                     &         $ c_{ij} $         & centroid distance between $ x_{i} $ and $ x_{j} $      \\
        $ \kappa $    & heat kernel                                           &       $ {bd}_{ij} $        & Bhattacharyya distance between $ x_{i} $ and $ x_{j} $ \\
        $ \Phi $     & a smooth function                                     &       $ \delta_{i} $       & variance in $ N(x_{i}) $                               \\
        $ \epsilon $   & fixed Parzen window                                   &        $ \nu_{i} $         & mean in $ N(x_{i}) $                                   \\
        $ \Lambda $   & diagonal matrix                                       &           $ Y $            & set of labels                                          \\
        $ L $      & graph Laplacian $ (L=\Lambda-W) $                     &        $ \lambda $         & smoothness in ambient and intrinsic dimension          \\ \hline
    \end{tabular}
    \end{singlespace}
\end{table*}
\section{Problem Description}
\label{sec:system_model_and_problem_description}
     If $ f:\mathcal{M}\rightarrow\mathbb{R} $ is a smooth function, then the bias and variance error terms given are \cite{laplacian02}
    \[ \frac{1}{\epsilon}\sum\limits_{j=1}^{n}{L_{ij}f(x_{j})}=\frac{1}{2}\Delta_{\mathcal{M}}f(x_{i})+O\bigg(\frac{1}{n^{1/2}\epsilon^{1+d/4}},\epsilon^{1/2}\bigg) \]
    where, $ L $ is the graph Laplacian, $ O(\epsilon^{1/2}) $ is bias error independent of sample size $ n $ and $ O(\frac{1}{n^{1/2}\epsilon^{1+d/4}}) $ represents the variance error. This shows that in the limit of $ n\rightarrow\infty $ and $ \epsilon\rightarrow0 $, discrete graph Laplacian pointwise converges to the continuous Laplace-Beltrami operator. Since, the sample size $ n $ is not controlled by the underlying model, it leaves only $ \epsilon $ to be tuned for convergence. A small $ \epsilon $ reduces the error, but a minimal value \cite{laplacian02} in low-density regions may result in a noisy estimator and a very large $ \epsilon $ might ignore features which could have been learned otherwise.
    
    In a setting of varying density regions, ideally, $ \epsilon $ should be large in low-density and small in high-density regions \cite{botev2010kernel} which is directly proportional to the maximum distance up to which surface remains linear \cite{scott1990feature}. A constant Parzen window may lead to inaccurate graph Laplacian convergence \cite{parzen1962estimation}.    
\section{Related Work}
\label{sec:related_work}
    The Parzen window for data points governed by a zero-mean, $ \sigma^{2} $ variance based normal distribution $ \mathcal{N}(0, \sigma^{2}) $ can be obtained using \cite{ConvergenceOfLaplacianEigenmapsbelkin2007,silverman1986density}
    \begin{equation}\label{eqn:empiricalFPW}
    \epsilon=\bigg(\frac{4\sigma^{5}}{3n}\bigg)^{\frac{1}{5}}\approx 1.06\sigma n^{1/5}
    \end{equation}
    $ \sigma $ is the standard deviation over $ n $ data points. However, when $ \mathcal{M} $ contains uneven sampled data, the spectrum of the Laplacian may not converge to the underlying Laplace-Beltrami operator and requires $ \epsilon $ to be tuned. The existing state-of-the-art Parzen window estimators can be broadly classified under two categories:
    
    \textbf{Globally fixed Parzen window:} These estimators assume that data is evenly sampled and hence, a global FPW gives the true affinity between two connected data points. Lepski's procedure \cite{chazal2016data} creates a setting to find an optimal estimator from the set of estimators by fixing a target quantity and tight upper bound on variance error of the estimator. Parzen window estimation based on geometrical consistency \cite{joncas2017improved} fix the target geometry of the manifold from given points of cloud. Further, $ \epsilon $ parameter is tuned to minimize the error for maximum geometric preservation.
    
    \textbf{Adaptive Parzen window:} An adaptive Parzen window tries to exploit the diverse linearity region and heterogeneous neighborhood size of the underlying manifold by defining a custom Parzen window for every neighborhood. Authors in  \cite{NIPS2006_3043} suggest finding a local Parzen window in each dimension by optimizing the function which minimizes the entropy on unlabeled data points. Similarly, \cite{Karasuyama2017} proposes to find the Parzen window estimator in each dimension by solving the local linear embedding function in every neighborhood. Self-tuning parameter \cite{zelnik2005self,Tasdemir:2015:ASC:2801150.2801385} suggests using the distance from the point of interest to its $ k $\textsuperscript{th} neighbor as its Parzen window. In  \cite{ManifoldParzenWindowsVincent2003}, the Parzen window is estimated from the normal distribution at each data point using the mean vector and covariance matrix. Non-local manifold Parzen window \cite{NonLocalManifoldParzenWindowBengio2006} uses neural networks to predict the width, density, and covariance matrix around each data point for identifying the non-local manifold density structure. Adaptive kernel density estimation \cite{zhang2018adaptive} proposes to approximate local Parzen window by subtracting the average edge weight in the local neighborhood from the sum of minimum and maximum edge weight of the same neighborhood. Entropic affinities (EA) \cite{entropicAffinities} calculates adaptive bandwidth on the user-defined perplexity hyper-parameter. Apart from heat kernel based affinity, the pairwise similarity on $ \mathcal{M} $ can also be determined using CONN \cite{Tasdemir:2012:VQB:2181340.2181723}, R-convolution kernels \cite{ROSSI20153357}, and sparsity connection \cite{li2018affinity}.
\section{Riemannian Manifold and Affinity}
\label{sec:proposed_technique}
    In manifold learning and manifold regularization, it is assumed that input data intrinsically lies on a lower dimensional manifold embedded in the higher dimension ambient space. On a smooth Riemannian manifold $ (\mathcal{M},g) $, the geometry is contained in the field of metric tensors. At each point $ x_{i} $, the tensor $ g(x_{i}) $ defines an inner product on the tangent space and a metric in a neighborhood of $ N(x_{i}) $ via the exponential map.
    On $ \mathcal{M} $, the heat kernel encapsulates the distribution of geodesic distances and the solution is given by exponentiating the Laplacian eigen system over time \cite{kernelJayasumana2015}. In the presence of infinite sampled data points, the heat kernel $ (\kappa) $ derived graph Laplacian converges to its respective Laplace-Beltrami spectrum \cite{convergenceSinger,singer2016spectral}.
    \[ \Delta_{\epsilon,n}f(x_{i}):=\frac{1}{n+\epsilon^{d+2}}\sum_{j=1}^{n}\kappa\bigg(\frac{x_{i}-x_{j}}{\epsilon}\bigg)(f(x_{i})-f(x_{j})) \]
    The affinity in discrete graph can be defined as a distance metric $ e_{ij} $ on connected pair of data points $ x_{i} $ and $ x_{j} $ of the graph $ G $,
    \[ e_{ij}=\parallel x_{i}-x_{j}\parallel^{2}_{2} \]
    The data points $ x_{j} $s are considered to be neighbors of $ x_{i} $ i.e. $ x_{j}\in N(x_{i}) $, if $ x_{j} $ lies in the locally linear region of $ \mathcal{M} $ around $ x_{i} $. In such a linear region, $ e_{ij} $ exhibits slow spatial variations, but due to density variation on $ \mathcal{M} $, it may vary rapidly and unpredictably. Therefore, the edge weight needs to be replaced by affinity with coefficients that decay with dissimilarity. The heat kernel based affinity $ a_{ij} $ employed over Euclidean distance gives large value for spatially near points and decays exponentially as the distance increases.
    \begin{equation}
    \centering
    a_{ij} = \frac{1}{C}\exp\bigg(\frac{-e_{ij}}{2\epsilon^{2}}\bigg)
    \label{equation:basicAffinity}
    \end{equation}
    here, $ C>0 $ is the normalizing constant. The dynamically varying curvature due to unevenly sampled data points on $ \mathcal{M} $ form sub-groups with different density. Hence, the focus is to find $ \epsilon $ based on the respective neighborhoods of two connected data points $ x_{i} $ and $ x_{j} $.
\subsection{Variable Parzen Window (VPW)}
\label{sec:vpwam}
    In the input space, the probability density estimation at point $ x_{i} $ on $ \mathcal{M} $ is given by
    \[ \hat{P}(x_{i})=\frac{1}{|N(x_{i})|}\sum\limits_{j\in N(x_{i})}\phi(x_{i}-x_{j}, \epsilon) \]
    $ \phi(\cdot) $ is the window function and $ \epsilon $ represents the Parzen window. If $ \phi(\cdot) \text{ and } \epsilon $ are chosen correctly, and the true probability density is constant in the chosen region, then $ \hat{P}(x_{i}) $ converges to its true density \cite{NonLocalManifoldParzenWindowBengio2006,parzen1962estimation}. Ideally, $ \epsilon $ should be tuned based on the density of the region on the manifold. Let $ R $ be a small region on manifold following Euclidean properties $ \int_{R} P(x) dx $, only $ l $ out of $ n $ data points fall inside $ R $ with probability $ q $ i.e. $ \bigg(\substack{n\\\\l} \bigg)q^{l}(1-q)^{n-l} $. The expected fraction of $ l $ is calculated by obtaining expectation from $ E[l/n]=q $ and variance by $ var[l/n]=q(1-q)/n $. As $ n\rightarrow \infty $, the variance becomes $ 0 $ and the fraction peaks around the expectation, $ k\approx nq $, then, $ q\approx P(x)\cdot\mu $. Here, $ \mu $ is the volume of $ \mathcal{M} $ centered at $ x $, $ \mu=\sum_{i=1}^{n}\mu_{i}\delta_{x_{i}} $. $ \mathcal{M} $ is further divided into measurable subsets $ V_{i}\subset R, i=1,2\dots n $ and $ vol(V_{i})=\mu_{i} $. Assume $ G=G(X,W,\mu,\rho) $ be a weighted undirected graph with vertices $ X $, weight $ W $, volume $ \mu $ and injectivity radius $ \rho $ in a $ D $ dimensional ambient space manifold $ \mathcal{M} $. Then, the Parzen window $ \epsilon_{ij} $ between two connected data points $ x_{i} $ and $ x_{j} $ can be approximated using an edge with a constant weight between them \cite{GraphDiscretizationBurago2013}
    \[ \epsilon_{ij}=\frac{2(D+2)}{\mathcal{V}_{D}\rho^{D+2}}\mu_{i}\mu_{j} \]
    $ \mathcal{V}_{D} $ is a unit ball volume in $ \mathbb{R}^{D} $. The degree of a vertex in the graph can be approximately expressed as $ \frac{\mathcal{V}_{D}\rho^{D}}{\mu_{i}} $ which is same as the number of neighbors, therefore,
    \[ |N(x_{i})|=\frac{\mathcal{V}_{D}\rho^{D}}{\mu_{i}};|N(x_{j})|=\frac{\mathcal{V}_{D}\rho^{D}}{\mu_{j}} \]
    \[ \implies \epsilon_{ij}=\frac{2(D+2)}{|N(x_{i})||N(x_{j})|}\frac{\mathcal{V}_{D}\rho^{D}}{\rho^{2}} \]
    If the weights $ W $ are constant, then the weight at the vertex $ \mu_{i}=\mu_{j}=\mu_{0} $ and $ \mu_{0}=\frac{\mathcal{V}_{D}\rho^{D}}{|N(x_{0})|} $ where, $ |N(x_{0})|=\frac{1}{n}\sum_{i=1}^{n}|N(x_{i})| $ is the average number of neighbors.
    \[ \therefore \mathcal{V}_{D}\rho^{D}=\mu_{0}|N(x_{0})| \]
    replacing $ \mathcal{V}_{D}\rho^{D} $ in $ \epsilon_{ij} $,
    \begin{equation}
    \centering
    \implies \epsilon_{ij}=\frac{2(D+2)}{|N(x_{i})||N(x_{j})|}\frac{\mu_{0}|N(x_{0})|}{\rho^{2}}=\overbrace{\frac{2(D+2)|N(x_{0})|}{|N(x_{i})||N(x_{j})|}}^{\varepsilon_{ij}}\underbrace{\frac{\mu_{0}}{\rho^{2}}}_{\bar{\varepsilon}_{ij}}
    \label{equation:localParzenCalculation}
    \end{equation}
    To determine the affinity between a data point $ x_{i} $ and its neighbor $ x_{j} $, we introduce weights on edges depending on the Euclidean distance $ e_{ij} $ normalized by $ \rho $. The probability of sharing similar labels between data points $ x_{i} $ and $ x_{j} $ is higher when they are spatially closer. Hence, the affinity between points $ x_{i} $ and $ x_{j}  $ can be expressed as a function $ \kappa(e_{ij},\rho,\epsilon_{ij}) $
    \[ a_{ij}=\kappa\bigg( \frac{e_{ij}}{\rho} \frac{1}{\epsilon_{ij}} \bigg) \]
    since, $ \epsilon_{ij}=\varepsilon_{ij}\bar{\varepsilon}_{ij} $
    \begin{equation}
    \centering
    \implies a_{ij} = \kappa\bigg(\frac{e_{ij}}{\varepsilon_{ij}} (\rho/\mu_{0})\bigg)
    \label{equation:finalAffinity}
    \end{equation}
    where, $ \kappa $ is the heat kernel, thus, $ a_{ij}=\frac{1}{C}\exp\bigg(-\frac{e_{ij}}{\varepsilon_{ij}} (\rho/\mu_{0})\bigg) $. $ \mu_{0} $ and $ \rho $ are unknown and depend respectively on the underlying distribution of the data points and the extent to which the manifold is flat. Further, $ (\rho/\mu_{0}) $ factor needs to be adjusted to account for the sampling unevenness and flatness range of the local manifold region.
    
    It can be observed that for an even sampled data points cloud on $ \mathcal{M} $ with constant curvature, $ (\rho/\mu_{0}) $ also remains constant for the neighborhoods of all data points. In the case of $ \mathcal{M} $ with varying curvature due to unevenly sampled data points or when the extent of locally linear region is gauged by density of data points in the neighborhood, the ratio $ (\rho/\mu_{0}) $ is observed to be different for different neighborhoods. This needs to be factored in the affinity calculation.
\subsection{Affinity Adjustment}
\label{sec:lnas}
    On the manifold ($ \mathcal{M} $), we must be able to estimate the local data geometry so that the graph Laplacian converges to the Laplace-Beltrami operator. $ \epsilon $ is dependent on the extent to which the underlying sampling density of the point cloud is constant. The heat kernel estimation with FPW tends to smoothen crest and trough of the distribution \cite{botev2010kernel}. It however is oblivious to local variability. This requires affinity adjustment to incorporate the local variation in data density.
    
    It is clear that the effect of uneven sampling manifests in the form of variation in curvature which effects the extent of local linear region around a data point and the change in the neighborhood size around different data points. This also effects the pairwise affinity between two neighboring data points and introduces a bias which needs to be factored in the affinity computation.
    
    \paragraph{\textbf{Non-local means based affinity adjustment:}} The affinity bias between two data points which is a function of $ (\rho/\mu_{0}) $ that induces a deviation in the size of the neighborhood due to the change in the linear region and hence, the number of data points in the region. This induced effect can be viewed as the difference in neighborhood sizes between two adjacent data points $ x_{i} $ and $ x_{j} $. The edge weight $ e_{ij} $ needs to be adjusted to offset this bias.
    
    Let $ x_{i} $ and $ x_{j} $ be two data points on $ \mathcal{M} $ where, $ x_{j}\in N(x_{i}) $ i.e. $ x_{j} $ is a member of $ x_{i} $'s neighborhood. Based on the manifold assumption, the closeness between these two data points calculated (Eqn. \ref{equation:basicAffinity}) determines whether they shall share a similar label or not. The Euclidean distance between these two data points, assumed to be on a flat surface defines a direct relationship between them. However, to balance for the uneven sampling of data, the effect on individual neighborhoods should be considered. This measure of affinity can be seen as a non-local similarity between two neighborhoods centered around data points $ x_{i} $ and $ x_{j} $. This follows from the non-local means algorithm \cite{nonlocalmean} based on a non-local averaging of all neighbors in respective neighborhoods. 
    For discrete noisy $ x_{j\in N(x_{i})} $, the estimated value $ b_{i} $ is a average of all distances in the neighborhood
    \[ b_{i}=\frac{1}{|N(x_{i})|}\sum_{j\in N(x_{i})}e_{ij} \]
    where the weights $ e_{ij} $, depend on the similarity between the each pair of neighbors $ x_{i} $ and $ x_{j} $.
    
    We assume that the data points are generated by a stationary random process. Thus, for an $ x_{i} $, the non-local means algorithm converges to the conditional expectations of $ x_{i} $, given its neighboring data points. A similarity measure between two neighbors must take into account the conditional expectations of the observations. Accordingly, the similarity between two neighbors $ x_{i} $ and $ x_{j} $ must consider the spatial distance between them conditioned on the similarity between their respective neighborhoods. We modify the above notion of non-local means to develop a measure of similarity between neighborhoods of two data points.
    \[ b_{ij}=\parallel b_{i}-b_{j}\parallel_{2}^{2} \]
    The non-local means based measure $ b_{ij} $ considers the geometrical configuration in a whole neighborhood and will be significant when the data points are uneven sampled. The difference $ b_{ij} $, thus, represents the difference the conditional expectations of $ x_{i} $ and $ x_{j} $ given their respective neighborhoods. The affinity based on edge weight $ e_{ij} $ adjusted by $ b_{ij} $ neighborhood similarity is given by
    \begin{equation}\label{equation:lnas}
    \centering
    a_{ij} = \frac{1}{C}\exp\bigg(-\frac{e_{ij}^{2}}{\varepsilon_{ij}^{2}}\bigg)\exp\bigg(-\frac{b_{ij}^{2}}{\varepsilon_{ij}^{2}}\bigg)
    \end{equation}
    here, $ C>0 $ is a normalizing constant and $ \varepsilon_{ij}>0 $ is the normalizing constant for variable Parzen window inside small region $ (x_{i},x_{j})\in R $.
    \paragraph{\textbf{Centroid distance based affinity adjustment:}} The bias induced by $ (\rho/\mu_{0}) $ can be viewed as being introduced by a change in the parameters of the underlying distribution in the neighborhoods of connected data points \cite{NonLocalImageDenoisingBuades2008}. This can be considered as the distribution of data points around both $ x_{i} $ and $ x_{j} $ respectively in the affinity calculation. Thus, the bias may be offset by modifying $ e_{ij} $ by the distance between the centroids of the neighborhoods around $ x_{i} $ and $ x_{j} $ respectively. 
    We define the similarity as the Euclidean distance between the centroids of the neighborhoods of $ x_{i} $ and $ x_{j} $, respectively along with $ e_{ij} $. In the neighborhood $ N(x_{i}) $, the centroid is given by
    \[ c_{i}=\int_{R}\frac{1}{|N(x_{i})|}\sum_{j\in N(x_{i})}x_{j} dx \]
    $ R $ is the assumed linear region, $ N(x_{i}) $ contains the local neighbors of $ x_{i} $ and $ c_{i} $ is the centroid of the neighborhood. The shift due to varying density is captured by $ \gamma_{i} $ in $ N(x_{i})=c_{i}+\gamma_{i} $. The spatial closeness between any $ x_{i} $ and $ x_{j} $ needs to be modified by the geometric similarity between the neighborhood centers of the two data points. The combination of this spatial closeness and geometric similarity discards the $ (\rho/\mu_{0}) $ bias and takes care of the non-uniform distribution of data in $ R $. The centroid based adjustment factor can be calculated as
    \[ c_{ij}=\parallel c_{i}-c_{j}\parallel^{2}_{2} \]
    $ c_{ij} $ is the distance between centroids of $ x_{i} $ and $ x_{j} $ utilized to adjust the unevenness in the sampling of the data point cloud. Then, affinity including centroid distance is
    \begin{equation}
    \centering
    a_{ij} = \frac{1}{C}\exp\bigg(-\frac{e_{ij}^{2}}{\varepsilon_{ij}^{2}}\bigg)\exp\bigg(-\frac{c_{ij}^{2}}{\varepsilon_{ij}^{2}}\bigg)
    \label{equation:cbas}
    \end{equation}
    here, $ C>0 $ is a normalizing constant and $ \varepsilon_{ij}>0 $ is the normalizing constant for variable Parzen window in a small region $ (x_{i},x_{j})\in R $.
    \paragraph{\textbf{Bhattacharyya distance based affinity scale:}} In addition to the centroids, the variances in the neighborhoods may also be considered for adjustment. This can be done by using Bhattacharyya distance between two connected neighborhoods. Given two observations $ x_{i}\text{ and } x_{j} $ such that $ x_{j}\in N(x_{i}) $, as the local geometrical properties around them differs due to different neighborhood size and $ \rho $. It can be assumed that both points have been drawn from two separate distributions which hold true for their respective neighbors. This additional factor based on local distribution properties along with the Euclidean metric $ e_{ij} $ helps to balance the effect of $ (\rho/\mu_{0}) $. Bhattacharyya distance measures the similarity between two probability distributions over the same domain. The Bhattacharyya distance is
    \[ bd_{ij}=\frac{1}{4}\log\bigg(\frac{1}{4}\bigg(\frac{\delta_{i}^{2}}{\delta_{j}^{2}}+\frac{\delta_{j}^{2}}{\delta_{i}^{2}}+2\bigg)\bigg)+\frac{1}{4}\bigg(\frac{(\nu_{i}-\nu_{j})^{2}}{\delta_{i}^{2}+\delta_{j}^{2}}\bigg) \]
    where, $ bd_{ij} $ is the Bhattacharyya distance between $ x_{i}\text{ and }x_{j} $, $ \delta_{i}^{2} $ and $ \nu_{i} $ are the variance and mean of $ N(x_{i}) $ respectively. The final affinity using Euclidean distance $ e_{ij} $ with $ bd_{ij} $ adjustment is given by
    \begin{equation}\label{equation:bdas}
    a_{ij} = \frac{1}{C}\exp\bigg(-\frac{e_{ij}^{2}}{\varepsilon_{ij}^{2}}\bigg)\exp\bigg(-\frac{{bd}_{ij}^{2}}{\varepsilon_{ij}^{2}}\bigg)
    \end{equation}
    here, $ C>0 $ is a normalizing constant and $ \varepsilon_{ij}>0 $ is the normalizing constant for variable Parzen window specific to the small region $ (x_{i},x_{j})\in R $.
    
\subsection{Graph Laplacian manifold regularization}
\label{sec:graphLapReg}
    Given, affinity $ W=\{a_{ij}\}_{i=j=1}^{n} $ obtained from methods defined in the previous section, compute a diagonal matrix containing the sum of each row $ \Lambda=\bigg\{\sum_{j=1}^{n}a_{ij}\bigg\}_{i=1}^{n} $. Then, graph Laplacian is calculated from
    $ L = \Lambda-W $ where, $ L $ holds the graph $ G $'s spectrum and defines the divergence of the function at every data point on $ \mathcal{M} $. The objective function to find the optimal candidate function $ f $ is defined by
    \[ f^*= \argmin_{f \in \mathcal{H}_{K}} \Psi(Y_{m},X_{m},f) + {\lambda_{A} \parallel f \parallel^{2}}+ {\lambda_{I} \parallel R(f) \parallel^{2}} \]
    here, $ \Psi $ is a loss function, $ X $ is the input observations, $ Y $ contains the labels for respective $ m $ number of samples, $ \lambda_{A}\text{ and }\lambda_{I} $ defines the function smoothness weight on ambient and intrinsic space respectively. Manifold regularization over $ f $ using $ n-m $ unlabeled samples is obtained through
    \[ R(f)= \frac{1}{2} \sum_{i=j=1}^{n}(f(x_{i})-f(x_{j}))^{2}\, a_{ij} \]
    \[ \implies R(f)=\sum_{i=1}^{n} f(x_{i})^{2} \sum_{j=1}^{n}a_{ij}-\sum_{i=j=1}^{n}a_{ij}f(x_{i})f(x_{j})=f^{T}Df-f^{T}Wf=f^{T}Lf \]
    where, $ \sum_{j=1}^{n}a_{ij}=\Lambda $ and $ \sum_{i=j=1}^{n}a_{ij}=W $. The extended Representer Theorem \cite{manifold_regularization} states that optimal $ f $ exists in $ \mathcal{H}_{K} $ and is given by $ f^{*}(x)= \sum_{i=1}^{n} \alpha_{i} \vartheta(x_{i}, x) $ here, $ \vartheta $ is a positive-definite real-valued kernel and $ \alpha_{i} $ is the representation coefficient.
    \[ \therefore\mathbf{f}= \bigg [\sum_{i=1}^{n} \alpha_{i} \vartheta(x_{i}, x_{1}), \dots, \sum_{i=1}^{n} \alpha_{i} \vartheta(x_{i}, x_{n})\bigg ]^{T}= \boldsymbol{\vartheta}\boldsymbol{\alpha} \]
    here, $ \boldsymbol{\vartheta} $ is the kernel gram matrix and $ \boldsymbol{\alpha} $ is a vector of representation coefficients. 
    The final prediction function of semi-supervised graph Laplacian based regression least squares classifier (LapRLSC) \cite{belkinOnManifold,manifold_regularization} is obtained by replacing $ \Psi $ with square loss and taking the partial derivative on $ \frac{\partial f}{\partial\boldsymbol{\alpha}}=0 $ 
    \begin{equation}\label{equation:prediction}
    \centering
    \boldsymbol{\alpha}^{*}= \big (\boldsymbol{\vartheta}_{m}\boldsymbol{\vartheta}_{m}^{T} +\lambda_{A} \boldsymbol{\vartheta} +\lambda_{I}\boldsymbol{\vartheta}L\boldsymbol{\vartheta}\big)^{-1}\boldsymbol{\vartheta}_{m}Y_{m}
    \end{equation}
\section{Experiment and Analysis}
\label{sec:results_and_discussion}
    In this section, we evaluate the proposed VPW estimator on various synthetic and real-world data sets\footnote{Code available at https://github.com/gitr00ki3/vpw}. The performance has been further compared with three FPWs and three adaptive Parzen window methods. VPW with non-local means, centroid and Bhattacharyya based affinity have been denoted using VPW{$ _{B} $}, VPW{$ _{C} $} and VPW{$ _{BD} $} respectively. Similarly, FPW methods have been represented using FPW, FPW{$ _{\hat{\mu}} $} and FPW{$ _{\sigma} $} containing user-defined, mean distance ($ \frac{1}{n}\sum_{i=1}^{n}a_{i} $) and empirical (Eqn. \ref{eqn:empiricalFPW}) values. As VPW works by defining bandwidth for each pair of connected vertices, it is desirable to compare it with existing adaptive local Parzen window methods denoted by K\textsubscript{7} \cite{zelnik2005self}, MMM \cite{zhang2018adaptive}, and EA\footnote{Default perplexity=number of neighbors-1} \cite{entropicAffinities}. In the experiments, Laplacian eigenmap and LapRLSC have been used for non-linear dimensionality reduction and semi-supervised classification, respectively. The number of nearest neighbor parameter has been denoted by $ |N| $. 
\subsection{Toroidal Helix}
\label{sec:toroidal_helix}
    \begin{figure*}[!h]
        \centering
        \subfloat[Toroidal helix]{\includegraphics[width=0.16\linewidth,clip,keepaspectratio]{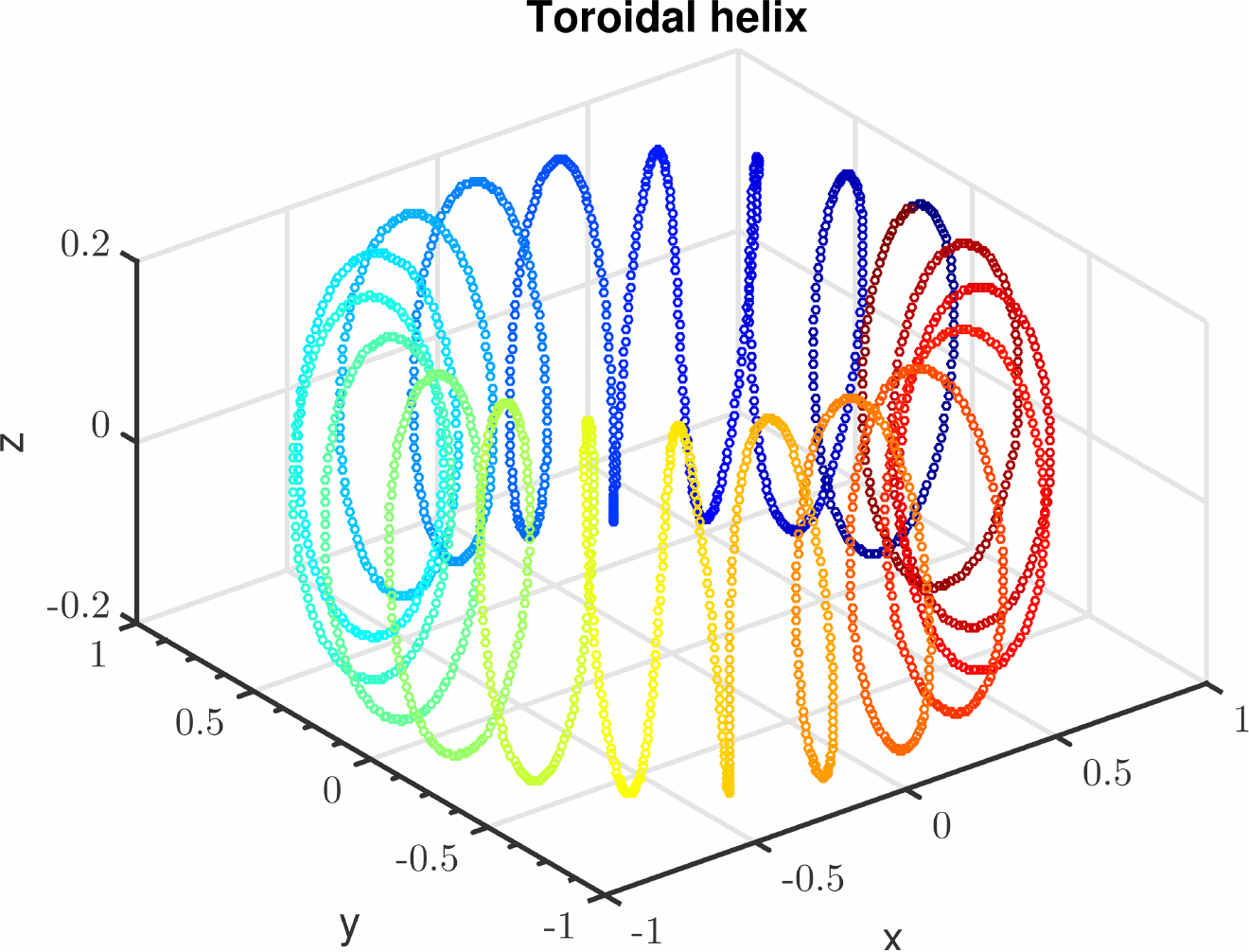}%
            \label{fig_toroidal}}
        \hfil
        \subfloat[FPW Laplacian]{\includegraphics[width=0.16\linewidth,clip,keepaspectratio]{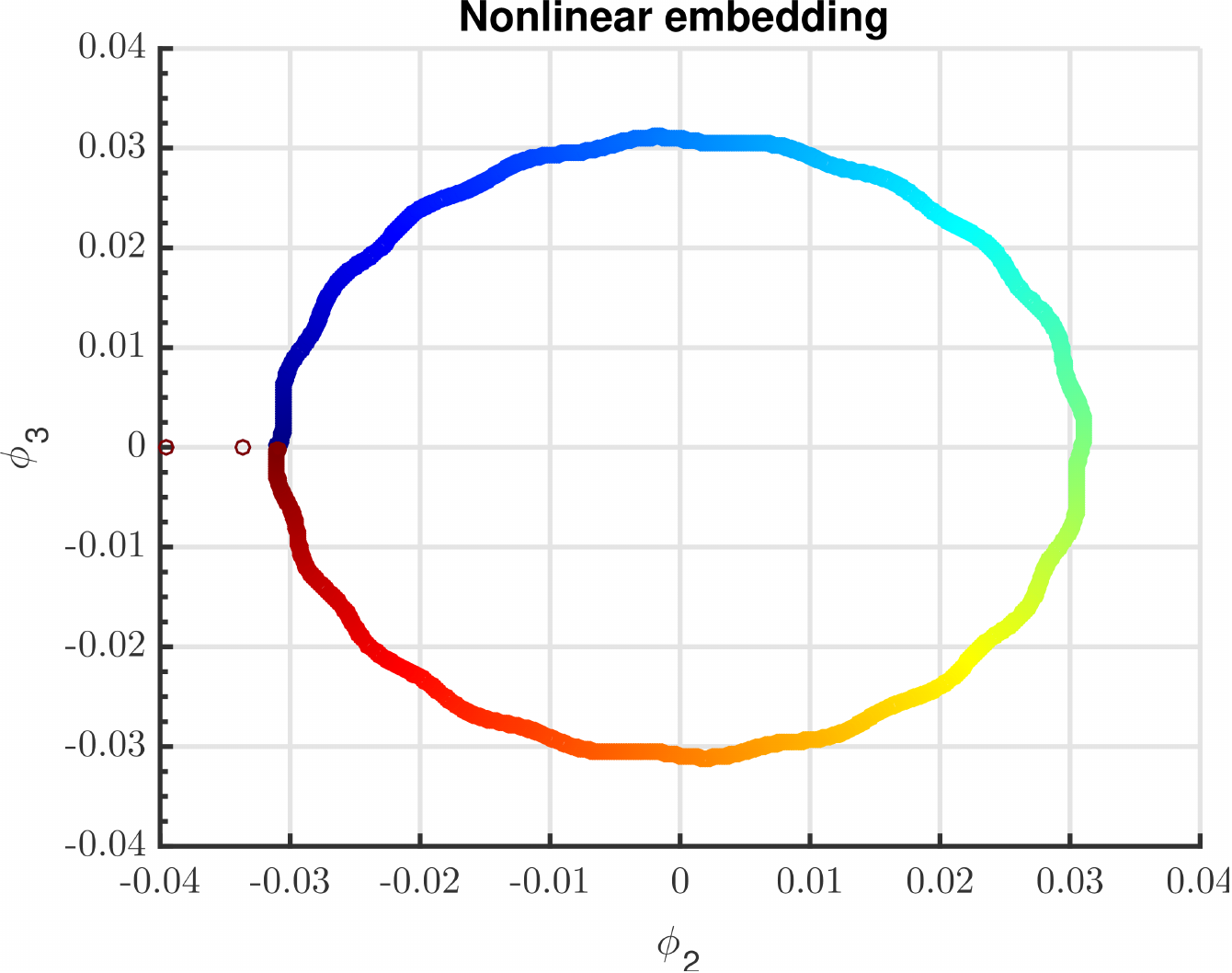}%
            \label{fig_toroidalLap}}
        \hfil
        \subfloat[FPW{$ _{\hat{\mu}} $} Laplacian]{\includegraphics[width=0.16\linewidth,clip,keepaspectratio]{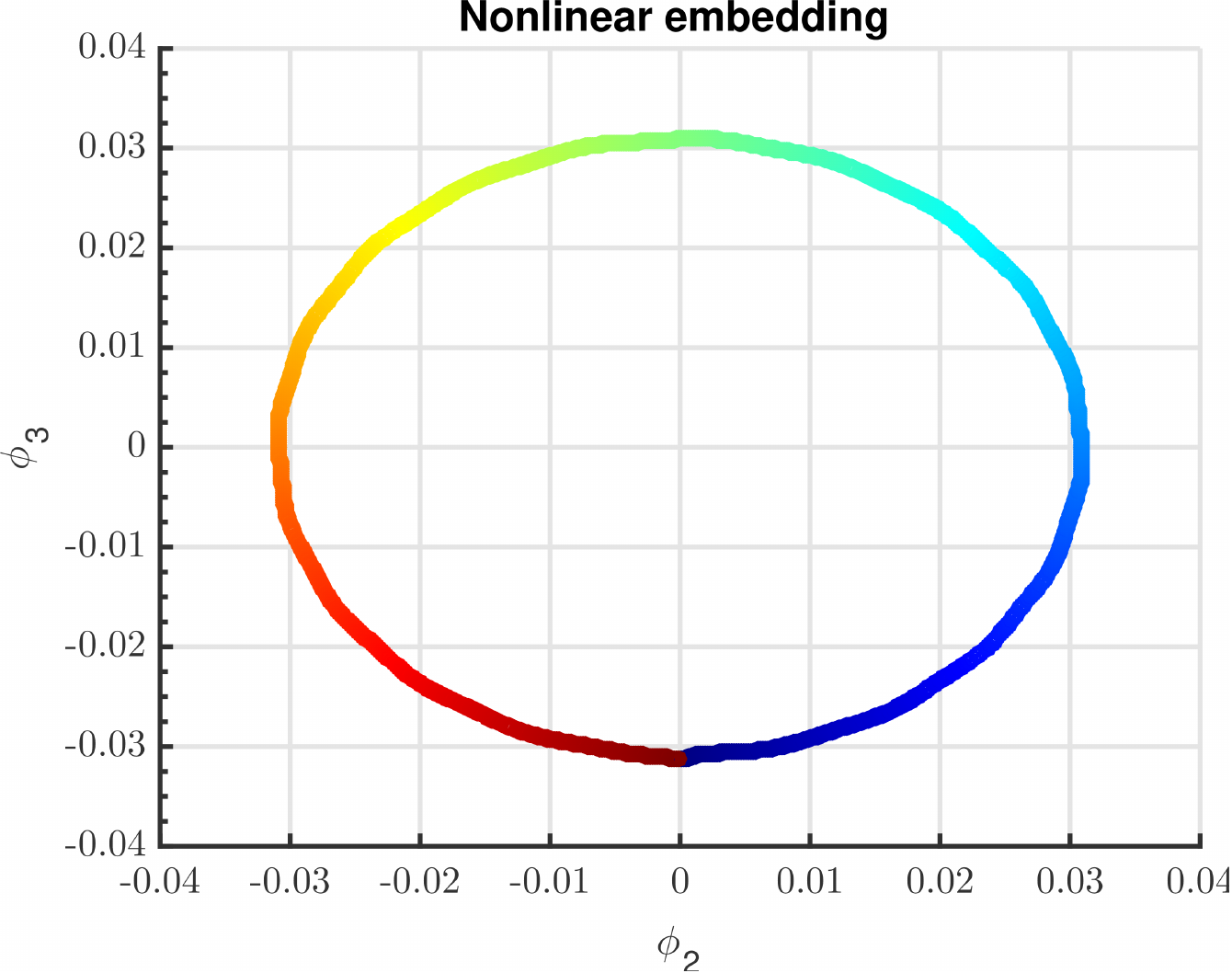}%
            \label{fig_toroidal_mu}}
        \hfil
        \subfloat[FPW{$ _{\sigma} $} Laplacian]{\includegraphics[width=0.16\linewidth,clip,keepaspectratio]{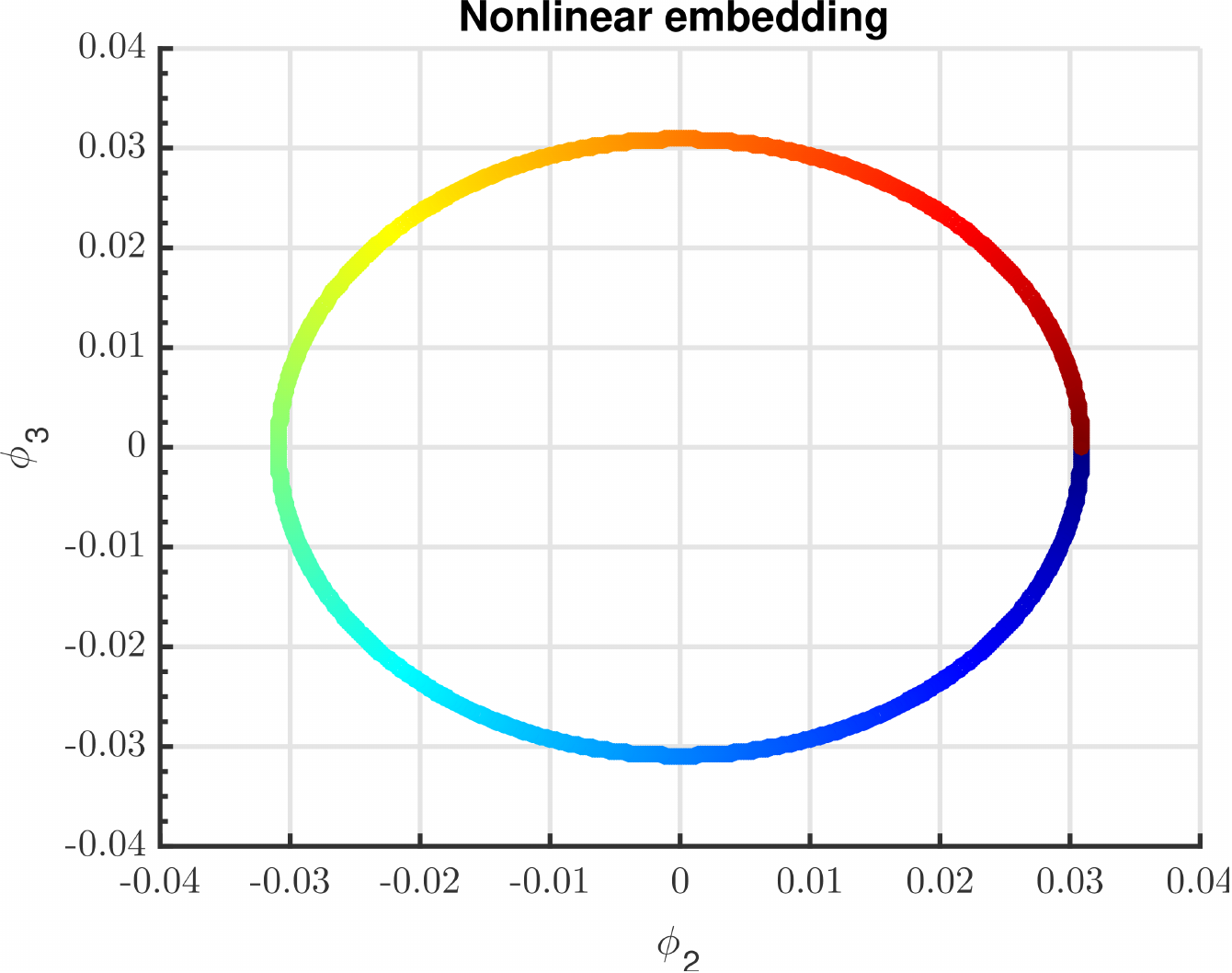}%
            \label{fig_toroidal_csig}}
        \hfil
        \subfloat[K\textsubscript{7}]{\includegraphics[width=0.16\linewidth,clip,keepaspectratio]{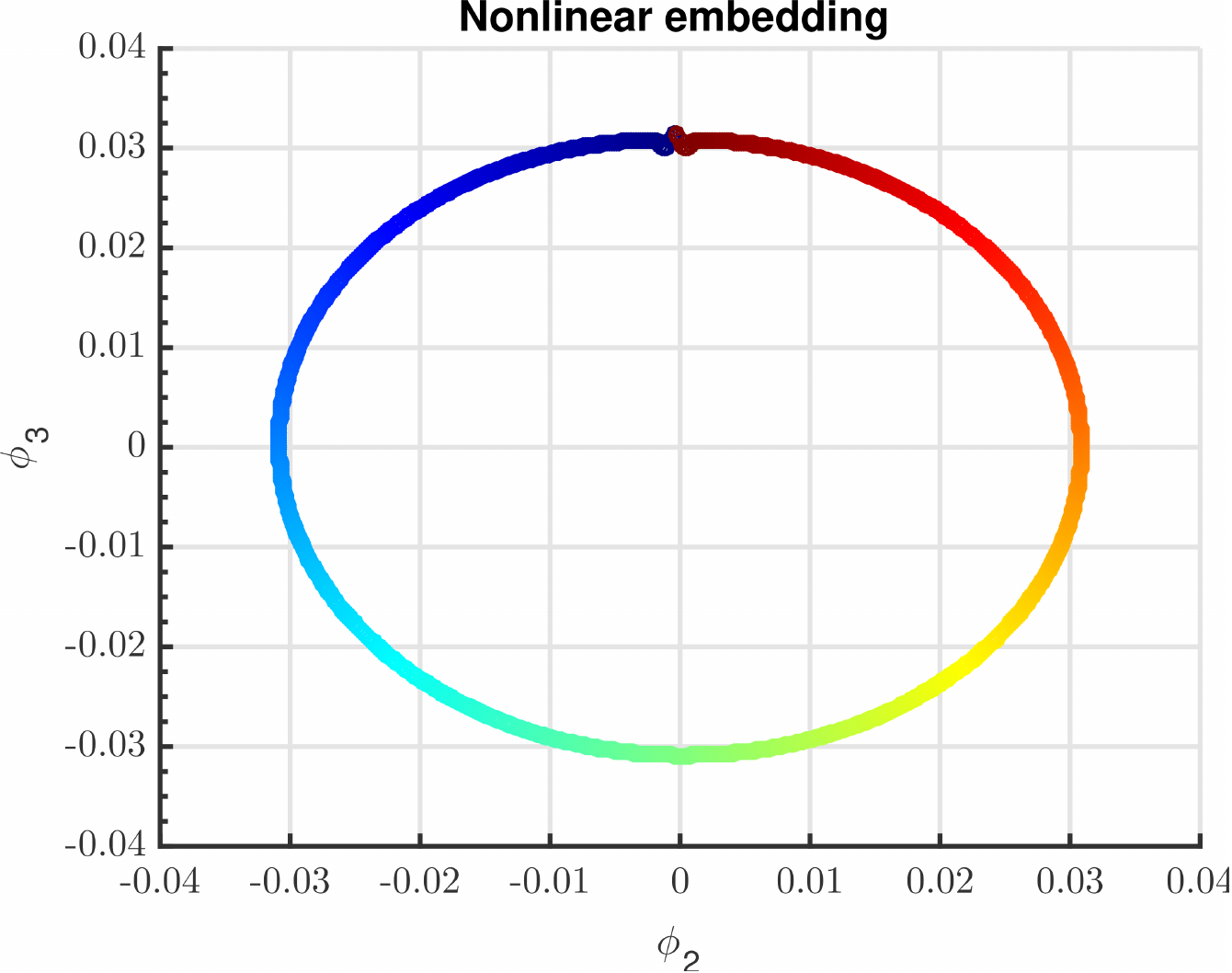}%
            \label{fig_toroidal_k7}}
        \hfil\\
        \subfloat[MMM]{\includegraphics[width=0.16\linewidth,clip,keepaspectratio]{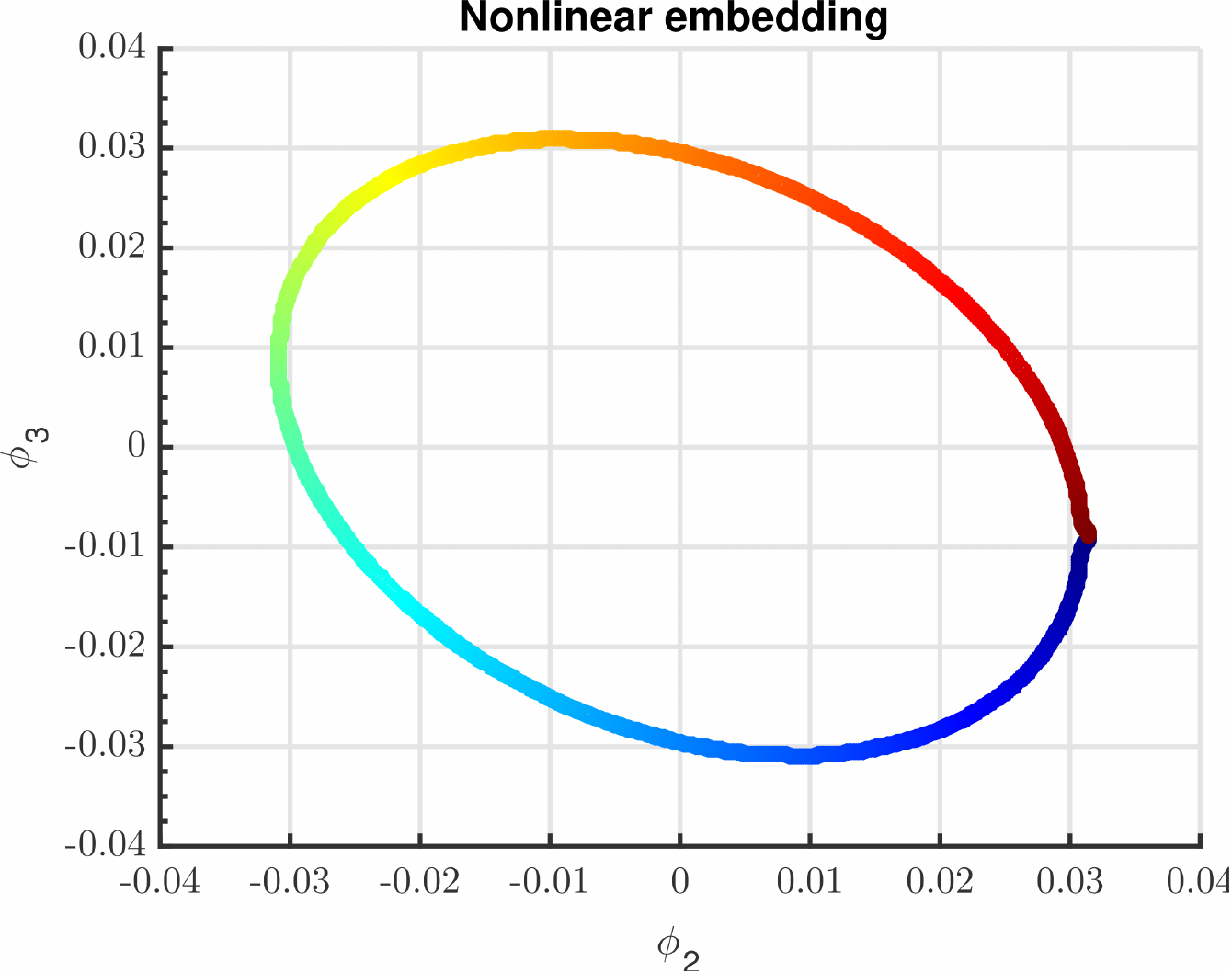}%
            \label{fig_toroidal_mmm}}
        \hfil
        \subfloat[EA]{\includegraphics[width=0.16\linewidth,clip,keepaspectratio]{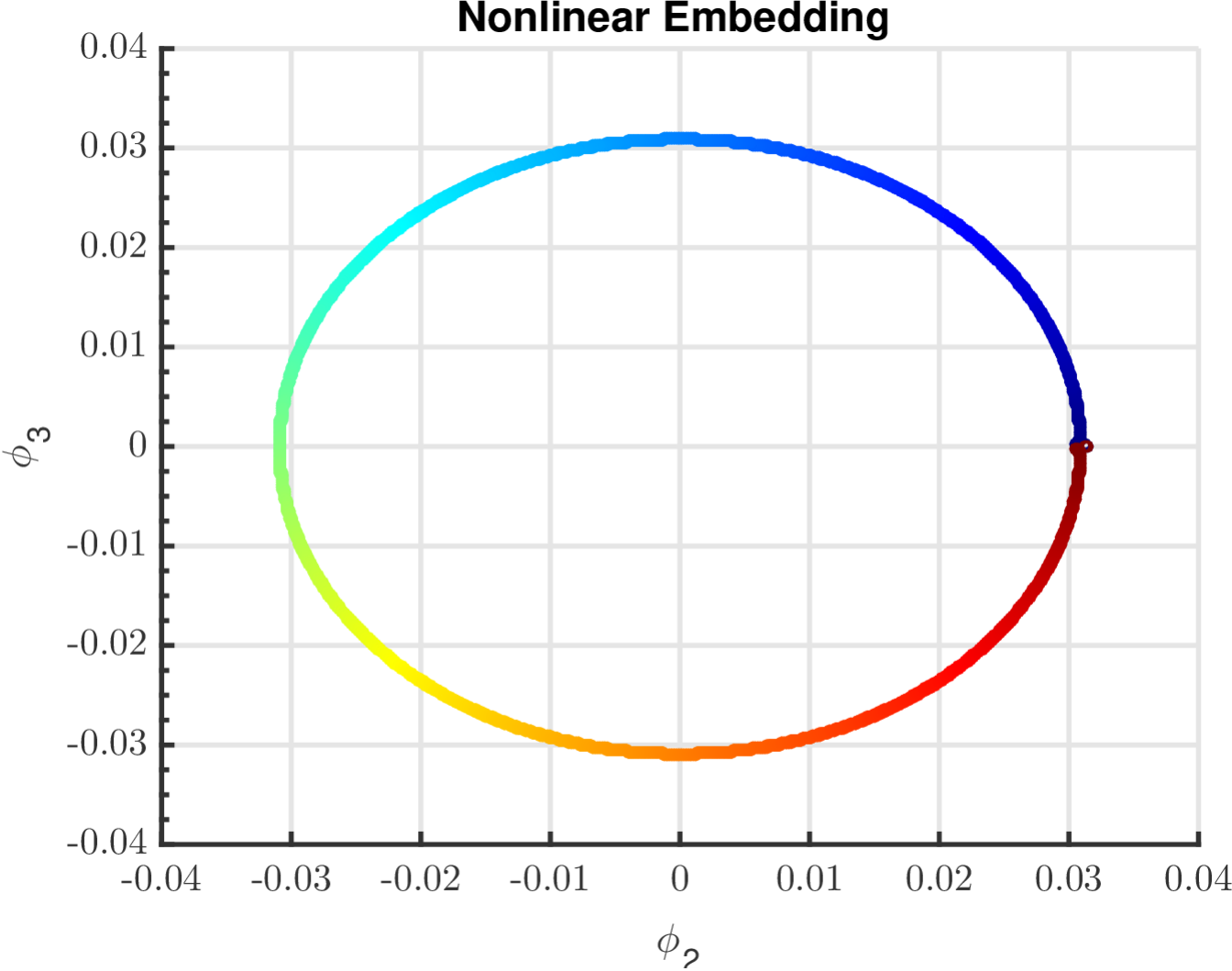}%
            \label{fig_toroidal_ea}}
        \hfil
        \subfloat[VPW{$ _B $} Laplacian]{\includegraphics[width=0.16\linewidth,clip,keepaspectratio]{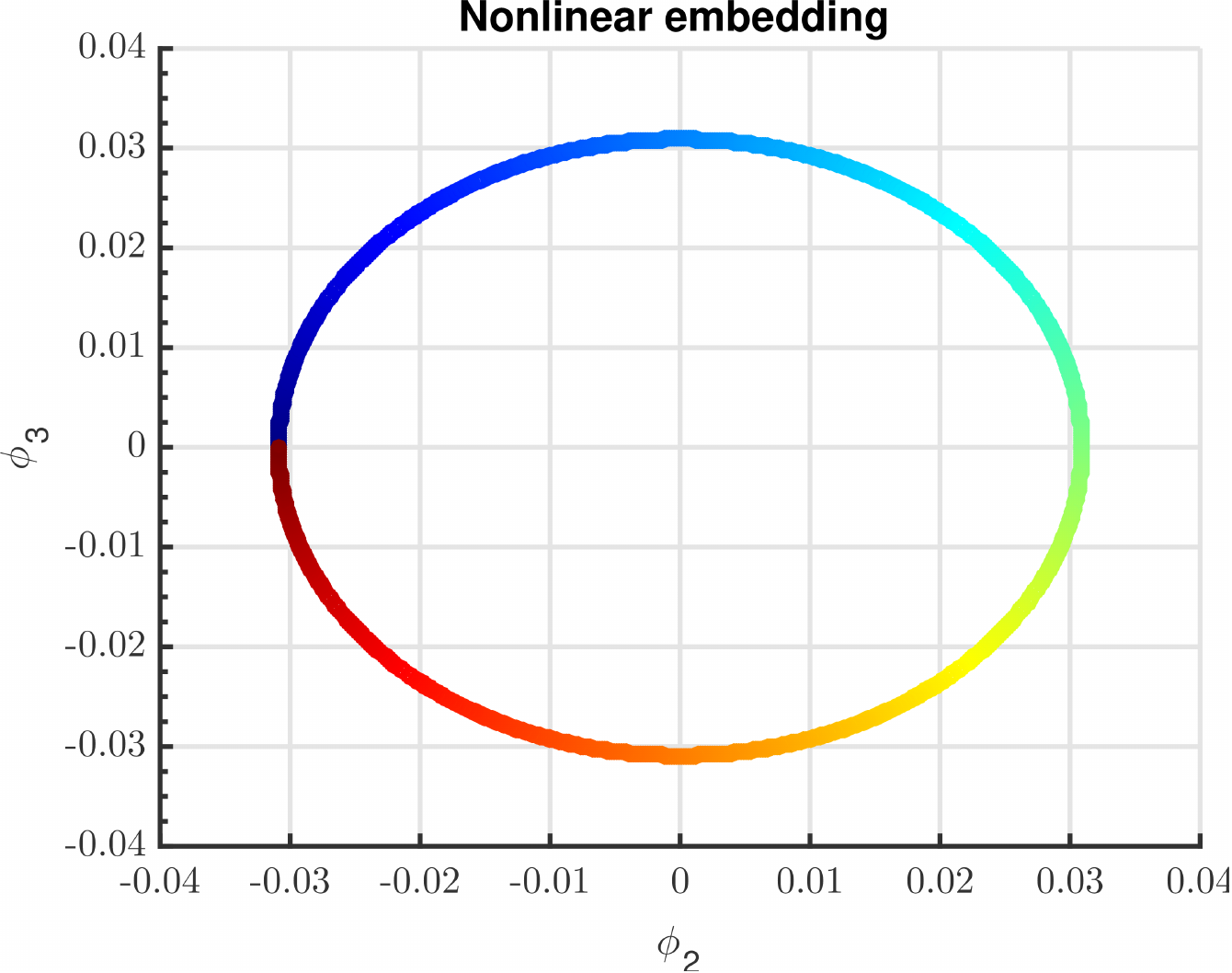}%
            \label{fig_toroidalVpw}}
        \hfil
        \subfloat[Eigenvalues]{\includegraphics[width=0.16\linewidth,clip,keepaspectratio]{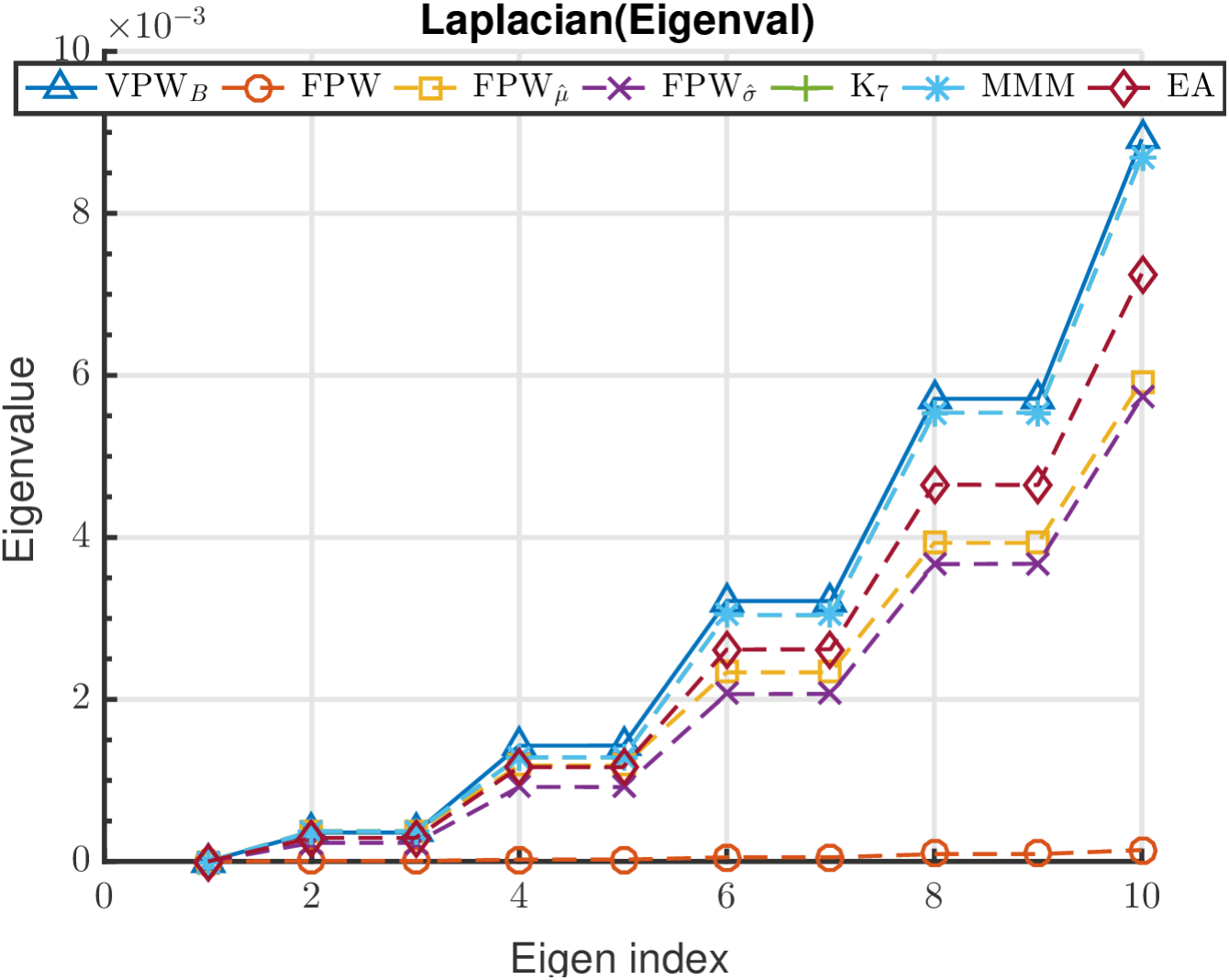}%
            \label{fig_toroidal_eigen}}
        \caption{Nonlinear dimensionality reduction (Toroidal helix)}
        \label{fig_toroidalhelix}
    \end{figure*}
    Data samples on a Toroidal helix is shown in Fig. \subref{fig_toroidal}. It contains $ 2095 $ data points embedded in $ \mathbb{R}^{3} $ and it is well known that the Toroidal helix is originally a 2D circle embedded in higher dimensional ambient space. Fig. \subref{fig_toroidalLap} to \subref{fig_toroidalVpw} show the low dimensional representation obtained using graph Laplacian with different Parzen window estimators. As evident, both FPW{$ _{\sigma} $} and VPW{$ _B $} are able to extract and preserve the true intrinsic geometry of Toroidal helix. Lower Parzen window values set in FPW and FPW{$ _{\hat{\mu}} $} leave unwanted curls in the final representation. The adaptive Parzen window estimators K\textsubscript{7}, MMM and EA are able to extract a smooth circle with a knot as shown in Fig. \subref{fig_toroidal_k7}, Fig. \subref{fig_toroidal_mmm}, and Fig. \subref{fig_toroidal_ea}, respectively.
    
    The results and eigenvalue comparison as shown in Fig. \subref{fig_toroidal_eigen} supports the fact that large eigenvalues include rich eigen-function counterparts in Laplace-Beltrami on the manifold which is why VPW{$ _B $} and FPW{$ _{\sigma} $} outperformed other methods.
\subsection{Brain computer interface}
\label{sec:brain_computer_interface}
    Brain computer interface (BCI) is an interface between electroencephalographic (EEG) signals from different imagery areas of mind and devices attached to its respective controller. 
    In this experiment, the raw EEG micro-volts signal has been used to train the LapRLSC model. The results show that proposed VPW{$ _C $} affinity outperforms other Parzen window estimators.
\subsubsection{HaLT data set}
\label{sec:halt}
    The large EEG motor imagery data set \cite{kaya_binli_ozbay_yanar_mishchenko_2018} contains five BCI paradigms experimental records, including HaLT. HaLT (Hand Leg Tongue) is an extension of the 3-state classic paradigm. It includes left leg, right leg, tongue, left hand, right hand, and passive imagery mental states. 
    In the data collection stage, each of the six movements was shown with an image on the computer screen for 1 second and respective 21 channels EEG readings were saved. Each such action consisted of approximately $ 170 $ frames of micro-volt data. Based on the action marker in the data, each such $ 170\times 21 $ frame was extracted and reshaped to $ 1\times 3570 $ vector. By appending all such frames, the final data set consisted of $ 2408\times 3570 $ matrix\footnote{Passive imagery readings were not included.}. 
    The training and test data set were created by randomly dividing each action data into two halves leading to a $ 10 $ binary classification model.\\
    \begin{figure}[!h]
        \centering
        \subfloat[LapRLSC (unlabeled set)]{\includegraphics[width=0.25\linewidth,clip,keepaspectratio]{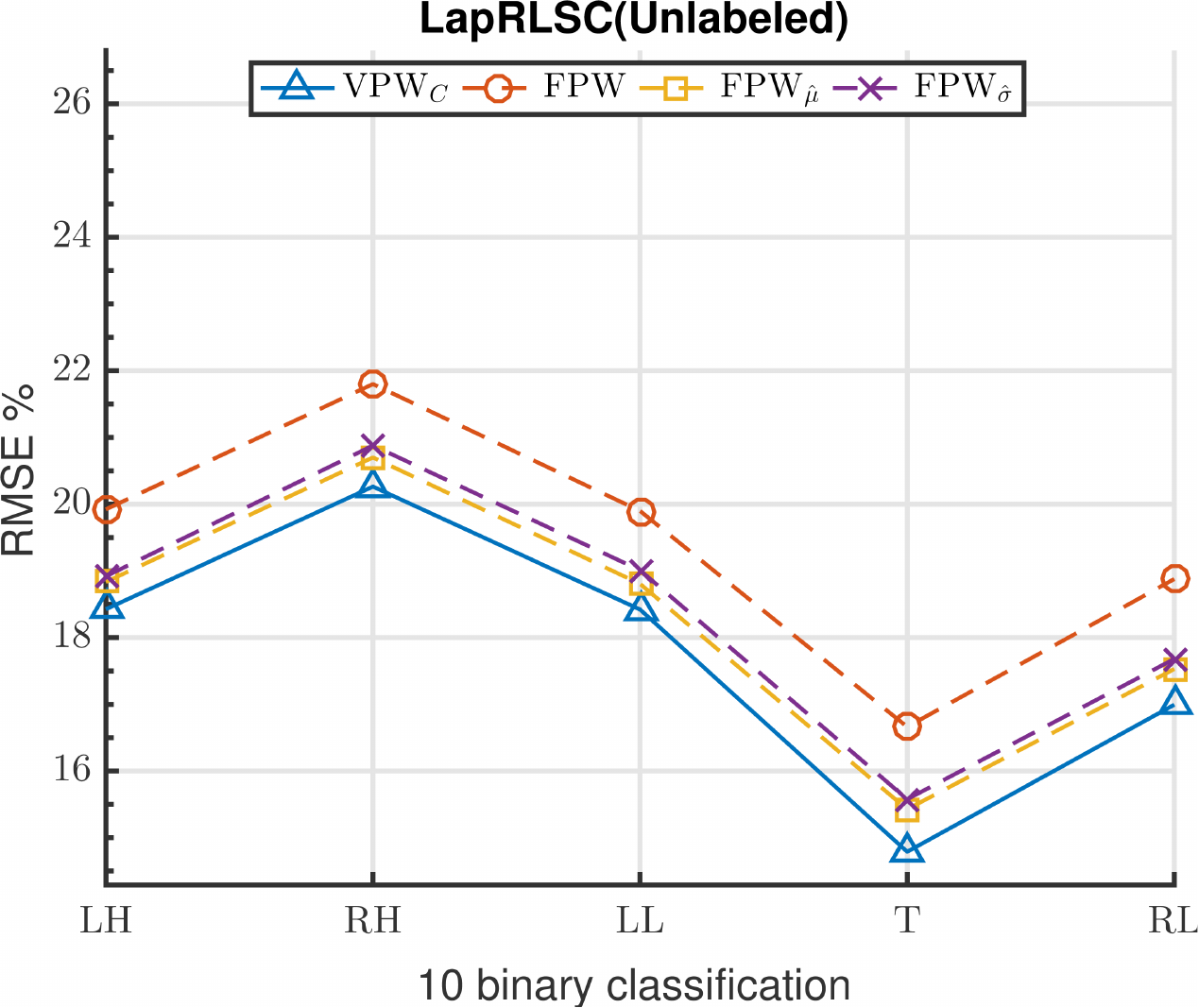}%
            \label{fig_bci_LapRlsc_u}}
        \hfil
        \subfloat[LapRLSC (test set)]{\includegraphics[width=0.25\linewidth,clip,keepaspectratio]{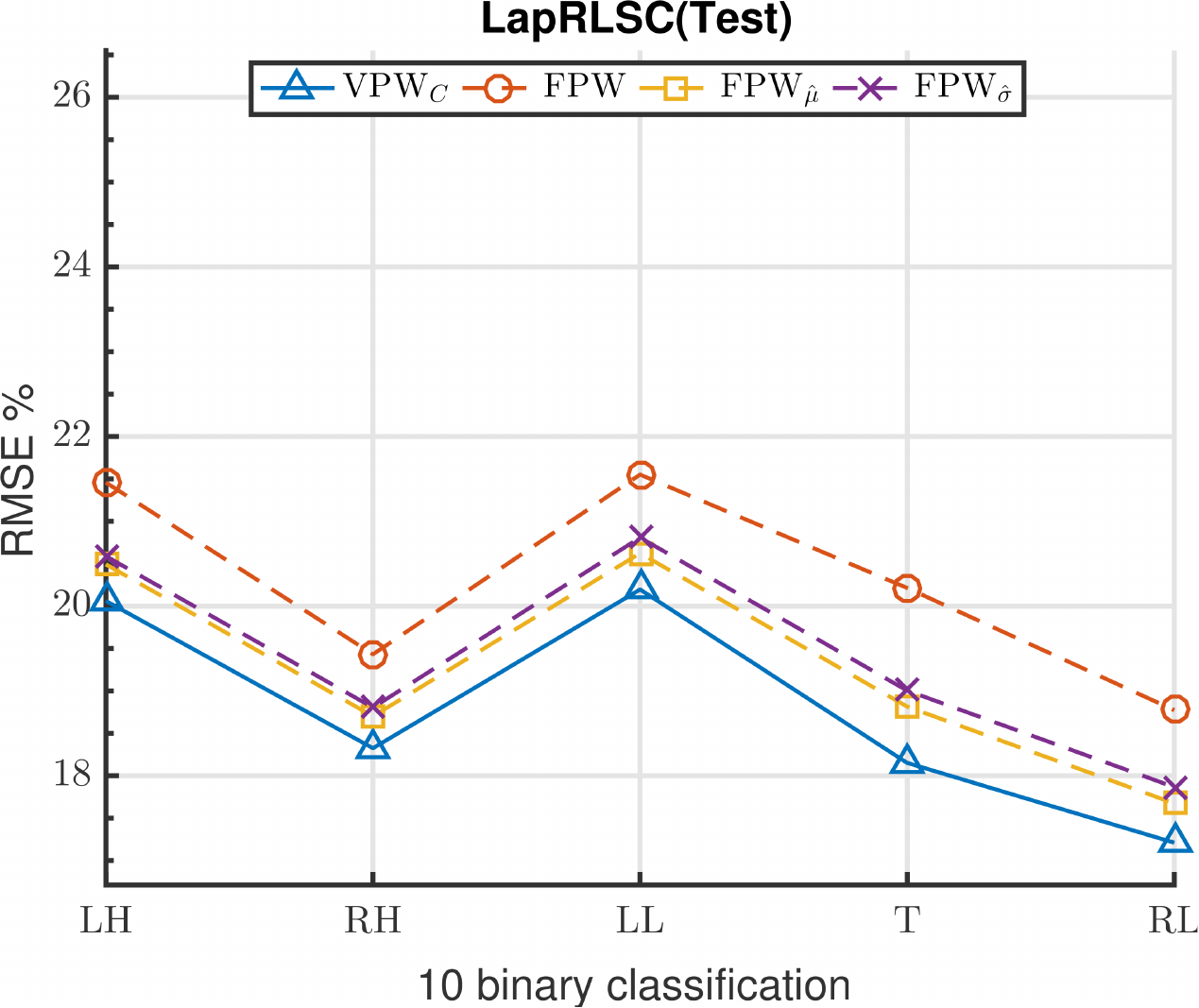}%
            \label{fig_bci_LapRlsc_t}}
        \hfil
        \subfloat[Eigenvalue]{\includegraphics[width=0.25\linewidth,clip,keepaspectratio]{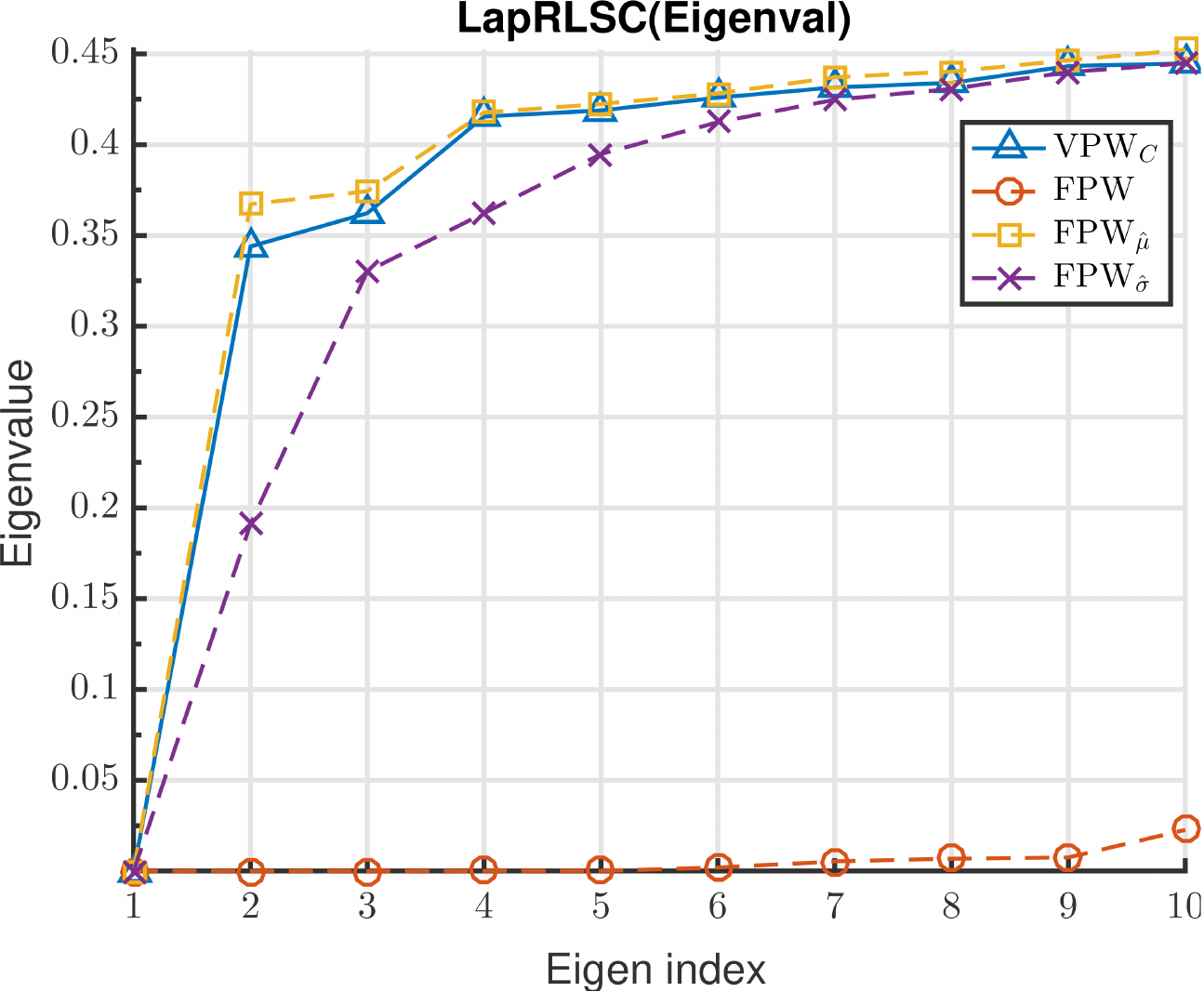}%
            \label{fig_bci_ev}}
        \caption{LapRLSC mean error and eigenvalue comparison between FPW, FPW$ _{\hat{\mu}} $, FPW$ _{\sigma} $, and VPW$ _{C} $ on HaLT data set}
        \label{fig_bci}
    \end{figure}
    \begin{table}[!h]
        \caption{HaLT mean error (standard deviation) with varying $ |N| $}
        \label{table:bci}
        \centering
        \small\addtolength{\tabcolsep}{-5pt}
        \begin{singlespace}\scriptsize
        \begin{tabular}{l*{10}{m{1cm}}}
            \hline\hline
            \multirow{2}{*}{\bfseries Affinity} &              \multicolumn{2}{c}{$ |N| $=7}              &              \multicolumn{2}{c}{$ |N| $=8}              &              \multicolumn{2}{c}{$ |N| $=9}              &             \multicolumn{2}{c}{$ |N| $=10}              &             \multicolumn{2}{c}{$ |N| $=11}              \\
            &                    \thead{et} &                    \thead{eu} &                    \thead{et} &                    \thead{eu} &                    \thead{et} &                    \thead{eu} &                    \thead{et} &                    \thead{eu} &                    \thead{et} &                    \thead{eu} \\ \hline
            FPW                                 &          $ 20.07 $ ($ 1.48 $) &          $ 19.12 $ ($ 1.88 $) &          $ 19.96 $ ($ 0.93 $) &          $ 19.57 $ ($ 2.01 $) &          $ 20.13 $ ($ 1.80 $) &          $ 18.97 $ ($ 2.19 $) &          $ 19.72 $ ($ 0.35 $) &          $ 19.65 $ ($ 1.82 $) &          $ 20.18 $ ($ 0.75 $) &          $ 19.63 $ ($ 2.01 $) \\
            FPW$_{\hat{\mu}}$                   &          $ 18.82 $ ($ 0.98 $) &          $ 18.07 $ ($ 1.99 $) &          $ 18.84 $ ($ 1.37 $) &          $ 18.20 $ ($ 2.22 $) &          $ 19.23 $ ($ 1.00 $) &          $ 18.10 $ ($ 2.32 $) &          $ 18.73 $ ($ 1.22 $) &          $ 18.43 $ ($ 2.29 $) &          $ 18.77 $ ($ 1.30 $) &          $ 17.88 $ ($ 1.87 $) \\
            FPW$_{\sigma}$                &          $ 19.43 $ ($ 1.27 $) &          $ 18.52 $ ($ 2.40 $) &          $ 19.06 $ ($ 1.40 $) &          $ 17.97 $ ($ 1.86 $) &          $ 19.00 $ ($ 1.58 $) &          $ 18.61 $ ($ 2.25 $) &          $ 19.07 $ ($ 1.37 $) &          $ 18.35 $ ($ 1.81 $) &          $ 19.04 $ ($ 1.32 $) &          $ 18.04 $ ($ 2.00 $) \\
            K$_{7}$                             &          $ 18.75 $ ($ 1.30 $) &          $ 17.73 $ ($ 2.22 $) &          $ 18.93 $ ($ 0.97 $) &          $ 17.75 $ ($ 2.36 $) &          $ 18.13 $ ($ 1.89 $) &          $ 18.10 $ ($ 1.86 $) &          $ 18.99 $ ($ 1.13 $) &          $ 18.24 $ ($ 2.66 $) &          $ 18.42 $ ($ 1.88 $) &          $ 17.94 $ ($ 1.91 $) \\
            MMM                                 &          $ 19.07 $ ($ 1.72 $) &          $ 18.11 $ ($ 1.89 $) &          $ 19.06 $ ($ 1.13 $) &          $ 18.33 $ ($ 2.09 $) &          $ 18.70 $ ($ 1.41 $) &          $ 18.06 $ ($ 2.41 $) &          $ 18.32 $ ($ 1.60 $) &          $ 18.22 $ ($ 1.88 $) &          $ 18.85 $ ($ 1.20 $) &          $ 18.26 $ ($ 1.71 $) \\
            EA & $ 19.69 $ ($ 1.26 $) & $ 18.84 $ ($ 2.06 $) & $ 19.69 $ ($ 1.26 $) & $ 18.84 $ ($ 2.06 $) & $ 19.69 $ ($ 1.26 $) & $ 18.84 $ ($ 2.06 $) & $ 19.69 $ ($ 1.26 $) & $ 18.84 $ ($ 2.06 $) & $ 19.69 $ ($ 1.26 $) & $ 18.84 $ ($ 2.06 $) \\
            VPW$_{C}$                           & $ \mathbf{16.83} $ ($ 1.72 $) & $ \mathbf{16.00} $ ($ 3.41 $) & $ \mathbf{16.45} $ ($ 1.32 $) & $ \mathbf{16.47} $ ($ 2.95 $) & $ \mathbf{17.90} $ ($ 1.22 $) & $ \mathbf{16.52} $ ($ 2.78 $) & $ \mathbf{17.18} $ ($ 1.44 $) & $ \mathbf{16.45} $ ($ 2.06 $) & $ \mathbf{17.25} $ ($ 1.40 $) & $ \mathbf{15.54} $ ($ 2.30 $) \\ \hline
        \end{tabular}
        \end{singlespace}
    \end{table}
    Each binary classifier was executed $ 20 $ times with $ 12 $ randomly labeled samples each for $ \{+1,-1\} $ classes. Fig. \ref{fig_bci} shows the results of the BCI classification for both unlabeled and test sets. The X-axis shows the actions performed LH, RH, LL, T and RL representing left hand, right hand, left leg, tongue and right leg respectively and Y-axis shows the classification error. The task vs. classification error shows that VPW$ _{C} $ outperforms other estimators in both unlabeled and test set by computing optimal affinity between the data points. Similar to the previous experiment, the performance of estimators is directly linked with their eigenvalues as shown in Fig. \subref{fig_bci_ev}. It contains $ 10 $ smallest eigenvalues of all four estimators and their accuracy is in the same order i.e. VPW$ _{C} $ gave best results followed by FPW$ _{\hat{\mu}} $, FPW$ _{\sigma} $ and FPW. It proves that estimator with large and smooth eigenvalues results in better manifold regularization. The results also validate that proposed affinity based SSL works accurately even with sparse labels and raw EEG micro-volt data.

    As the adaptive Parzen window estimators change with neighborhood properties; the comparative study between them and VPW$ _{C} $ is performed by varying the $ |N| $ value to create the neighborhoods. The result has been listed in Table \ref{table:bci}. It shows that VPW$ _{C} $ consistently gave accurate results across all $ |N| $ values. Due to change in the neighborhood, it can not be ensured that the local neighborhood always contains data points that fall in a small region $ R $. Hence, the increase in $ |N| $ does not always increase the model's accuracy.
\subsection{Handwritten digit recognition}
\label{sec:handwritten_digit_recognition}
    Handwritten digit recognition has always been treated as a benchmark data set to evaluate any classification model. Due to the inherent high variance in samples, it has always proved to be a challenging task for classification. Here, we have validated the performance of VPW on three benchmark handwritten data set: Hasy, USPS, and MNIST.
\subsubsection{Hasy v2}
\label{sec:hasy}
    \begin{figure}[!h]
        \centering
        \subfloat[LapRLSC (unlabeled set)]{\includegraphics[width=0.25\linewidth,clip,keepaspectratio]{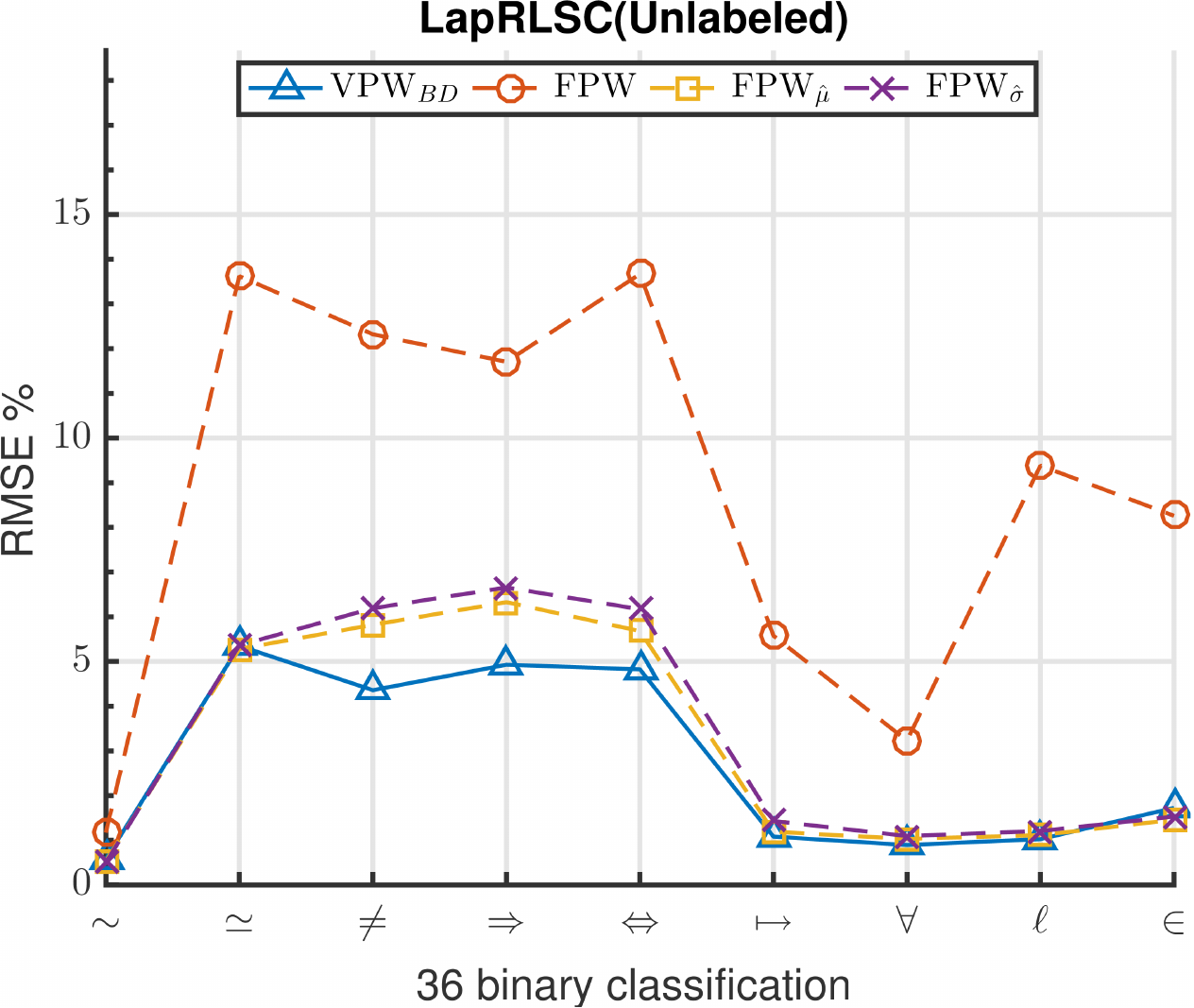}%
            \label{fig_hasy_LapRlsc_u}}
        \hfil
        \subfloat[LapRLSC (test set)]{\includegraphics[width=0.25\linewidth,clip,keepaspectratio]{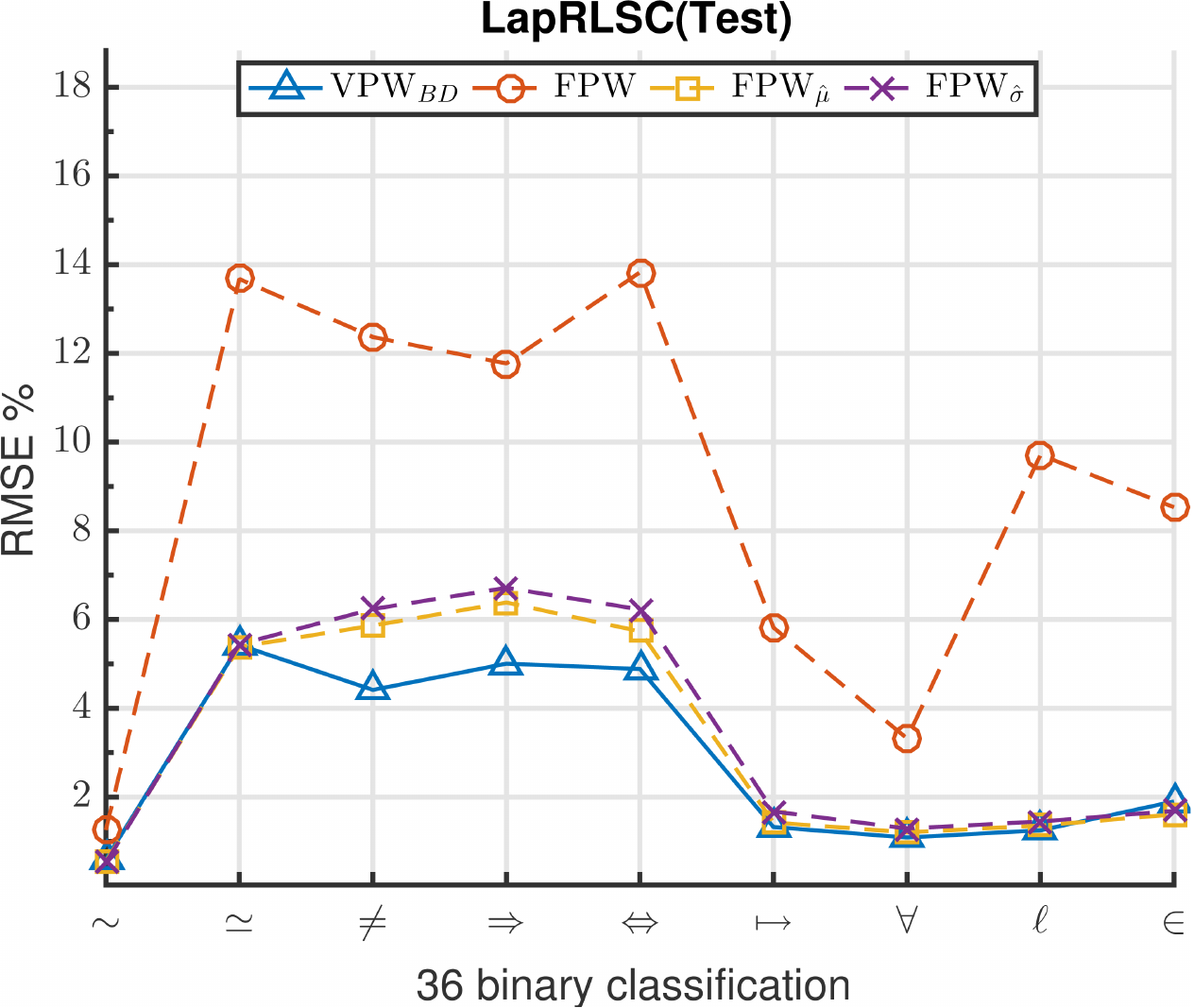}%
            \label{fig_hasy_LapRlsc_t}}
        \hfil
        \subfloat[Eigenvalue]{\includegraphics[width=0.25\linewidth,clip,keepaspectratio]{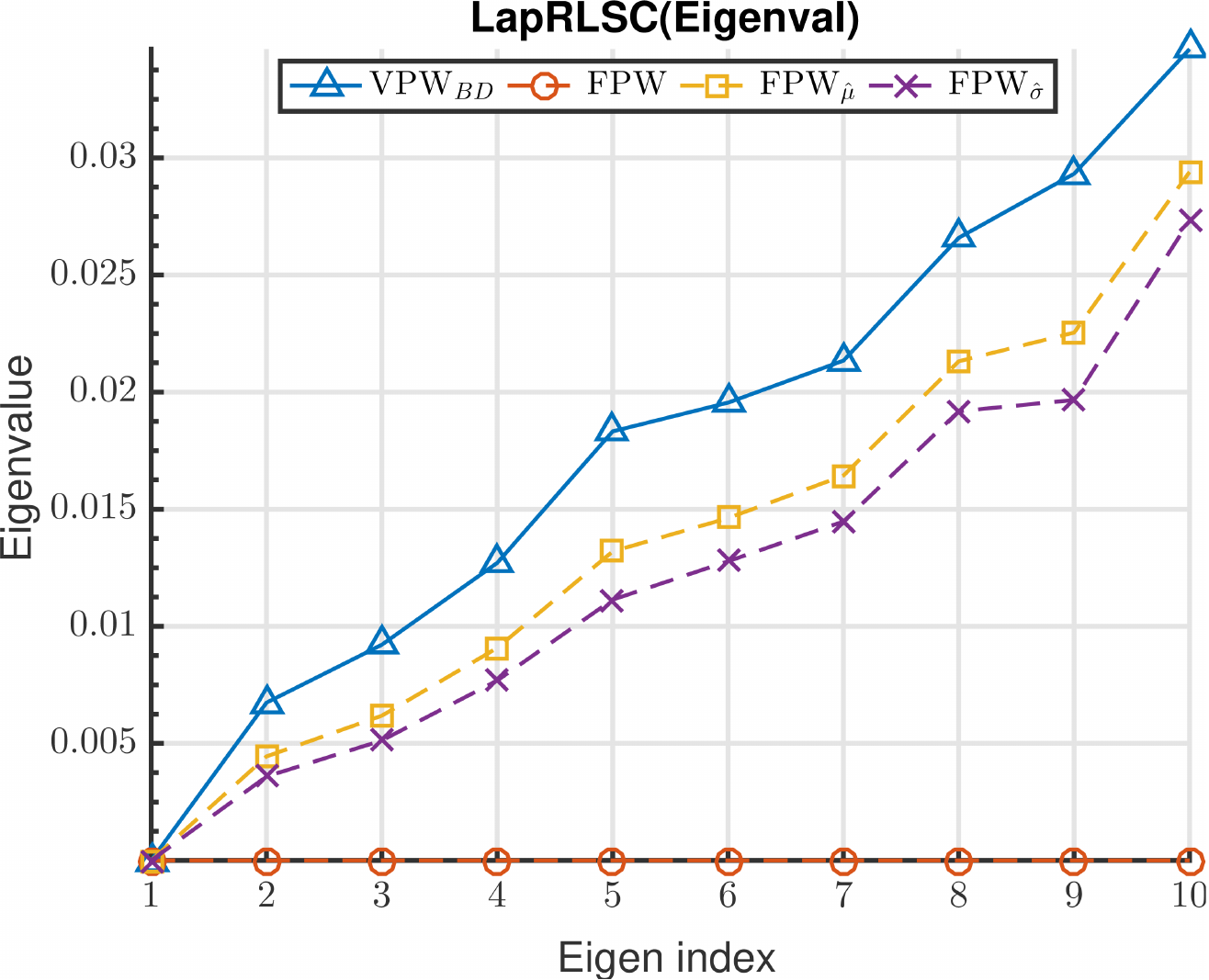}%
            \label{fig_hasy_ev}}
        \caption{LapRLSC mean error and eigenvalue comparison between FPW, FPW$ _{\hat{\mu}} $, FPW$ _{\sigma} $, and VPW$ _{BD} $ on Hasy\_v2 symbol data set}
        \label{fig_hasy}
    \end{figure}
    \begin{table}[!h]
        \caption{Hasy v2 mean error (standard deviation) with varying $ |N| $}
        \label{table:hasy}
        \centering
        \small\addtolength{\tabcolsep}{-5pt}
        \begin{singlespace}\scriptsize
        \begin{tabular}{l*{10}{m{1cm}}}
            \hline\hline
            \multirow{2}{*}{\bfseries Affinity} &            \multicolumn{2}{c}{$ |N| $=21}             &            \multicolumn{2}{c}{$ |N| $=22}             &            \multicolumn{2}{c}{$ |N| $=23}             &            \multicolumn{2}{c}{$ |N| $=24}             &            \multicolumn{2}{c}{$ |N| $=25}             \\
                                                &                   \thead{et} &                   \thead{eu} &                   \thead{et} &                   \thead{eu} &                   \thead{et} &                   \thead{eu} &                   \thead{et} &                   \thead{eu} &                   \thead{et} &                   \thead{eu} \\ \hline
            FPW                                 &          $ 3.45 $ ($ 1.77 $) &          $ 3.36 $ ($ 1.80 $) &          $ 3.45 $ ($ 1.77 $) &          $ 3.36 $ ($ 1.80 $) &          $ 3.45 $ ($ 1.77 $) &          $ 3.36 $ ($ 1.80 $) &          $ 3.45 $ ($ 1.77 $) &          $ 3.36 $ ($ 1.80 $) &          $ 3.45 $ ($ 1.77 $) &          $ 3.36 $ ($ 1.80 $) \\
            FPW$_{\hat{\mu}}$                   &          $ 2.19 $ ($ 1.23 $) &          $ 2.17 $ ($ 1.27 $) &          $ 2.22 $ ($ 1.25 $) &          $ 2.20 $ ($ 1.28 $) &          $ 2.24 $ ($ 1.26 $) &          $ 2.16 $ ($ 1.33 $) &          $ 2.26 $ ($ 1.27 $) &          $ 2.17 $ ($ 1.35 $) &          $ 2.27 $ ($ 1.28 $) &          $ 2.19 $ ($ 1.36 $) \\
            FPW$_{\sigma}$                &          $ 2.24 $ ($ 1.25 $) &          $ 2.16 $ ($ 1.31 $) &          $ 2.28 $ ($ 1.26 $) &          $ 2.20 $ ($ 1.33 $) &          $ 2.30 $ ($ 1.27 $) &          $ 2.23 $ ($ 1.34 $) &          $ 2.32 $ ($ 1.29 $) &          $ 2.25 $ ($ 1.36 $) &          $ 2.34 $ ($ 1.29 $) &          $ 2.26 $ ($ 1.36 $) \\
            K$_{7}$                             &          $ 2.67 $ ($ 1.31 $) &          $ 2.53 $ ($ 1.47 $) &          $ 2.50 $ ($ 1.19 $) &          $ 2.38 $ ($ 1.36 $) &          $ 2.64 $ ($ 1.14 $) &          $ 2.59 $ ($ 1.24 $) &          $ 2.74 $ ($ 1.14 $) &          $ 2.68 $ ($ 1.23 $) &          $ 2.73 $ ($ 1.19 $) &          $ 2.75 $ ($ 1.23 $) \\
            MMM                                 &          $ 2.30 $ ($ 1.38 $) &          $ 2.21 $ ($ 1.46 $) &          $ 2.60 $ ($ 1.59 $) &          $ 2.50 $ ($ 1.66 $) &          $ 2.59 $ ($ 1.66 $) &          $ 2.50 $ ($ 1.74 $) &          $ 2.63 $ ($ 1.69 $) &          $ 2.54 $ ($ 1.77 $) &          $ 2.77 $ ($ 1.73 $) &          $ 2.60 $ ($ 1.85 $) \\
            EA & $ 2.28 $ ($ 1.40 $) & $ 2.16 $ ($ 1.47 $) & $ 2.31 $ ($ 1.42 $) & $ 2.19 $ ($ 1.48 $) & $ 2.33 $ ($ 1.43 $) & $ 2.22 $ ($ 1.50 $) & $ 2.36 $ ($ 1.44 $) & $ 2.25 $ ($ 1.51 $) & $ 2.39 $ ($ 1.46 $) & $ 2.28 $ ($ 1.53 $) \\
            VPW$_{BD}$                           & $ \mathbf{1.88} $ ($ 1.04 $) & $ \mathbf{1.80} $ ($ 1.07 $) & $ \mathbf{1.89} $ ($ 1.03 $) & $ \mathbf{1.81} $ ($ 1.07 $) & $ \mathbf{1.89} $ ($ 1.03 $) & $ \mathbf{1.81} $ ($ 1.07 $) & $ \mathbf{1.93} $ ($ 1.05 $) & $ \mathbf{1.85} $ ($ 1.10 $) & $ \mathbf{1.98} $ ($ 1.11 $) & $ \mathbf{1.91} $ ($ 1.15 $) \\ \hline
        \end{tabular}
        \end{singlespace}
    \end{table}
    The publicly available Hasy\_v2 data set \cite{DBLP:journals/corr/Thoma17} similar to the MNIST data set contains $ 168233 $ single symbols across $ 369 $ classes. 
    Here, each image consists of $ 32\times 32 $ black and white pixels. Since, many symbol categories included less than $ 51 $ samples, hence, to avoid data imbalance, in this experiment we used $ 9 $ symbols which contained more than $ 800 $ images individually. The training and testing data set were created by dividing the number of images in each symbol in two halves. Complete classification model consisted of $ 36 $ binary LapRLSC classifiers where each model was trained $ 20 $ times using $ 2 $ random images labeled in both $ +1 $ and $ -1 $ class. Fig. \ref{fig_hasy} shows the unlabeled and test result comparison between FPW and VPW$ _{BD} $ Parzen window estimators. In unlabeled set as shown in Fig. \subref{fig_hasy_LapRlsc_u}, except for symbols $ \ell $ and $ \in $, VPW$ _{BD} $ outperformed the existing FPW estimators. As for $ \sim $ symbol, VPW$ _{BD} $ based model gave $ \approx 100\% $ accuracy.
    
    The eigenvalue analysis is shown in Fig. \subref{fig_hasy_ev} confirms that higher eigenvalue results in better manifold regularization. The adaptive Parzen window estimators have also been compared with VPW$ _{BD} $ based on the change in $ |N| $ value as listed in Table \ref{table:hasy}. It shows that as the $ |N| $ increases, the accuracy of the model dip due to unwanted cross category edge connections and while other estimators hogged around $ \approx 97\% $ accuracy, VPW$ _{BD} $ outperformed them by increasing the underlying model's accuracy to $ \approx 99\% $.
\subsubsection{USPS}
\label{sec:usps}
    \begin{figure}[!h]
        \centering
        \subfloat[LapRLSC (unlabeled set)]{\includegraphics[width=0.25\linewidth,clip,keepaspectratio]{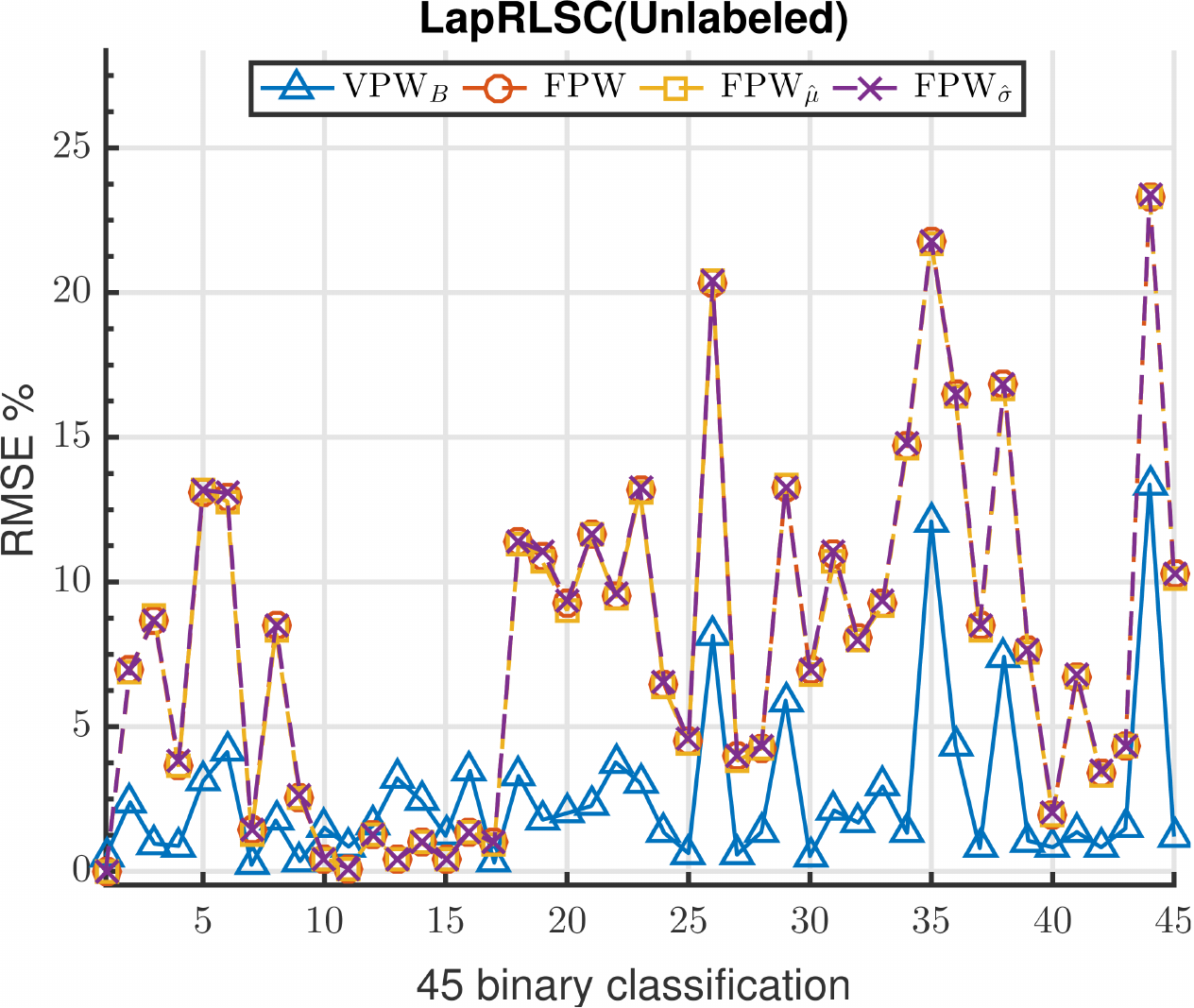}%
            \label{fig_45ClassificationProblem_LapRLSC_u}}
        \hfil
        \subfloat[LapRLSC (test set)]{\includegraphics[width=0.25\linewidth,clip,keepaspectratio]{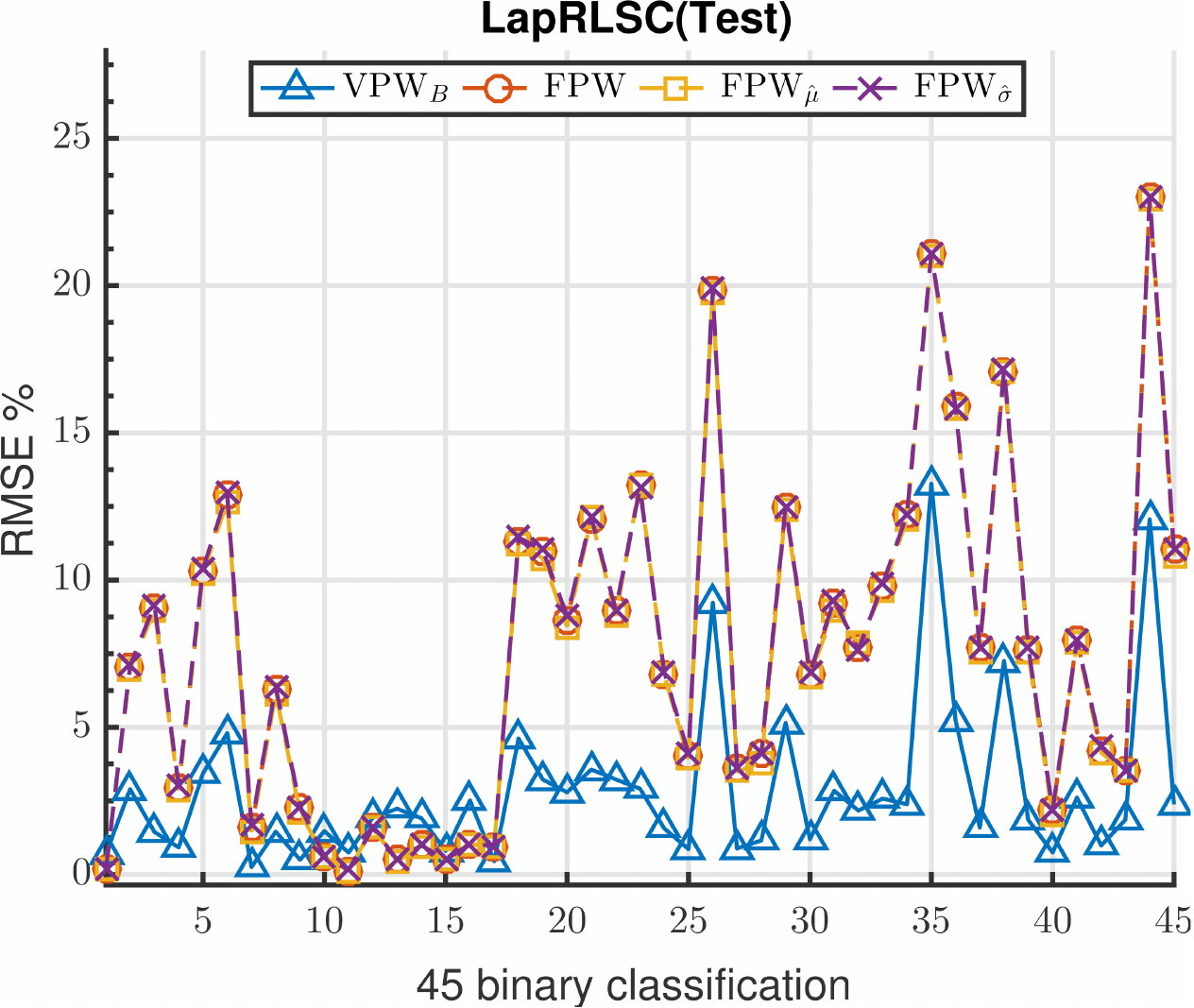}%
            \label{fig_45ClassificationProblem_LapRLSC_t}}
        \hfil
        \subfloat[Eigenvalue]{\includegraphics[width=0.25\linewidth,clip,keepaspectratio]{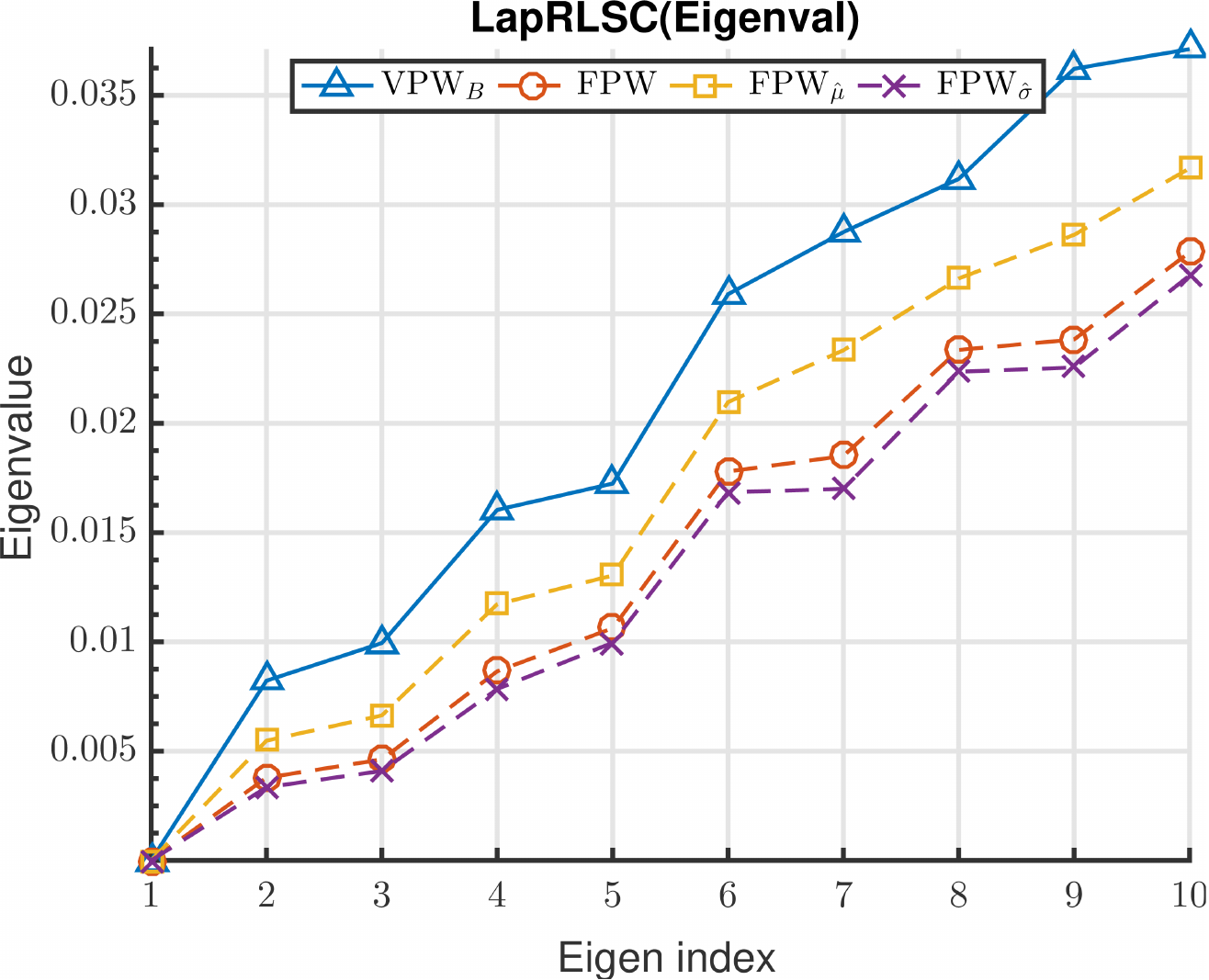}%
            \label{fig_usps_ev}}
        \caption{LapRLSC mean error and eigenvalue comparison between FPW, FPW$ _{\hat{\mu}} $, FPW$ _{\sigma} $, and VPW$ _{B} $ on USPS handwritten digit recognition}
        \label{fig_45ClassificationProblem}
    \end{figure}
    \begin{table}[!h]
        \caption{USPS mean error (standard deviation) with varying $ |N| $}
        \label{table:usps}
        \centering
        \small\addtolength{\tabcolsep}{-5pt}
        \begin{singlespace}\scriptsize
        \begin{tabular}{l*{10}{m{1cm}}}
            \hline\hline
            \multirow{2}{*}{\bfseries Affinity} &             \multicolumn{2}{c}{$ |N| $=7}             &             \multicolumn{2}{c}{$ |N| $=8}             &             \multicolumn{2}{c}{$ |N| $=9}             &            \multicolumn{2}{c}{$ |N| $=10}             &            \multicolumn{2}{c}{$ |N| $=11}             \\
                                                &                   \thead{et} &                   \thead{eu} &                   \thead{et} &                   \thead{eu} &                   \thead{et} &                   \thead{eu} &                   \thead{et} &                   \thead{eu} &                   \thead{et} &                   \thead{eu} \\ \hline
            FPW                                 &          $ 2.70 $ ($ 2.66 $) &          $ 2.25 $ ($ 2.55 $) &          $ 2.57 $ ($ 2.73 $) &          $ 2.12 $ ($ 2.54 $) &          $ 2.41 $ ($ 2.64 $) &          $ 2.06 $ ($ 2.54 $) &          $ 2.39 $ ($ 2.66 $) &          $ 2.05 $ ($ 2.53 $) &          $ 2.41 $ ($ 2.83 $) &          $ 2.01 $ ($ 2.59 $) \\
            FPW$_{\hat{\mu}}$                   &          $ 2.62 $ ($ 2.66 $) &          $ 2.28 $ ($ 2.62 $) &          $ 2.45 $ ($ 2.69 $) &          $ 2.14 $ ($ 2.60 $) &          $ 2.40 $ ($ 2.73 $) &          $ 2.12 $ ($ 2.61 $) &          $ 2.34 $ ($ 2.71 $) &          $ 2.10 $ ($ 2.62 $) &          $ 2.28 $ ($ 2.75 $) &          $ 2.08 $ ($ 2.68 $) \\
            FPW$_{\sigma}$                &          $ 2.71 $ ($ 2.61 $) &          $ 2.31 $ ($ 2.55 $) &          $ 2.58 $ ($ 2.63 $) &          $ 2.18 $ ($ 2.64 $) &          $ 2.53 $ ($ 2.82 $) &          $ 2.15 $ ($ 2.63 $) &          $ 2.50 $ ($ 2.85 $) &          $ 2.09 $ ($ 2.56 $) &          $ 2.37 $ ($ 2.71 $) &          $ 2.09 $ ($ 2.72 $) \\
            K$_{7}$                             &          $ 3.51 $ ($ 2.94 $) &          $ 3.43 $ ($ 3.18 $) &          $ 3.29 $ ($ 2.91 $) &          $ 3.21 $ ($ 3.05 $) &          $ 3.23 $ ($ 2.94 $) &          $ 3.17 $ ($ 3.01 $) &          $ 3.28 $ ($ 3.01 $) &          $ 3.25 $ ($ 3.09 $) &          $ 3.18 $ ($ 2.98 $) &          $ 3.17 $ ($ 3.06 $) \\
            MMM                                 &          $ 3.74 $ ($ 2.70 $) &          $ 3.42 $ ($ 2.85 $) &          $ 3.45 $ ($ 2.74 $) &          $ 3.15 $ ($ 2.77 $) &          $ 3.30 $ ($ 2.83 $) &          $ 3.06 $ ($ 2.86 $) &          $ 3.19 $ ($ 2.83 $) &          $ 2.95 $ ($ 2.82 $) &          $ 3.03 $ ($ 2.86 $) &          $ 2.86 $ ($ 2.84 $) \\
            EA & $ 3.93 $ ($ 4.72 $) & $ 4.72 $ ($ 4.31 $) & $ 4.68 $ ($ 5.42 $) & $ 4.99 $ ($ 4.94 $) & $ 4.83 $ ($ 6.06 $) & $ 5.58 $ ($ 5.58 $) & $ 5.16 $ ($ 6.56 $) & $ 6.07 $ ($ 6.42 $) & $ 5.73 $ ($ 6.87 $) & $ 6.69 $ ($ 6.99 $) \\
            VPW$_{B}$                           & $ \mathbf{1.80} $ ($ 2.71 $) & $ \mathbf{1.56} $ ($ 2.65 $) & $ \mathbf{1.62} $ ($ 2.62 $) & $ \mathbf{1.36} $ ($ 2.80 $) & $ \mathbf{1.43} $ ($ 2.67 $) & $ \mathbf{1.40} $ ($ 2.72 $) & $ \mathbf{1.47} $ ($ 2.83 $) & $ \mathbf{1.37} $ ($ 2.67 $) & $ \mathbf{1.40} $ ($ 2.68 $) & $ \mathbf{1.21} $ ($ 2.87 $) \\ \hline
        \end{tabular}
        \end{singlespace}
    \end{table}
    The USPS data set consists of handwritten digits $ 0-9 $. In pre-processing, each sample image is reduced to $ 100 $ dimensions using PCA, which constituted $ >90\% $ of total data variance. First $ 400 $ images from each digit were included in the training set and rest in the testing set. The experiment consisted of $ 45 $ binary LapRLSC classifiers. Fig. \ref{fig_45ClassificationProblem} shows the result of unlabeled and test set. The prominent error rate spikes at $ 18, 26, 29, 35, 38 \text{ and }44 $ in the unlabeled model of original FPW were significantly reduced using VPW$ _{B} $ as shown in Fig. \subref{fig_45ClassificationProblem_LapRLSC_u}). VPW$ _{B} $ also outperformed FPW estimators in the testing set as shown in Fig. \subref{fig_45ClassificationProblem_LapRLSC_t} except for $ 13^{th}\text{ and }14^{th} $ comparison where FPW gave more accurate results than VPW$ _{B} $.
    
    The ranking of estimators' performance can be interpreted from the eigenvalue study shown in Fig. \ref{fig_usps_ev}. High eigenvalues resulted in better manifold regularization and hence, increasing the underlying model's accuracy. The effect of varying $ |N| $ on mean error has been illustrated in Table \ref{table:usps}. The overall result trend shows that increase in $ |N| $ reduces the classification error. It also shows that VPW$ _{B} $ gives more accurate results as compared to other adaptive Parzen window estimators across all $ |N| $ values. The default graph sparsity $ |N| $ and perplexity $ |N|-1 $ on USPS for EA resulted in a very biased affinity and hence, the accuracy of the model dipped. However, when the perplexity was tuned to $ |N|\times 20 $ and fully connected graph was created, the accuracy of the model increased which has been listed in the Table \ref{table:usps}.
\subsubsection{MNIST}
\label{sec:mnist}
    \begin{figure}[!h]
        \centering
        \subfloat[LapRLSC (unlabeled set)]{\includegraphics[width=0.25\linewidth,clip,keepaspectratio]{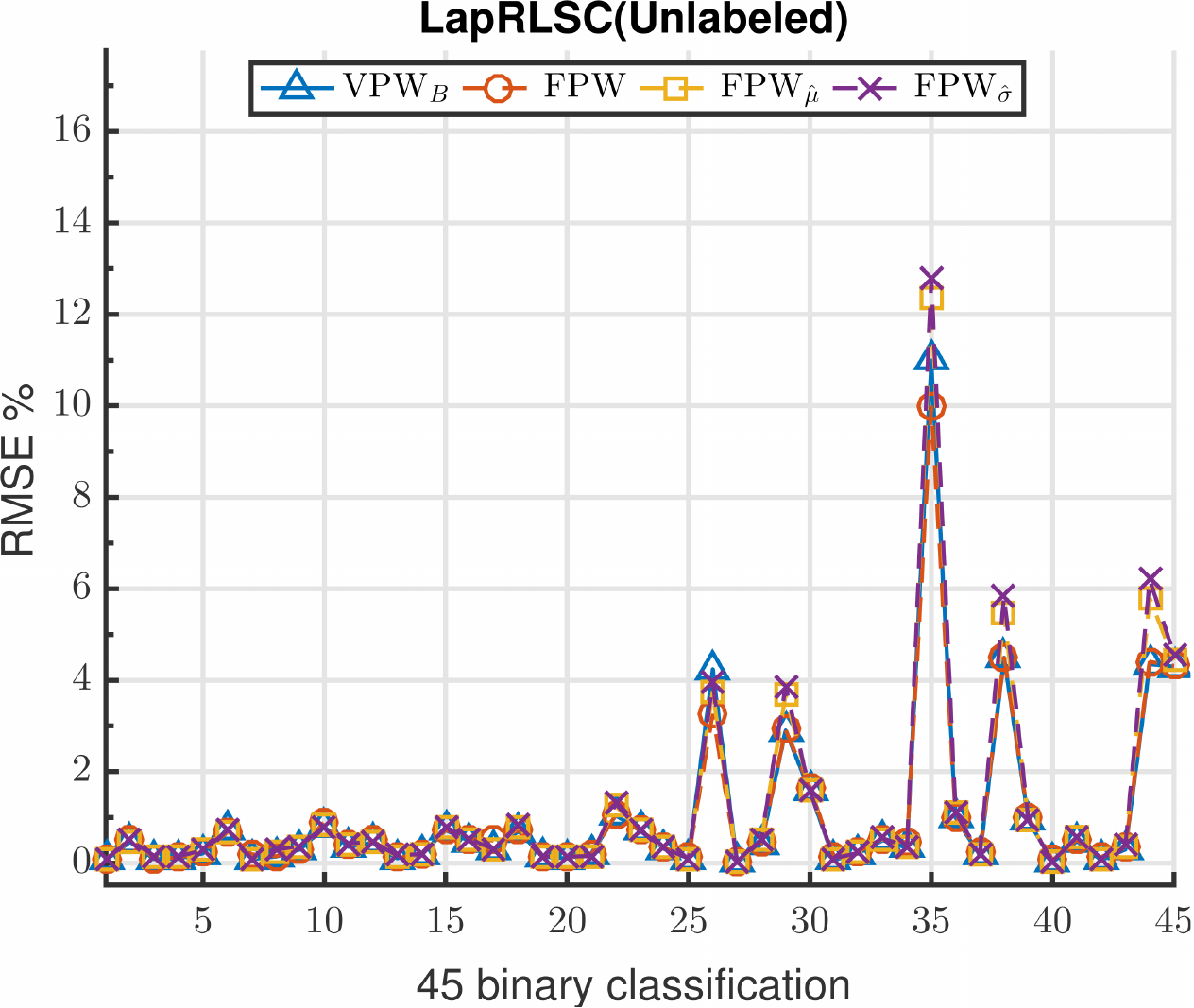}%
            \label{fig_mnist_laprlsc_u}}
        \hfil
        \subfloat[LapRLSC (test set)]{\includegraphics[width=0.25\linewidth,clip,keepaspectratio]{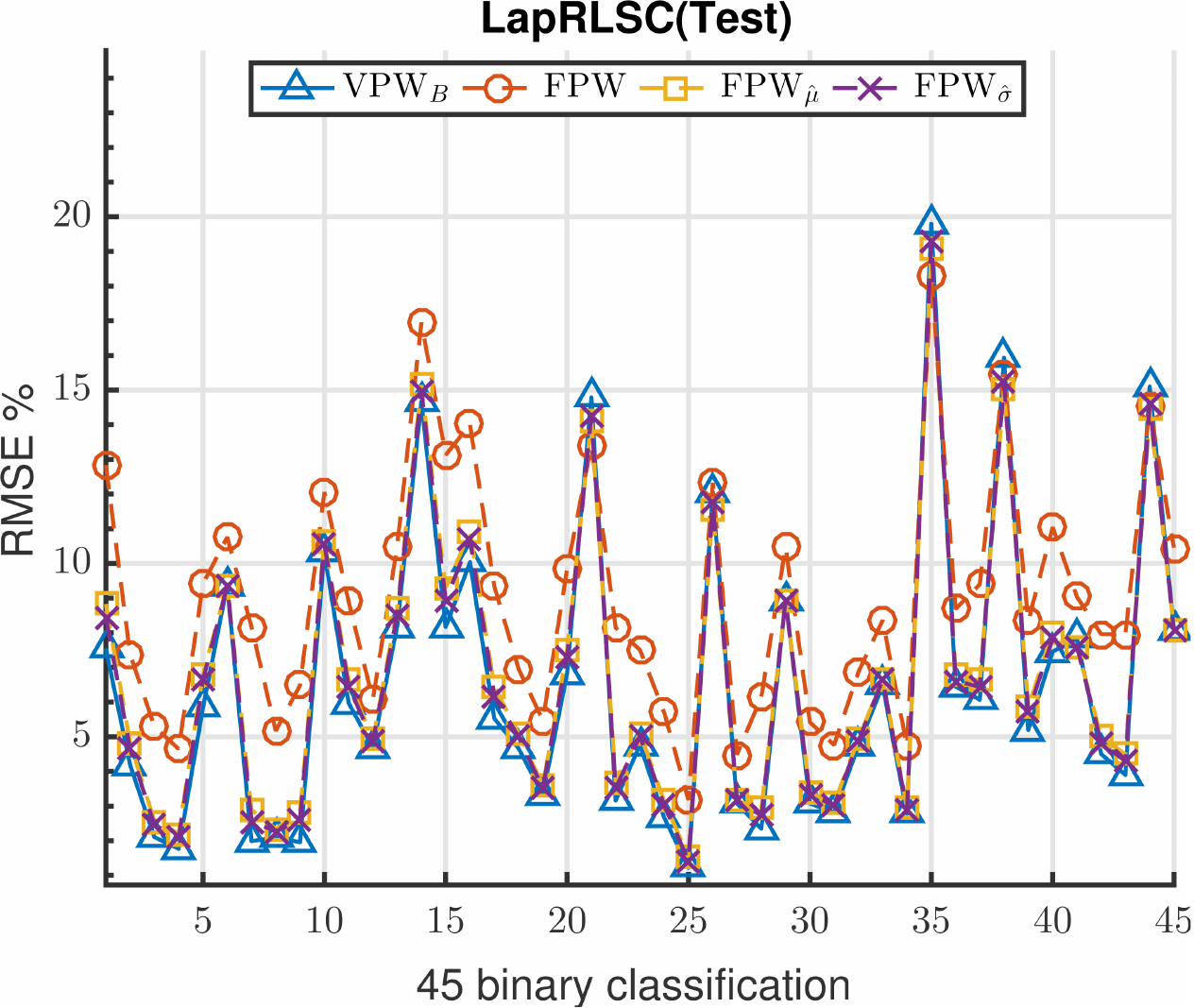}%
            \label{fig_mnist_laprlsc_t}}
        \hfil
        \subfloat[Eigenvalue]{\includegraphics[width=0.25\linewidth,clip,keepaspectratio]{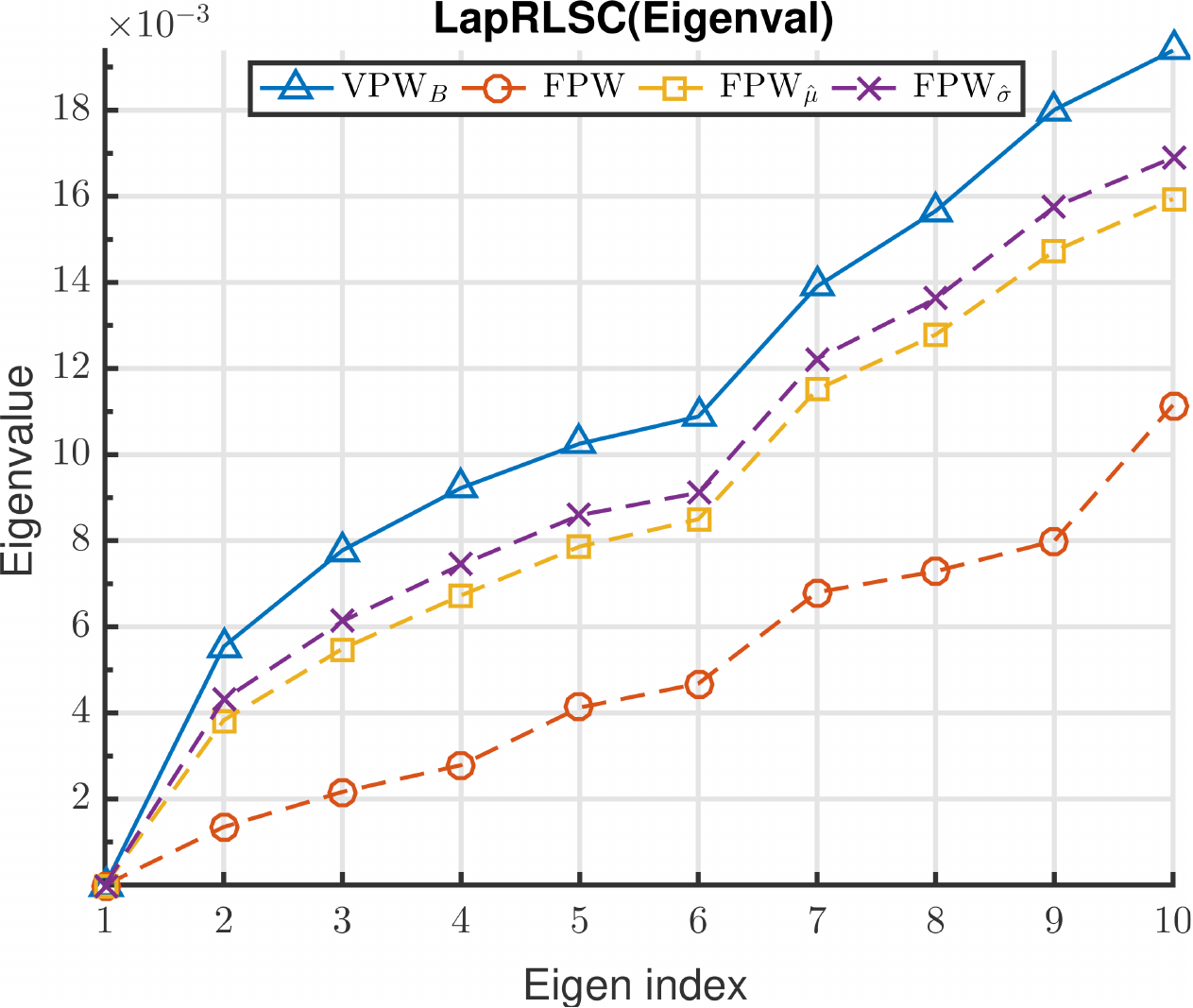}%
            \label{fig_mnist_ev}}
        \caption{LapRLSC mean error and eigenvalue comparison between FPW, FPW$ _{\hat{\mu}} $, FPW$ _{\sigma} $, and VPW$ _{B} $ on MNIST digit data set}
        \label{fig_mnist}
    \end{figure}
    \begin{table}[!h]
        \caption{MNIST mean error (standard deviation) with varying $ |N| $}
        \label{table:mnist}
        \centering
        \small\addtolength{\tabcolsep}{-5pt}
        \begin{singlespace}\scriptsize
        \begin{tabular}{l*{10}{m{1cm}}}
            \hline\hline
            \multirow{2}{*}{\bfseries Affinity} &             \multicolumn{2}{c}{$ |N| $=7}             &             \multicolumn{2}{c}{$ |N| $=8}             &             \multicolumn{2}{c}{$ |N| $=9}             &            \multicolumn{2}{c}{$ |N| $=10}             &            \multicolumn{2}{c}{$ |N| $=11}             \\
                                                &                   \thead{et} &                   \thead{eu} &                   \thead{et} &                   \thead{eu} &                   \thead{et} &                   \thead{eu} &                   \thead{et} &                   \thead{eu} &                   \thead{et} &                   \thead{eu} \\ \hline
            FPW                                 &          $ 8.21 $ ($ 4.75 $) & $ \mathbf{1.57} $ ($ 3.90 $) &          $ 8.17 $ ($ 4.78 $) & $ \mathbf{1.60} $ ($ 3.98 $) &          $ 8.14 $ ($ 4.84 $) & $ \mathbf{1.62} $ ($ 4.03 $) &          $ 8.17 $ ($ 4.94 $) & $ \mathbf{1.66} $ ($ 4.16 $) &          $ 8.19 $ ($ 5.02 $) & $ \mathbf{1.69} $ ($ 4.25 $) \\
            FPW$_{\hat{\mu}}$                   &          $ 6.83 $ ($ 5.71 $) &          $ 2.00 $ ($ 4.73 $) &          $ 6.82 $ ($ 5.74 $) &          $ 2.03 $ ($ 4.79 $) &          $ 6.81 $ ($ 5.78 $) &          $ 2.06 $ ($ 4.85 $) &          $ 6.85 $ ($ 5.88 $) &          $ 2.10 $ ($ 4.97 $) &          $ 6.87 $ ($ 5.94 $) &          $ 2.12 $ ($ 5.02 $) \\
            FPW$_{\sigma}$                &          $ 6.79 $ ($ 5.81 $) &          $ 2.07 $ ($ 4.82 $) &          $ 6.79 $ ($ 5.83 $) &          $ 2.10 $ ($ 4.88 $) &          $ 6.78 $ ($ 5.87 $) &          $ 2.12 $ ($ 4.92 $) &          $ 6.82 $ ($ 5.96 $) &          $ 2.17 $ ($ 5.06 $) &          $ 6.84 $ ($ 6.04 $) &          $ 2.19 $ ($ 5.10 $) \\
            K$_{7}$                             & $ \mathbf{6.14} $ ($ 6.00 $) &          $ 2.24 $ ($ 4.94 $) & $ \mathbf{6.17} $ ($ 6.05 $) &          $ 2.27 $ ($ 4.98 $) & $ \mathbf{6.20} $ ($ 6.10 $) &          $ 2.29 $ ($ 5.03 $) & $ \mathbf{6.23} $ ($ 6.19 $) &          $ 2.34 $ ($ 5.15 $) & $ \mathbf{6.26} $ ($ 6.23 $) &          $ 2.36 $ ($ 5.19 $) \\
            MMM                                 &          $ 7.09 $ ($ 5.88 $) &          $ 2.18 $ ($ 4.87 $) &          $ 7.09 $ ($ 5.91 $) &          $ 2.20 $ ($ 4.90 $) &          $ 7.09 $ ($ 5.96 $) &          $ 2.23 $ ($ 4.96 $) &          $ 7.13 $ ($ 6.06 $) &          $ 2.28 $ ($ 5.10 $) &          $ 7.14 $ ($ 6.09 $) &          $ 2.30 $ ($ 5.14 $) \\
            EA & $ 8.04 $ ($ 6.20 $) & $ 3.45 $ ($ 5.58 $) & $ 7.43 $ ($ 6.65 $) & $ 2.17 $ ($ 4.31 $) & $ 7.35 $ ($ 5.29 $) & $ 4.35 $ ($ 5.06 $) & $ 7.00 $ ($ 5.19 $) & $ 3.42 $ ($ 6.15 $) & $ 7.80 $ ($ 5.80 $) & $ 3.36 $ ($ 6.08 $) \\
            VPW$_{B}$                           &          $ 6.55 $ ($ 6.23 $) &          $ 2.30 $ ($ 5.17 $) &          $ 6.55 $ ($ 6.27 $) &          $ 2.33 $ ($ 5.23 $) &          $ 6.55 $ ($ 6.32 $) &          $ 2.36 $ ($ 5.29 $) &          $ 6.59 $ ($ 6.42 $) &          $ 2.41 $ ($ 5.42 $) &          $ 6.60 $ ($ 6.47 $) &          $ 2.43 $ ($ 5.46 $) \\ \hline
        \end{tabular}
        \end{singlespace}
    \end{table}
    MNIST \cite{mnist} is a pre-processed subset of NIST's special database 3 and 1 which contain binary images of handwritten digits. Each digit consists of $ 28\times 28 $ pixels image. The training set was created by randomly selecting $ 4000 $ samples from each digits' pool and remaining images became part of the test set. The binary comparison between each pair of digits required $ 45 $ binary LapRLSC classifier models similar to USPS experiment. Each binary model was evaluated $ 20 $ times with $ 2 $ out of $ 4000 $ training samples randomly labeled in both $ +1 $ and $ -1 $ class. Fig. \ref{fig_mnist} shows the result of LapRLSC on both unlabeled and test data set. In this experiment, all methods gave similar accuracy, especially in the unlabeled set as shown in Fig. \subref{fig_mnist_laprlsc_u}. In the test set, VPW$ _{B} $ gave better performance than FPW methods.
    
    The large eigenvalues for all estimators, as shown in Fig. \subref{fig_mnist_ev} results in similar performance. VPW$ _{B} $ was also compared with adaptive Parzen window estimators by varying $ |N| $ values as listed in Table \ref{table:mnist}. It shows that in the unlabeled set, FPW with user-defined Parzen window gave most accurate classification results while it fails to give similar accuracy in the test set due to function over-fitting. In the test set, K$ _{7} $ outperformed other estimators followed by VPW$ _{B} $. This shows that VPW$ _{B} $ exploits the true intrinsic geometrical properties leading to optimal manifold regularization. Similar to USPS data set, EA on MNIST data set also gave poor accuracy with default setting of graph sparsity $ |N| $ and perplexity $ |N|-1 $. By further tuning the perplexity parameter to $ |N|\times3 $ and building a fully connected graph, comparable results were obtained as listed in Table \ref{table:mnist}.
\subsection{Scene detection}
\label{sec:scene_detection}
    High resolution scene image classification poses a huge challenge due to its inherent high dimension and non-local feature similarity properties. Hence, an appropriate affinity metric is required to counter these effects and increase the underlying model's accuracy.
\subsubsection{UC Merced land use data}
\label{sec:ucmerced}
    \begin{figure}[!h]
        \centering
        \subfloat[LapRLSC (unlabeled set)]{\includegraphics[width=0.25\linewidth,clip,keepaspectratio]{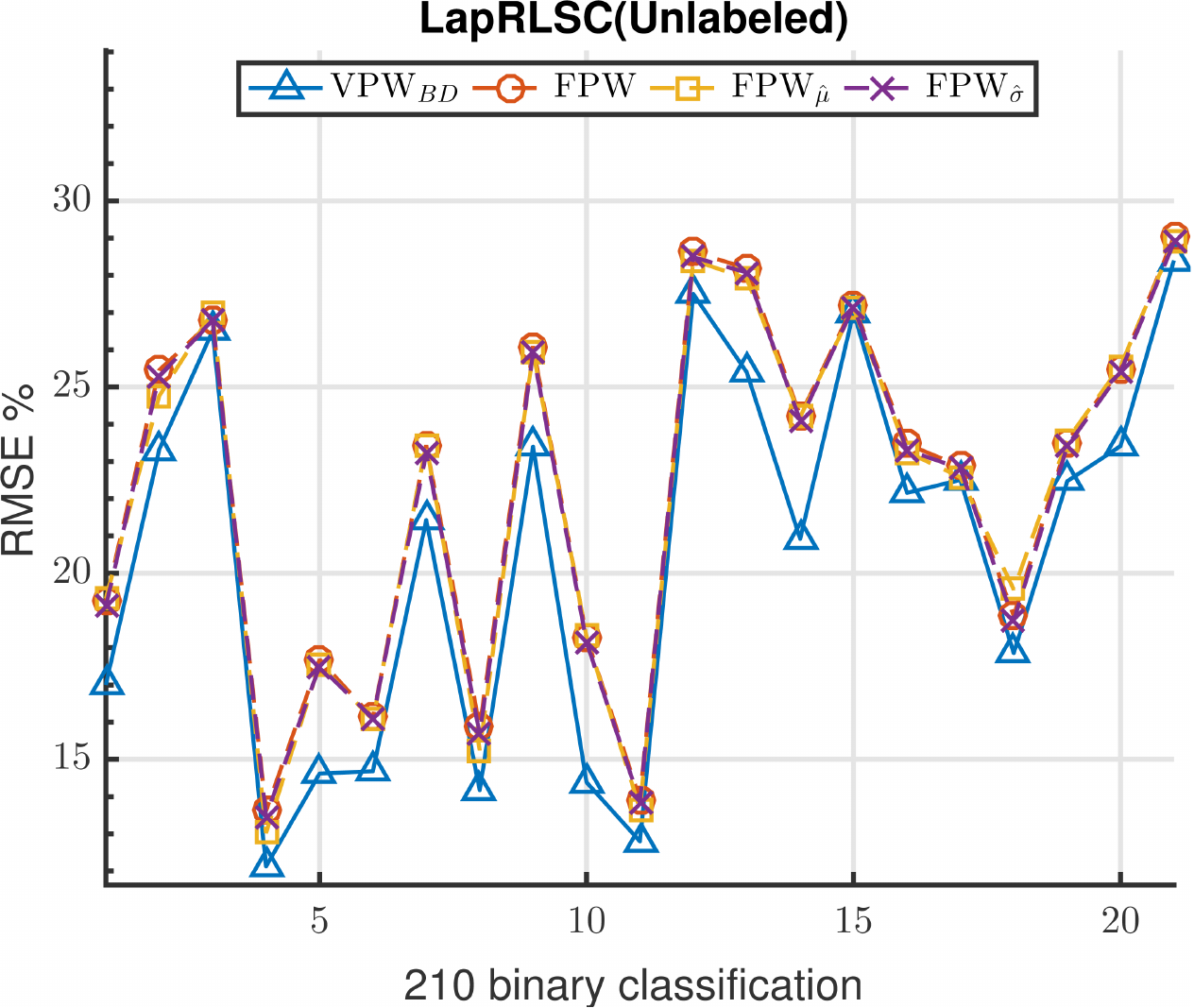}%
            \label{fig_ucmerced_LapRlsc_u}}
        \hfil
        \subfloat[LapRLSC (test set)]{\includegraphics[width=0.25\linewidth,clip,keepaspectratio]{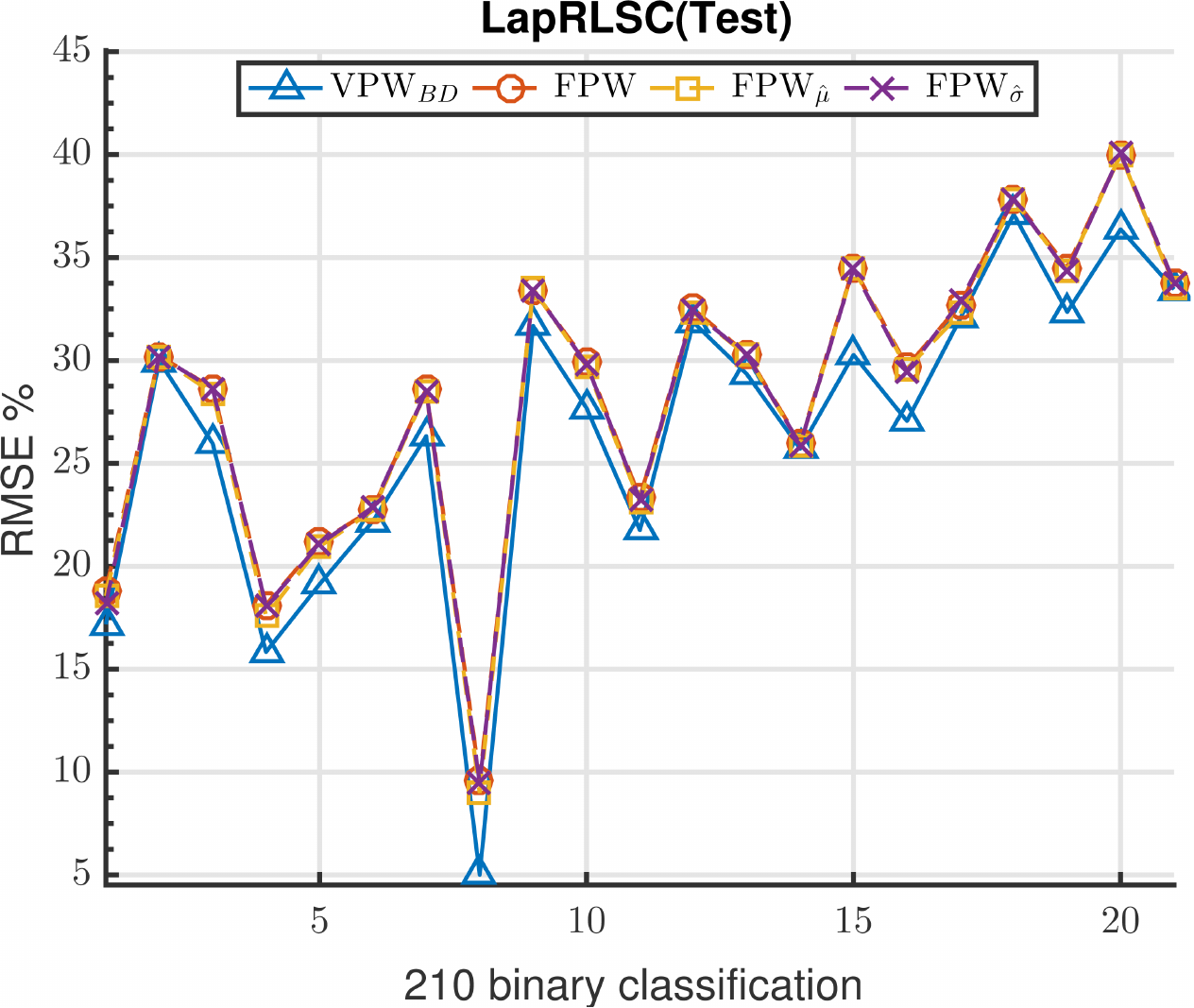}%
            \label{fig_ucmerced_LapRlsc_t}}
        \hfil
        \subfloat[Eigenvalue]{\includegraphics[width=0.25\linewidth,clip,keepaspectratio]{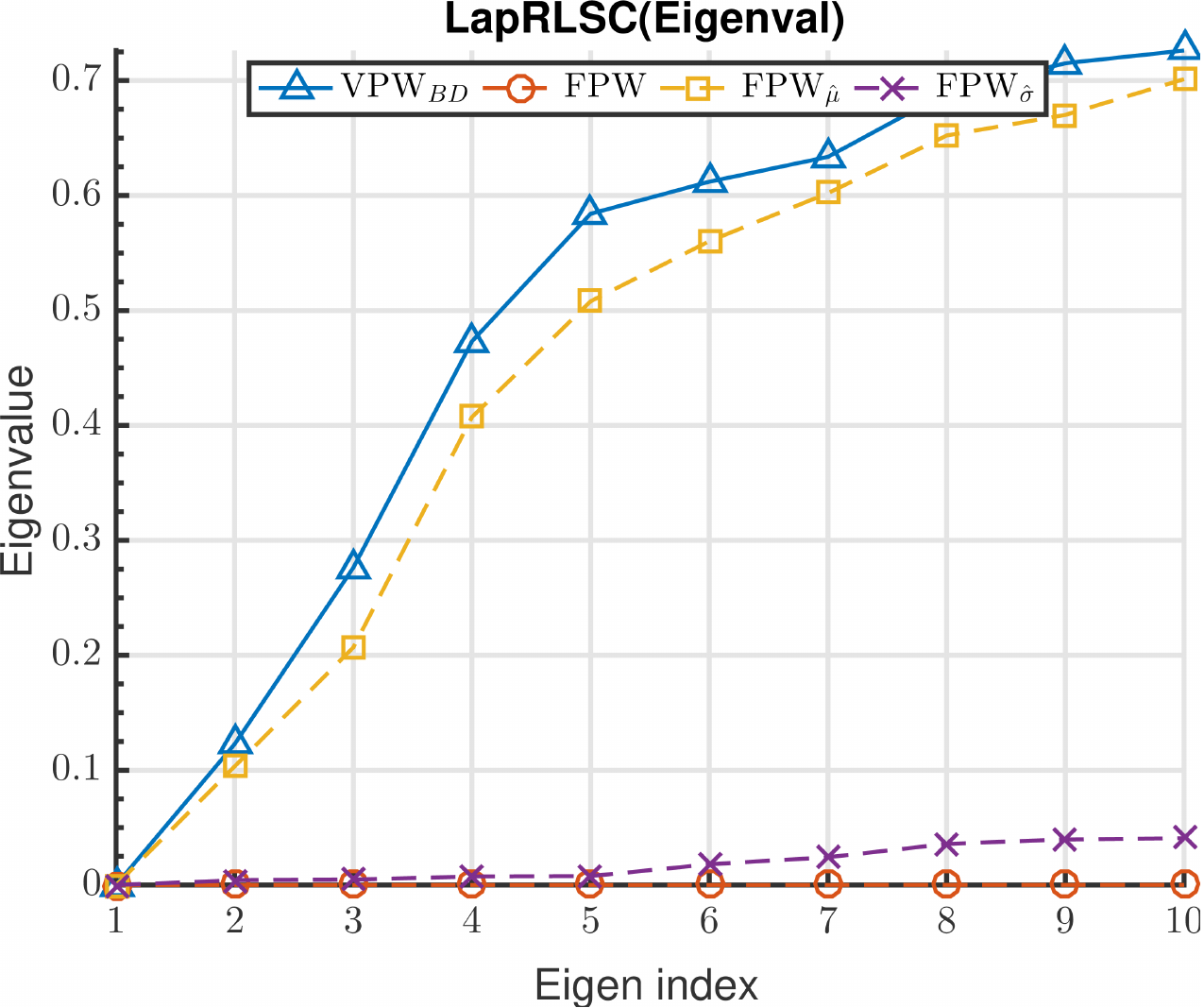}%
            \label{fig_ucmerced_ev}}
        \caption{LapRLSC mean error and eigenvalue comparison between FPW, FPW$ _{\hat{\mu}} $, FPW$ _{\sigma} $, and VPW$ _{BD} $ on UC Merced image}
        \label{fig_ucmerced}
    \end{figure}
    \begin{table}[!h]
        \caption{UC Merced Land image mean error (standard deviation) with varying $ |N| $}
        \label{table:ucmerced}
        \centering
        \small\addtolength{\tabcolsep}{-5pt}
        \begin{singlespace}\scriptsize
        \begin{tabular}{l*{10}{m{1cm}}}
            \hline\hline
            \multirow{2}{*}{\bfseries Affinity} &             \multicolumn{2}{c}{$ |N| $=31}              &             \multicolumn{2}{c}{$ |N| $=32}              &             \multicolumn{2}{c}{$ |N| $=33}              &             \multicolumn{2}{c}{$ |N| $=34}              &             \multicolumn{2}{c}{$ |N| $=35}              \\
            &                    \thead{et} &                    \thead{eu} &                    \thead{et} &                    \thead{eu} &                    \thead{et} &                    \thead{eu} &                    \thead{et} &                    \thead{eu} &                    \thead{et} &                    \thead{eu} \\ \hline
            FPW                                 &          $ 24.14 $ ($ 7.75 $) &          $ 16.64 $ ($ 4.83 $) &          $ 24.46 $ ($ 7.97 $) &          $ 16.36 $ ($ 4.81 $) &          $ 24.54 $ ($ 7.66 $) &          $ 16.69 $ ($ 5.25 $) &          $ 24.16 $ ($ 7.82 $) &          $ 16.72 $ ($ 5.00 $) &          $ 24.38 $ ($ 7.76 $) &          $ 16.61 $ ($ 5.08 $) \\
            FPW$_{\hat{\mu}}$                   &          $ 24.25 $ ($ 8.07 $) &          $ 16.76 $ ($ 4.86 $) &          $ 24.47 $ ($ 7.87 $) &          $ 16.64 $ ($ 4.77 $) &          $ 24.35 $ ($ 7.98 $) &          $ 16.66 $ ($ 5.18 $) &          $ 24.37 $ ($ 7.89 $) &          $ 16.67 $ ($ 4.71 $) &          $ 24.31 $ ($ 8.00 $) &          $ 16.48 $ ($ 5.09 $) \\
            FPW$_{\sigma}$                &          $ 24.44 $ ($ 7.92 $) &          $ 16.75 $ ($ 5.14 $) &          $ 24.41 $ ($ 7.87 $) &          $ 16.54 $ ($ 4.91 $) &          $ 24.51 $ ($ 7.92 $) &          $ 16.78 $ ($ 5.20 $) &          $ 24.51 $ ($ 7.74 $) &          $ 16.64 $ ($ 4.98 $) &          $ 24.23 $ ($ 7.93 $) &          $ 16.48 $ ($ 4.92 $) \\
            K$_{7}$                             &          $ 24.30 $ ($ 8.19 $) &          $ 16.61 $ ($ 5.04 $) &          $ 24.39 $ ($ 7.93 $) &          $ 16.46 $ ($ 5.07 $) &          $ 24.63 $ ($ 7.96 $) &          $ 16.55 $ ($ 4.80 $) &          $ 24.12 $ ($ 7.82 $) &          $ 16.69 $ ($ 4.86 $) &          $ 24.15 $ ($ 8.14 $) &          $ 16.71 $ ($ 4.79 $) \\
            MMM                                 &          $ 24.20 $ ($ 8.15 $) &          $ 16.87 $ ($ 4.81 $) &          $ 24.32 $ ($ 8.31 $) &          $ 16.48 $ ($ 4.75 $) &          $ 24.54 $ ($ 8.10 $) &          $ 16.66 $ ($ 4.83 $) &          $ 24.19 $ ($ 8.27 $) &          $ 16.44 $ ($ 4.82 $) &          $ 24.26 $ ($ 8.10 $) &          $ 16.54 $ ($ 4.90 $) \\
            EA & $ 25.10 $ ($ 7.95 $) & $ 17.67 $ ($ 5.00 $) & $ 25.11 $ ($ 7.93 $) & $ 17.65 $ ($ 5.00 $) & $ 25.11 $ ($ 7.93 $) & $ 17.64 $ ($ 4.98 $) & $ 25.10 $ ($ 7.94 $) & $ 17.61 $ ($ 4.98 $) & $ 25.11 $ ($ 7.93 $) & $ 17.61 $ ($ 4.99 $) \\
            VPW$_{BD}$                           & $ \mathbf{22.98} $ ($ 8.10 $) & $ \mathbf{15.53} $ ($ 5.37 $) & $ \mathbf{23.49} $ ($ 7.96 $) & $ \mathbf{15.64} $ ($ 5.15 $) & $ \mathbf{23.17} $ ($ 8.48 $) & $ \mathbf{15.15} $ ($ 5.44 $) & $ \mathbf{23.17} $ ($ 7.94 $) & $ \mathbf{15.66} $ ($ 4.67 $) & $ \mathbf{22.98} $ ($ 8.23 $) & $ \mathbf{15.23} $ ($ 4.75 $) \\ \hline
        \end{tabular}
        \end{singlespace}
    \end{table}
    The UC Merced land data set \cite{yang2010bag} consists of $ 21 $ categories (agriculture, airplane, forest, freeway, etc.) with each category consists of $ 100 $ high resolution images of $ 256\times 256\times 3 $ dimensions. The training and test set were created by randomly dividing images from each category in two halves. The whole classification model consisted of $ 210 $ binary LapRLSC models. They were executed $ 20 $ times with $ 2 $ labels randomly selected in both $ +1 $ and $ -1 $ class.

    Fig. \ref{fig_ucmerced} shows the result of comparison between FPW and VPW$ _{BD} $ Parzen window estimators. In this experiment, Euclidean weight $ e_{ij} $ with Bhattacharyya distance affinity adjustment $ bd_{ij} $ outperformed all other methods. The unlabeled and test set results are in shown in Fig. \subref{fig_ucmerced_LapRlsc_u} and Fig. \subref{fig_ucmerced_LapRlsc_t} respectively. A custom fitted Parzen window using VPW$ _{BD} $ avoided function over-fitting on seen unlabeled instances and led to accurate label propagation than other estimators. Due to similar features in agricultural land-baseball diamond, storage tank-sparse residential and tennis court-medium residential categories, the performance of VPW$ _{BD} $'s and FPW remained close.
    
    The eigenvalue comparison in Fig. \subref{fig_ucmerced_ev} shows that higher values of VPW$ _{BD} $ result in proper manifold regularization; thus, leading to a generic model. As Parzen window largely depends on the neighborhood size, the performance of methods with varying $ |N| $ has been listed in Table \ref{table:ucmerced}. The table contains the mean error percentage overall $ 21 $ categories along with their respective standard deviation. The general result trend shows that proposed VPW$ _{BD} $ outperforms all other adaptive bandwidth estimators. The increase in $ |N| $ value enhanced the accuracy of the underlying model by connecting more similar data points. This also increases the model's accuracy on categories sharing similar features. However, in a few cases, it adversely affected the readily separable categories by introducing affinity artificially between distant neighbors. Hence, the mean error first increases with $ |N| =31 $ and subsequently decreases with further increase in neighborhood size.
\subsubsection{Indoor scene recognition}
\label{sec:cvpr}
    \begin{figure}[!h]
        \centering
        \subfloat[LapRLSC (unlabeled set)]{\includegraphics[width=0.25\linewidth,clip,keepaspectratio]{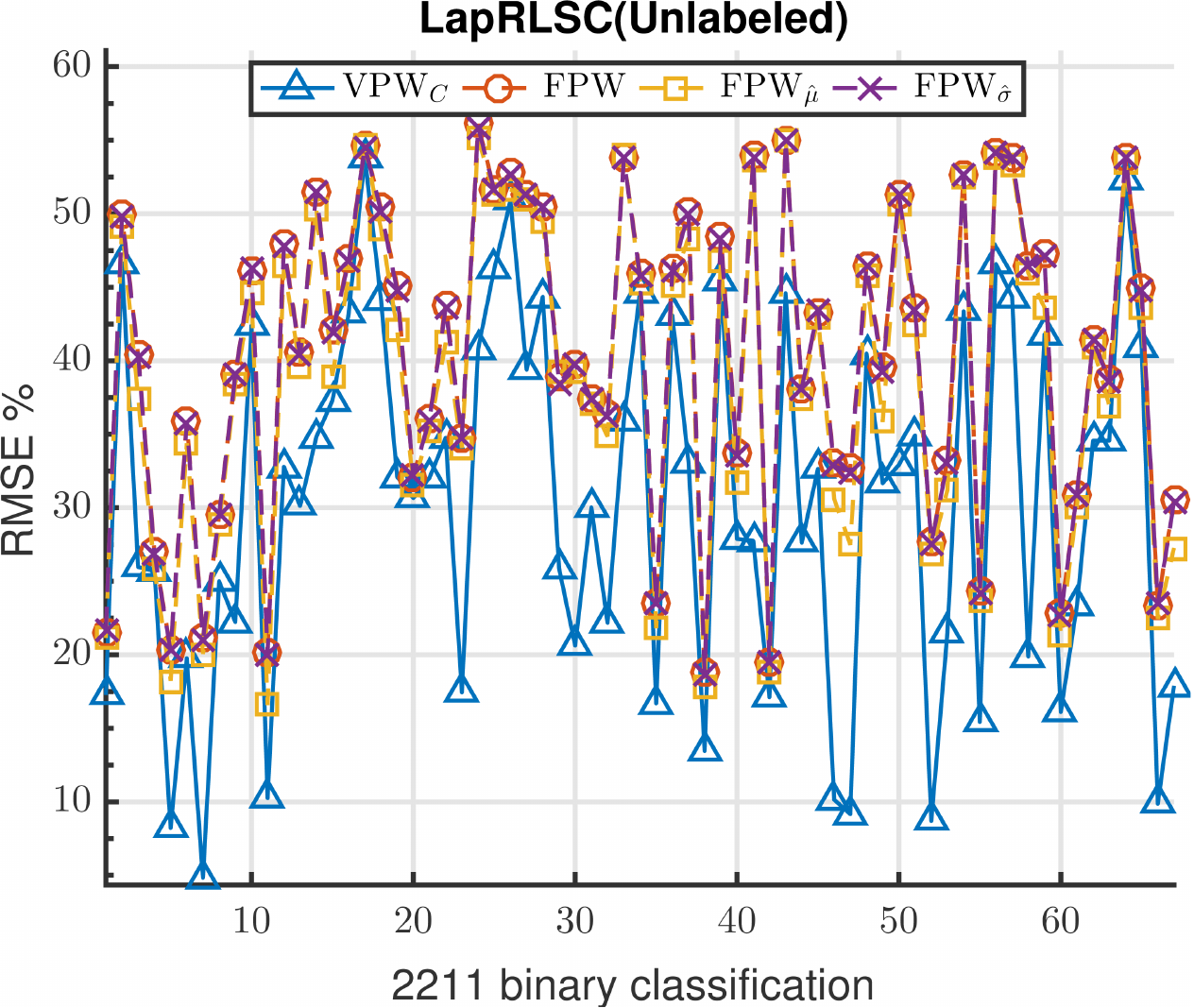}%
            \label{fig_cvpr_u}}
        \hfil
        \subfloat[LapRLSC (test set)]{\includegraphics[width=0.25\linewidth,clip,keepaspectratio]{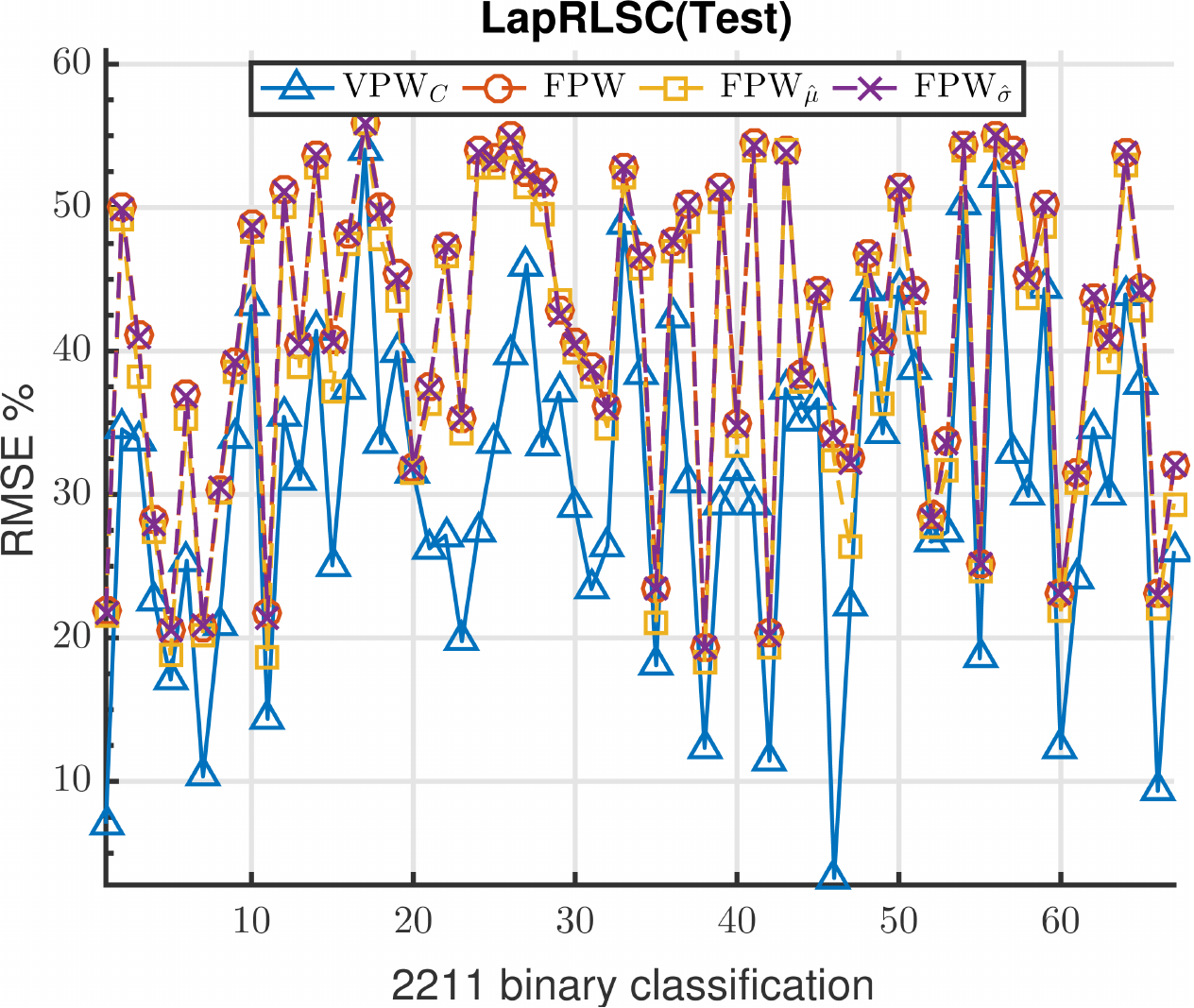}%
            \label{fig_cvpr_t}}
        \hfil
        \subfloat[Eigenvalue]{\includegraphics[width=0.25\linewidth,clip,keepaspectratio]{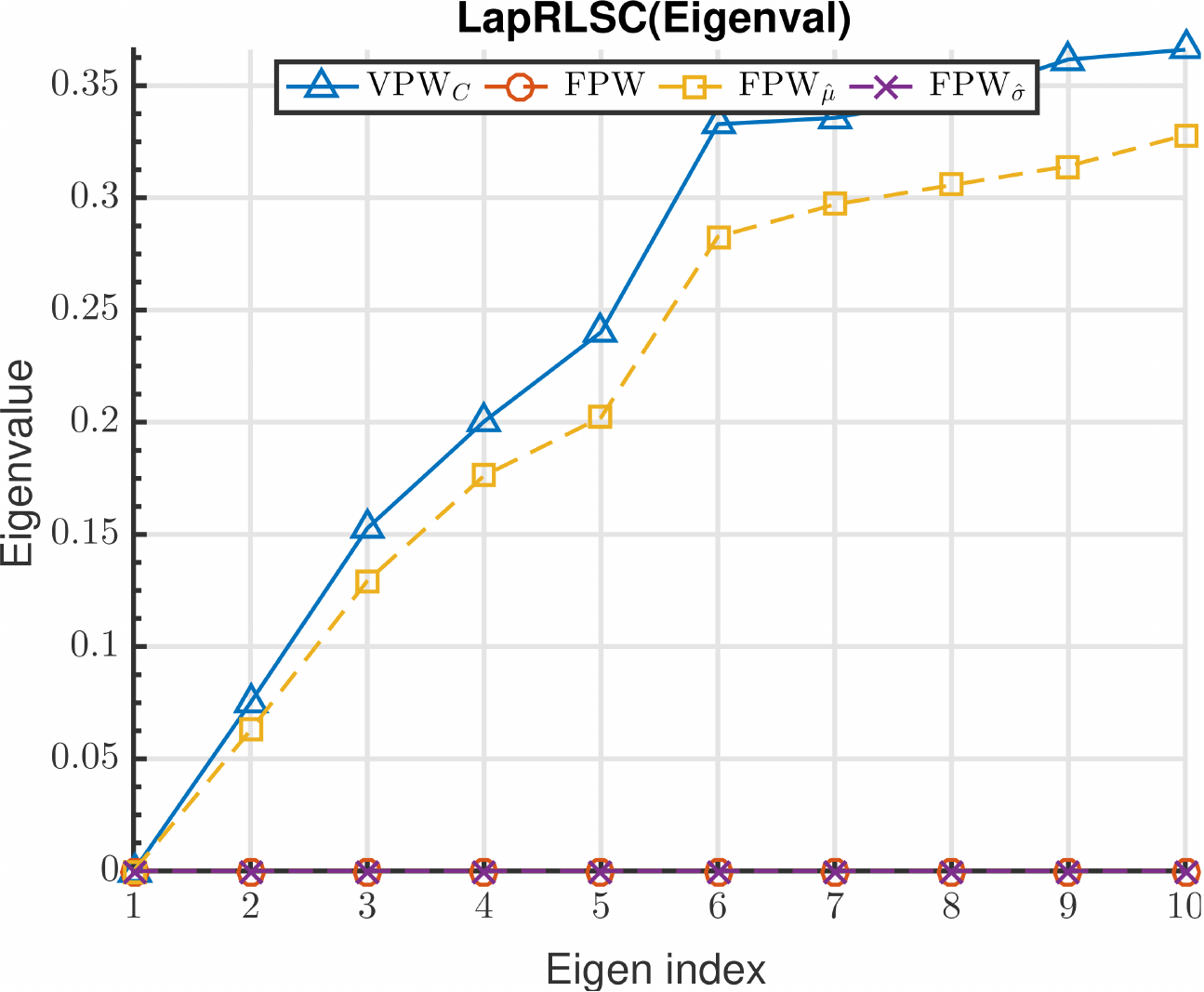}%
            \label{fig_cvpr_ev}}
        \caption{LapRLSC mean error and eigenvalue comparison between FPW, FPW$ _{\hat{\mu}} $, FPW$ _{\sigma} $, and VPW$ _{C} $ on CVPR'09 image}
        \label{fig_cvpr}
    \end{figure}
    \begin{table}[!h]
        \caption{CVPR'09 mean error (standard deviation) with varying $ |N| $}
        \label{table:cvpr}
        \centering
        \small\addtolength{\tabcolsep}{-6pt}
        \begin{singlespace}\scriptsize
        \begin{tabular}{l*{10}{m{1cm}}}
            \hline\hline
            \multirow{2}{*}{\bfseries Affinity} &              \multicolumn{2}{c}{$ |N| $=31}               &              \multicolumn{2}{c}{$ |N| $=32}               &              \multicolumn{2}{c}{$ |N| $=33}               &              \multicolumn{2}{c}{$ |N| $=34}               &              \multicolumn{2}{c}{$ |N| $=35}               \\
            &                     \thead{et} &                     \thead{eu} &                     \thead{et} &                     \thead{eu} &                     \thead{et} &                     \thead{eu} &                     \thead{et} &                     \thead{eu} &                     \thead{et} &                     \thead{eu} \\ \hline
            FPW                                 &          $ 33.25 $ ($ 10.14 $) &          $ 32.21 $ ($ 11.02 $) &          $ 33.26 $ ($ 10.14 $) &          $ 32.24 $ ($ 11.02 $) &          $ 33.33 $ ($ 10.21 $) &          $ 32.24 $ ($ 10.98 $) &          $ 33.37 $ ($ 10.13 $) &          $ 32.18 $ ($ 11.02 $) &          $ 33.32 $ ($ 10.15 $) &          $ 32.15 $ ($ 11.04 $) \\
            FPW$_{\hat{\mu}}$                   &          $ 33.36 $ ($ 10.10 $) &          $ 32.21 $ ($ 10.94 $) &          $ 33.22 $ ($ 10.10 $) &          $ 32.22 $ ($ 11.00 $) &          $ 33.39 $ ($ 10.13 $) &          $ 32.26 $ ($ 11.15 $) &          $ 33.34 $ ($ 10.10 $) &          $ 32.27 $ ($ 11.05 $) &          $ 33.35 $ ($ 10.17 $) &          $ 32.25 $ ($ 11.09 $) \\
            FPW$_{\sigma}$                &          $ 33.35 $ ($ 10.05 $) &          $ 32.22 $ ($ 11.06 $) &          $ 33.33 $ ($ 10.20 $) &          $ 32.23 $ ($ 10.96 $) &          $ 33.24 $ ($ 10.29 $) &          $ 32.15 $ ($ 11.01 $) &          $ 33.32 $ ($ 10.05 $) &          $ 32.29 $ ($ 11.03 $) &          $ 33.35 $ ($ 10.09 $) &          $ 32.15 $ ($ 11.02 $) \\
            K$_{7}$                             &          $ 33.21 $ ($ 10.17 $) &          $ 32.26 $ ($ 10.98 $) &          $ 33.33 $ ($ 10.19 $) &          $ 32.09 $ ($ 10.91 $) &          $ 33.48 $ ($ 10.13 $) &          $ 32.05 $ ($ 10.84 $) &          $ 33.43 $ ($ 10.13 $) &          $ 32.22 $ ($ 10.93 $) &          $ 33.28 $ ($ 10.15 $) &          $ 32.24 $ ($ 10.97 $) \\
            MMM                                 &          $ 33.31 $ ($ 10.22 $) &          $ 32.27 $ ($ 11.16 $) &          $ 33.40 $ ($ 10.19 $) &          $ 32.20 $ ($ 11.01 $) &          $ 33.32 $ ($ 10.25 $) &          $ 32.08 $ ($ 10.98 $) &          $ 33.32 $ ($ 10.16 $) &          $ 32.19 $ ($ 10.98 $) &          $ 33.23 $ ($ 10.12 $) &          $ 32.20 $ ($ 10.88 $) \\
            EA & $ 34.07 $ ($ 10.14 $) & $ 32.97 $ ($ 11.00 $) & $ 34.07 $ ($ 10.14 $) & $ 32.97 $ ($ 11.00 $) & $ 34.07 $ ($ 10.14 $) & $ 32.97 $ ($ 11.00 $) & $ 34.07 $ ($ 10.14 $) & $ 32.97 $ ($ 11.00 $) & $ 34.07 $ ($ 10.14 $) & $ 32.97 $ ($ 11.00 $) \\
            VPW$_{C}$                           & $ \mathbf{31.61} $ ($ 10.26 $) & $ \mathbf{30.65} $ ($ 10.95 $) & $ \mathbf{31.77} $ ($ 10.00 $) & $ \mathbf{30.62} $ ($ 11.07 $) & $ \mathbf{31.79} $ ($ 10.24 $) & $ \mathbf{30.55} $ ($ 11.00 $) & $ \mathbf{31.94} $ ($ 10.09 $) & $ \mathbf{30.86} $ ($ 11.03 $) & $ \mathbf{31.85} $ ($ 10.10 $) & $ \mathbf{30.76} $ ($ 11.03 $) \\ \hline
        \end{tabular}
        \end{singlespace}
    \end{table}
    The indoor scene recognition CVPR'09 data set \cite{quattoni2009recognizing} contains indoor images across $ 67 $ categories. Each location category consists of more than $ 100 $ images having $ \approx 500\times 350\times 3 $ pixels. As the image dimensions were inconsistent, during pre-processing, they were resized to $ 256\times 256\times 3 $ pixels. The training and testing data set were created by dividing images from each category into two halves. The complete classification model consisted of $ 2211 $ binary LapRLSC classifiers where each binary model was executed $ 20 $ times with $ 2 $ randomly labeled samples for both $ +1 $ and $ -1 $ classes.

    Due to a large number of categories sharing similar features, the performance of LapRLSC classifier degraded. In a few categories, the label propagation error went beyond $ 50\% $. However, in comparison with all adaptive Parzen window estimators, the proposed VPW$ _{C} $ gave more accurate results. The highest accuracy given by FPW in few categories was $ \approx 80\% $, in the same categories, VPW$ _{C} $ increased the model's accuracy to $ 90\% $. The VPW$ _{C} $ based affinity even brought down the mean error across various categories below $ 50\% $ by discarding the affinity drift towards high-density regions. The centroid distance based affinity adjustment was able to identify true distribution properties around the point of interest and hence, discards the effects of uneven sampling.
\begin{table*}[!t]
    \caption{Random walk using affinity adjustments on real-world data set}
    \label{table:fig_randomwalk}
    \centering
    \begin{singlespace}\scriptsize
    \begin{tabular}{|m{0.15\linewidth}|m{0.15\linewidth}|m{0.15\linewidth}|m{0.15\linewidth}|}
        \hline
        \theadm{Adjustment$ \rightarrow $}{Data set$ \downarrow $} & \theadb{$ b_{ij} $}          & \theadb{$ c_{ij} $}          & \theadb{$ bd_{ij} $}          \\ \hline
        \theadb{HaLT}                                              & \tgraph{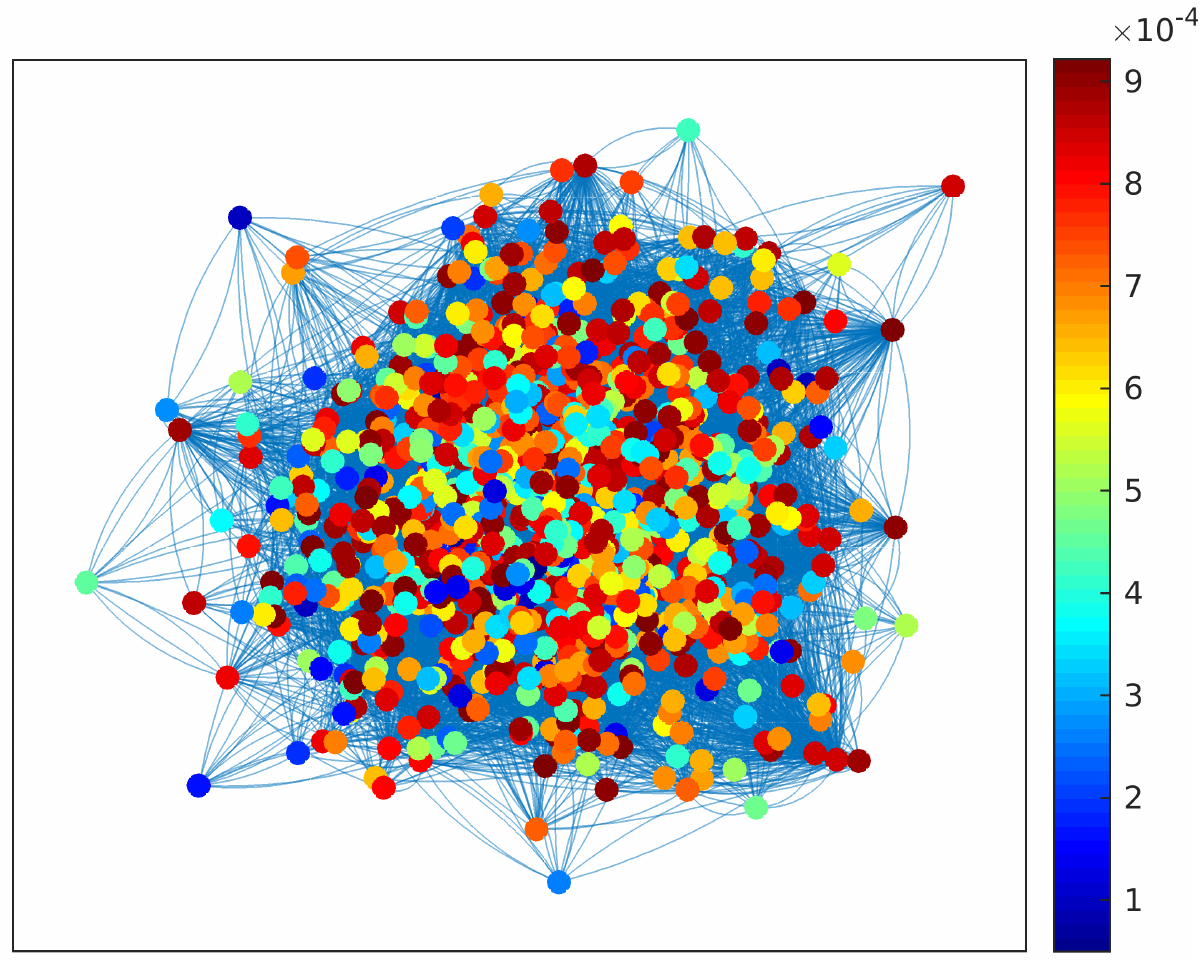}    & \tgraph{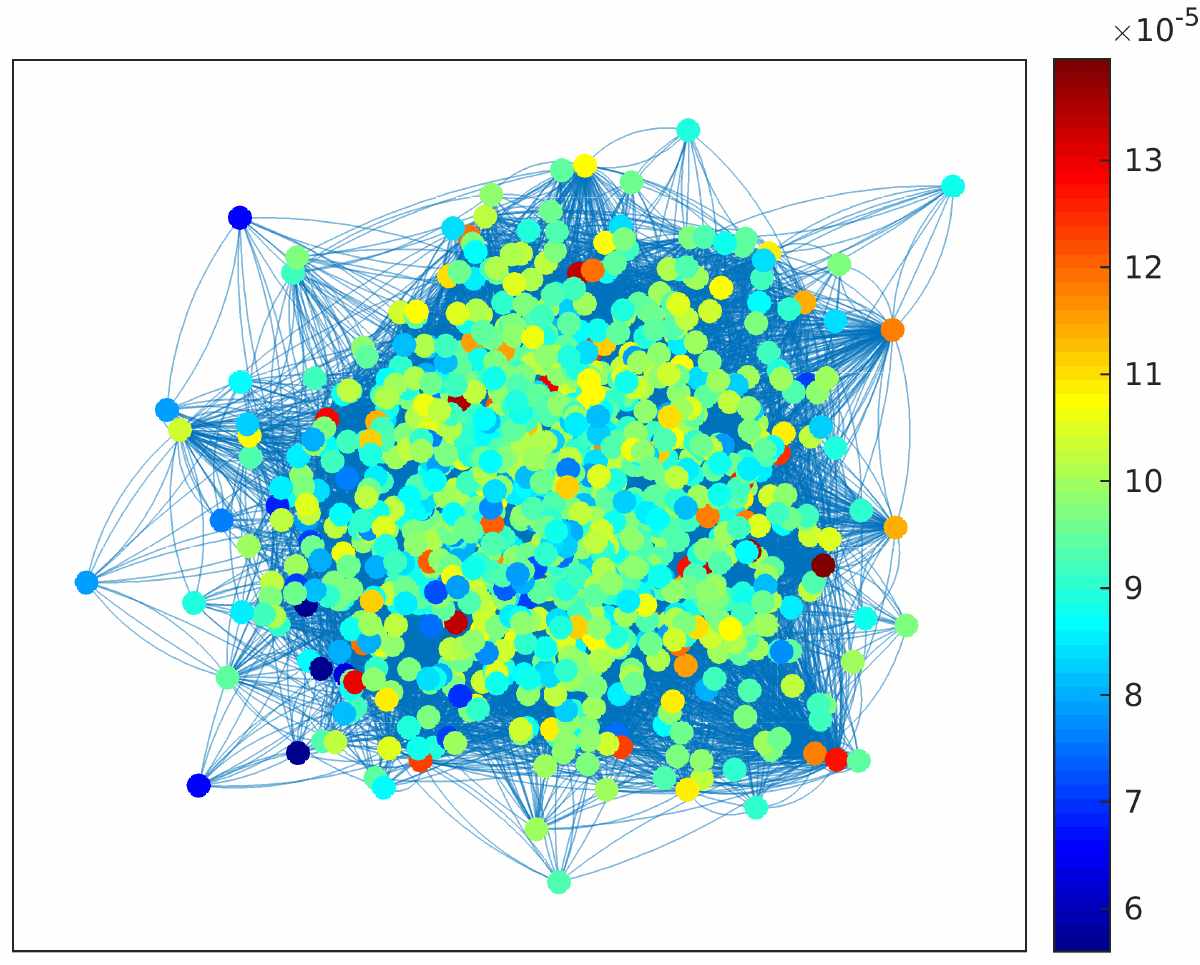}    & \tgraph{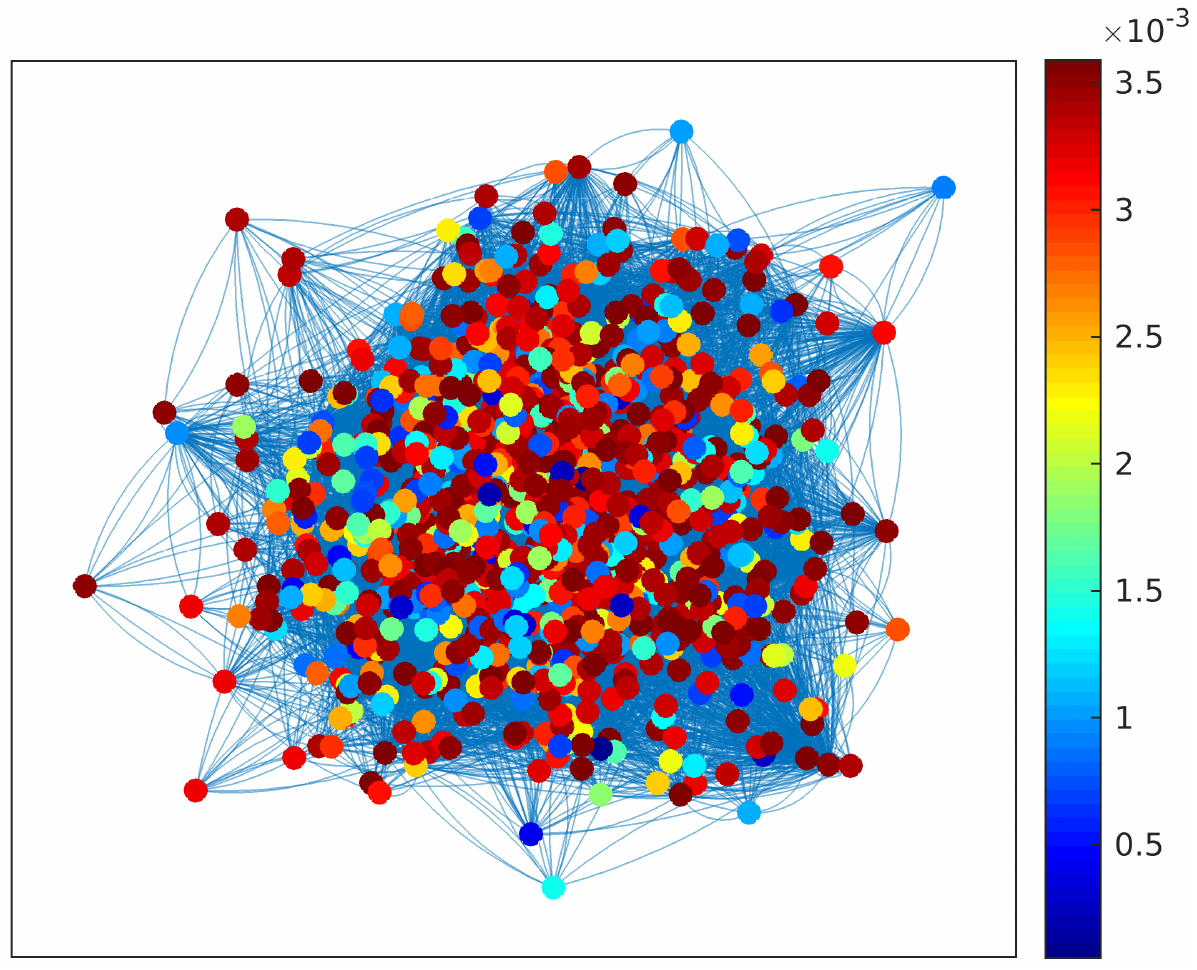}    \\ \hline
        \theadb{Hasy\_v2}                                          & \tgraph{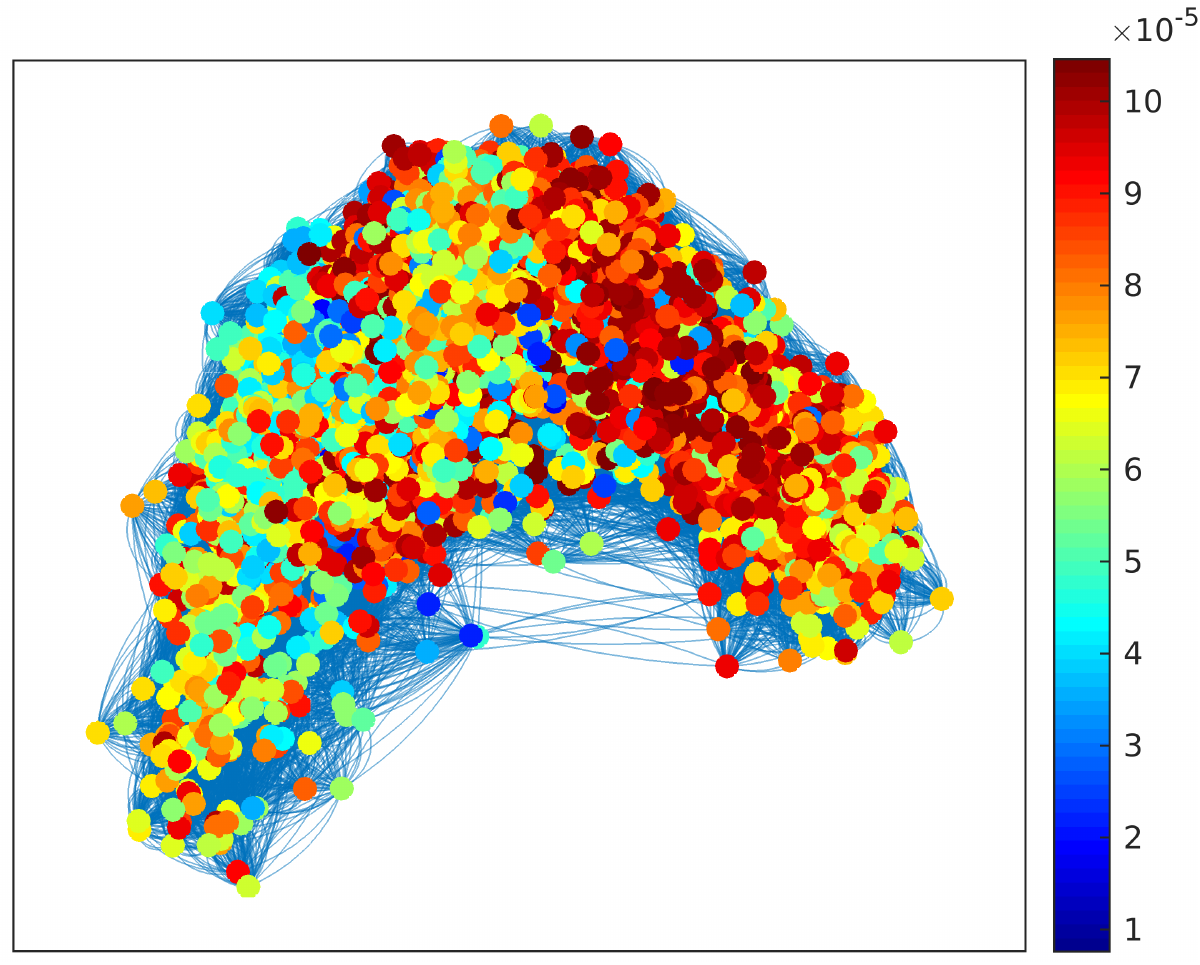}        & \tgraph{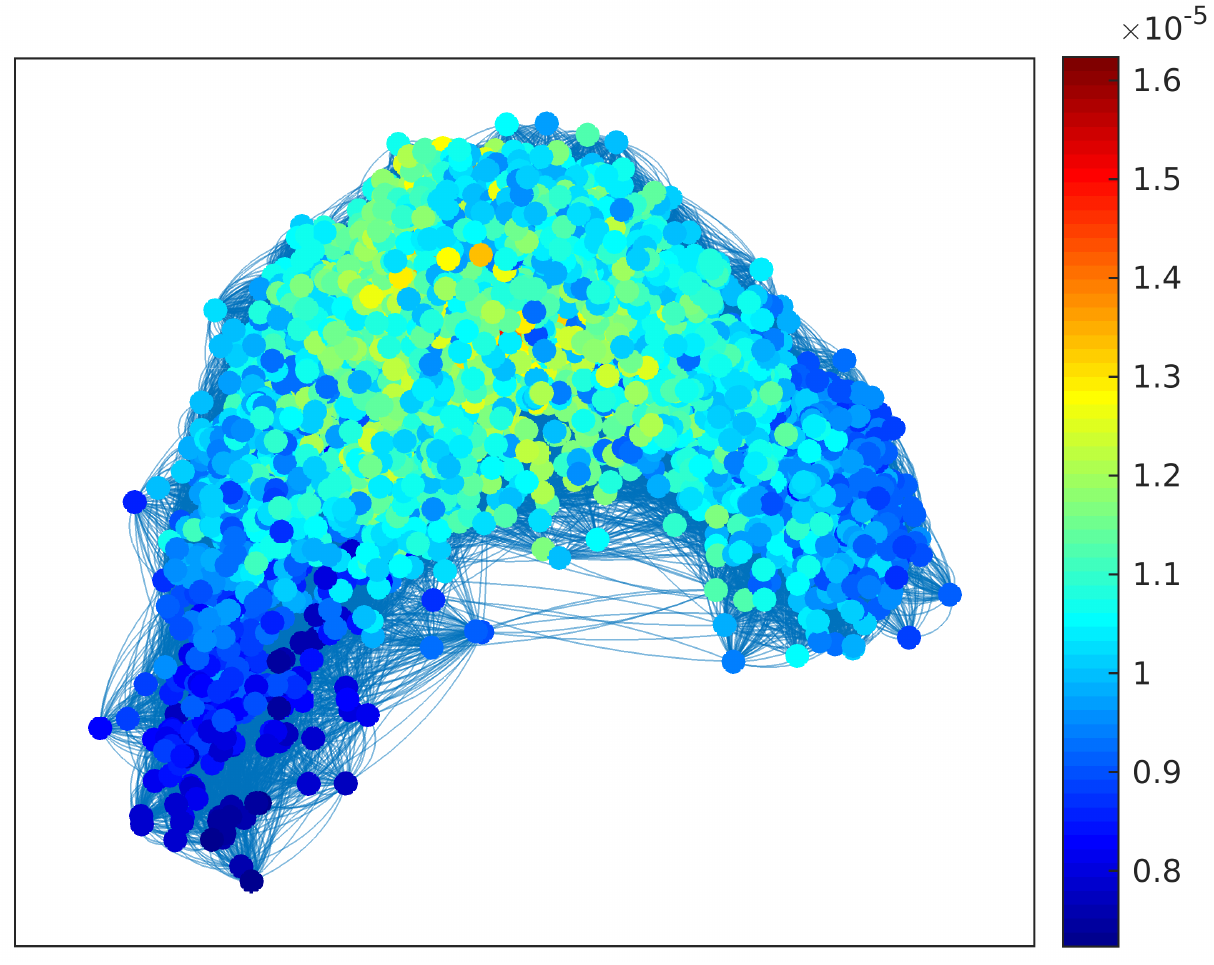}        & \tgraph{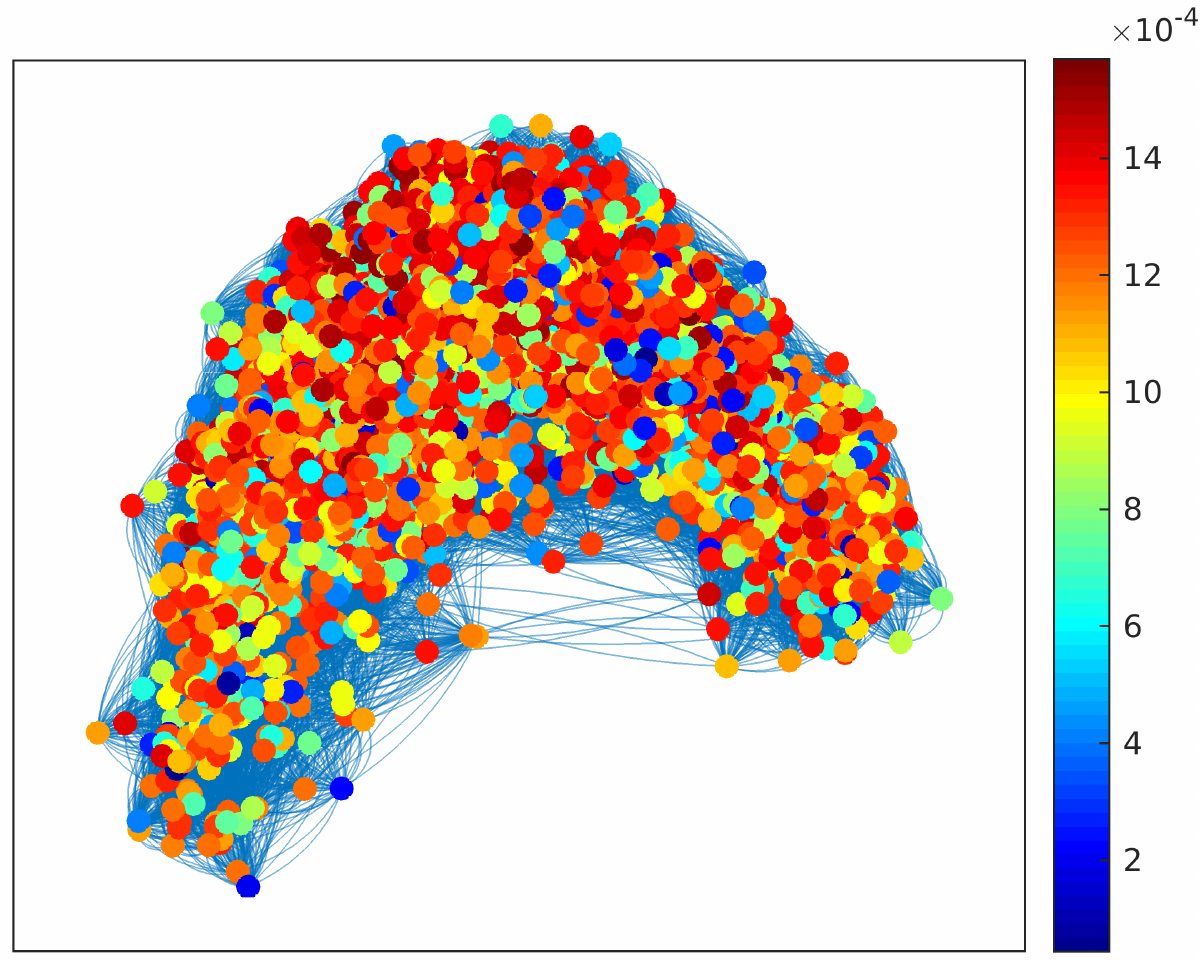}        \\ \hline
        \theadb{USPS}                                              & \tgraph{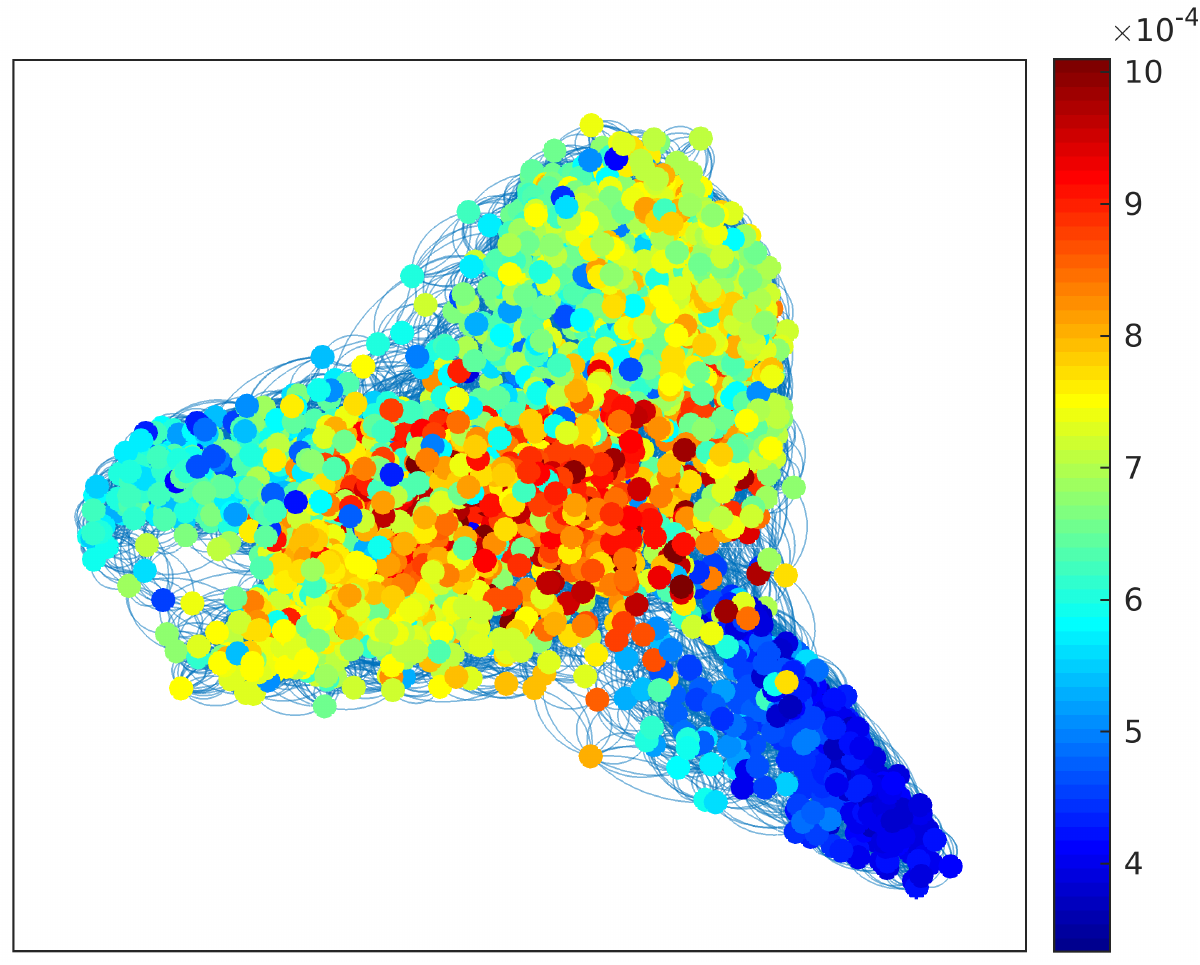}         & \tgraph{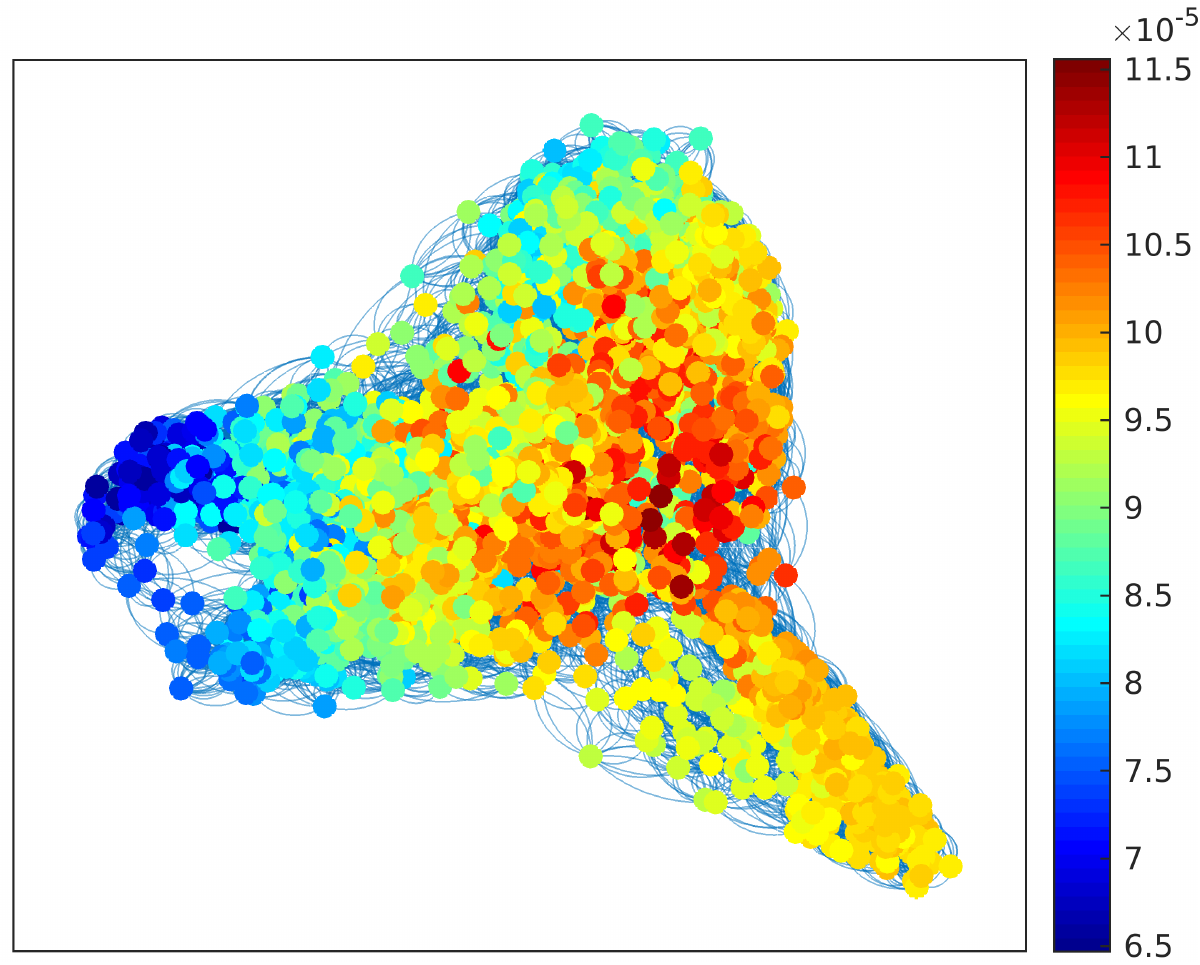}         & \tgraph{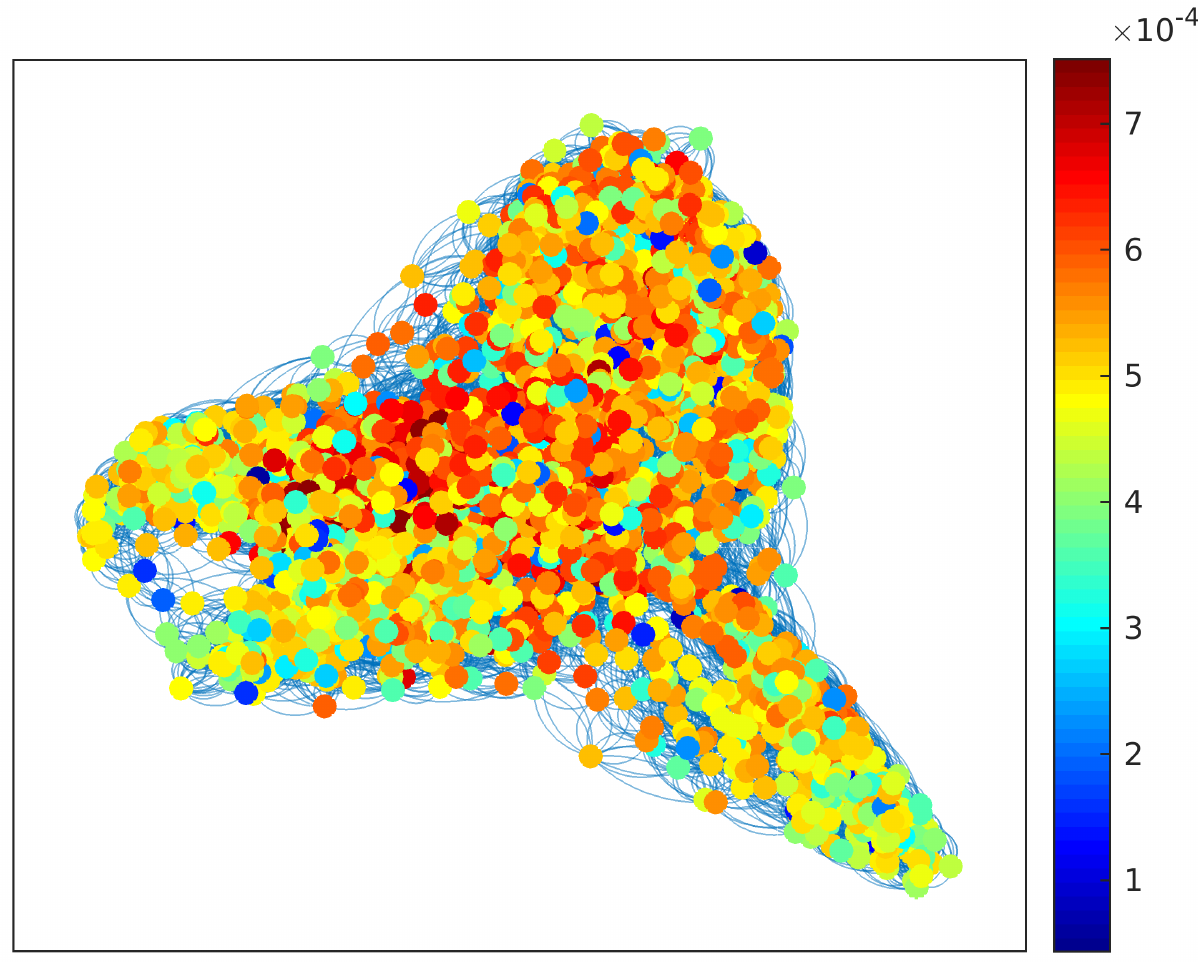}         \\ \hline
        \theadb{MNIST}                                             & \tgraph{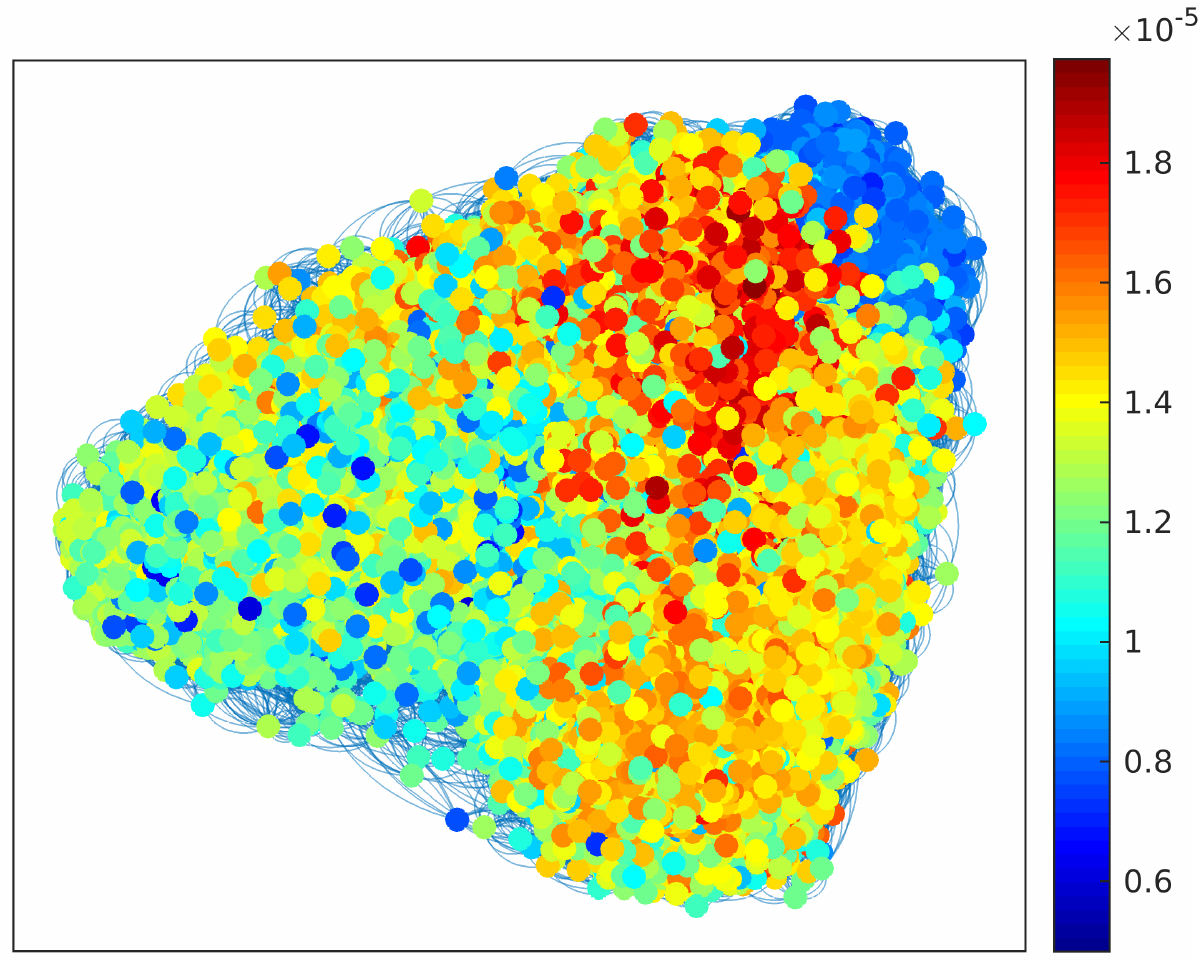}        & \tgraph{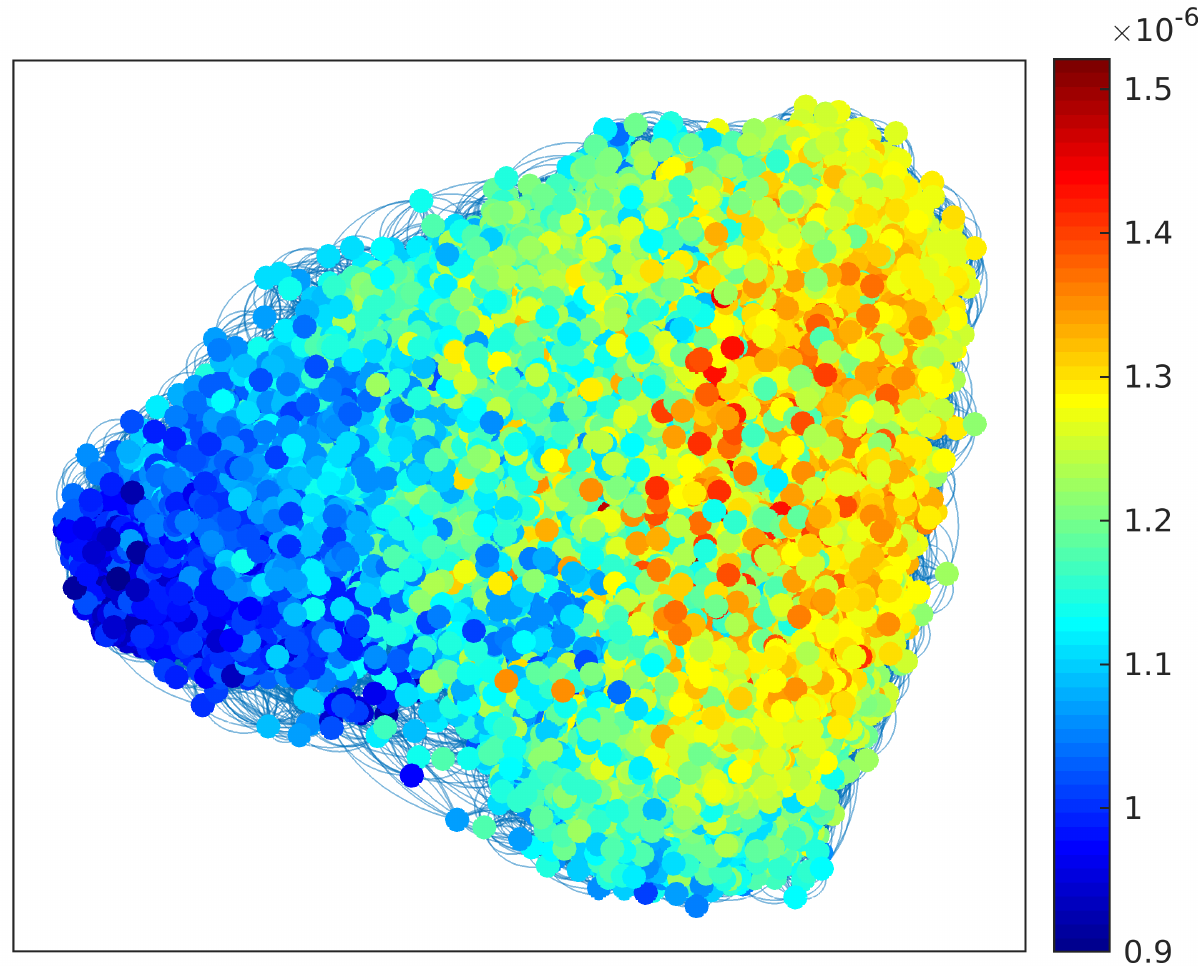}        & \tgraph{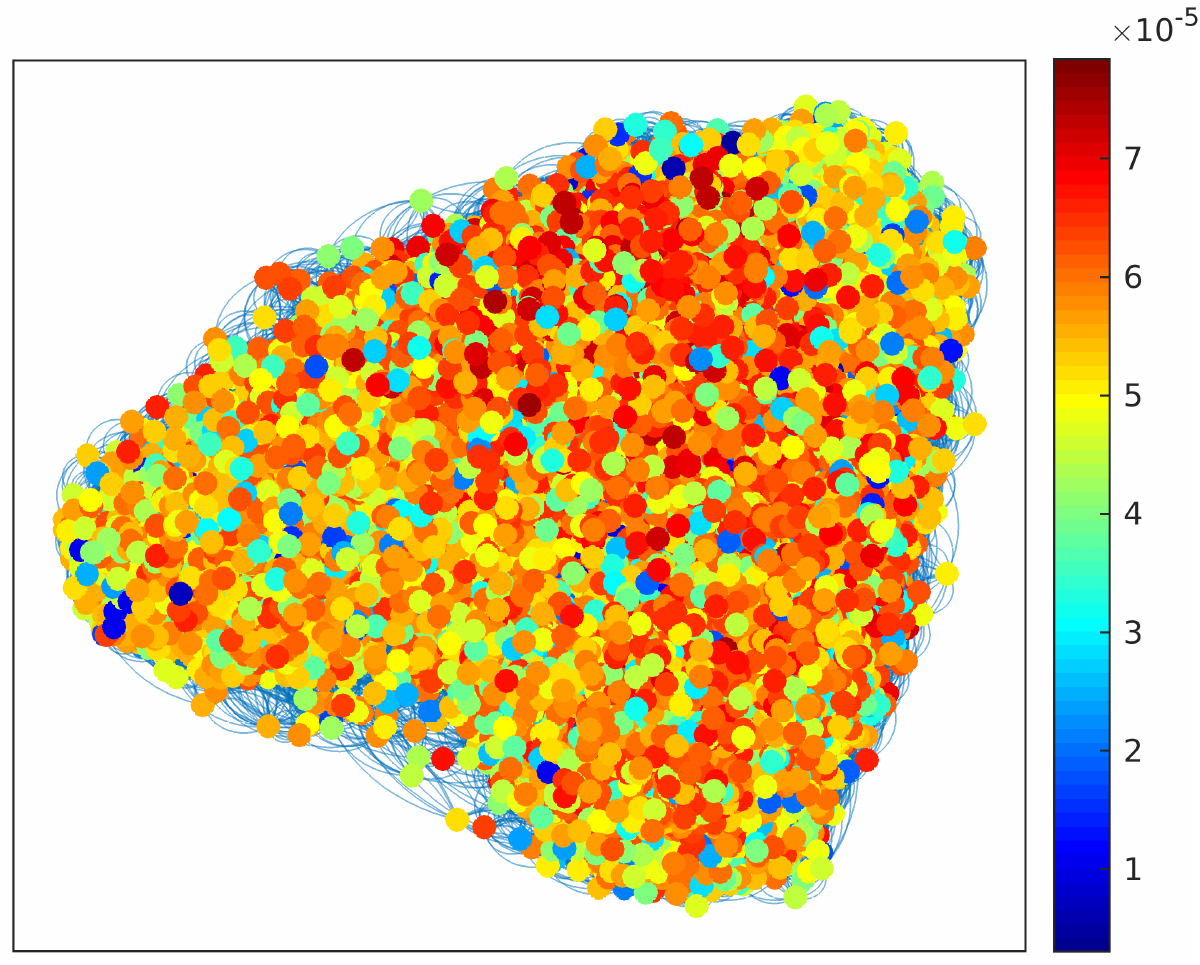}        \\ \hline
        \theadb{UC Merced}                                         & \tgraph{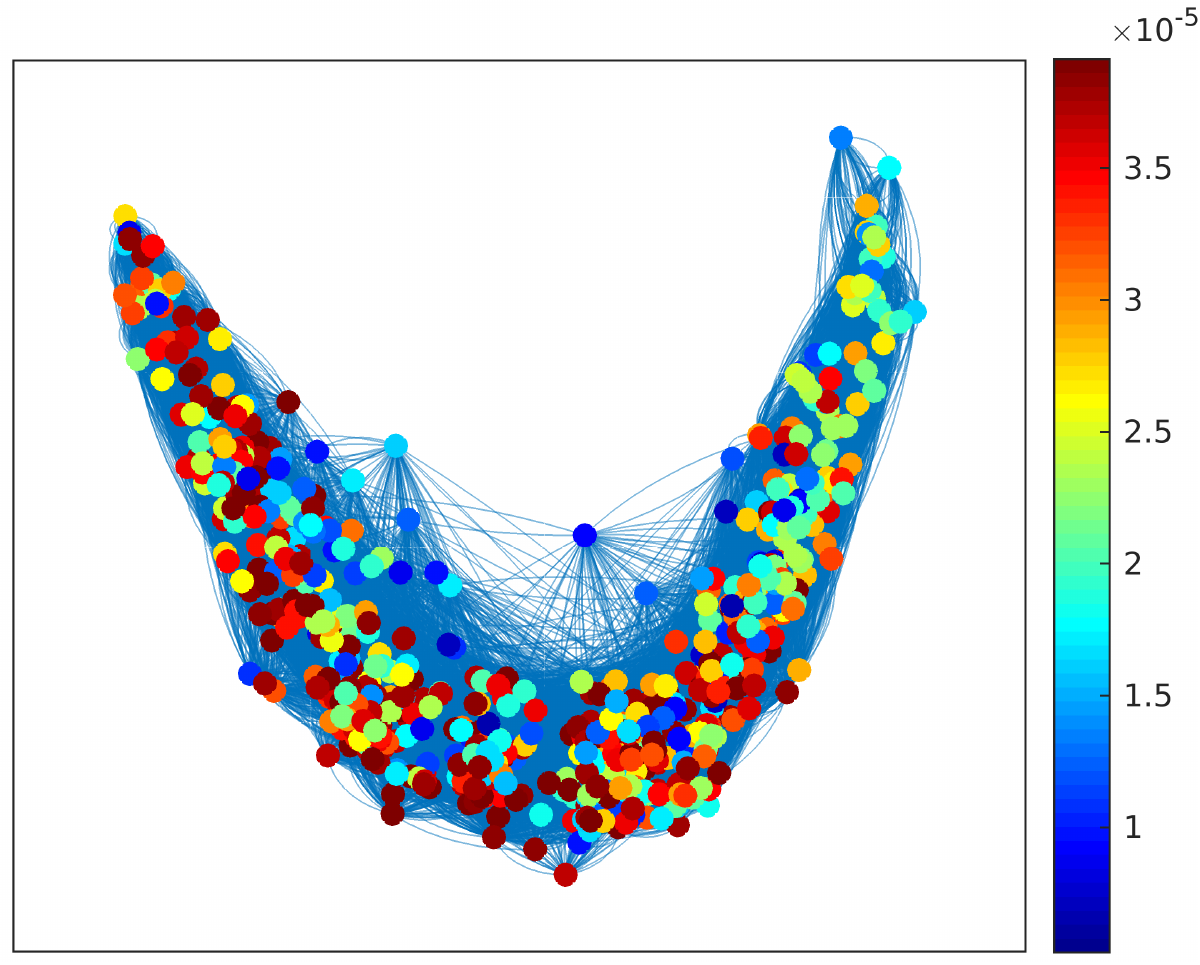} & \tgraph{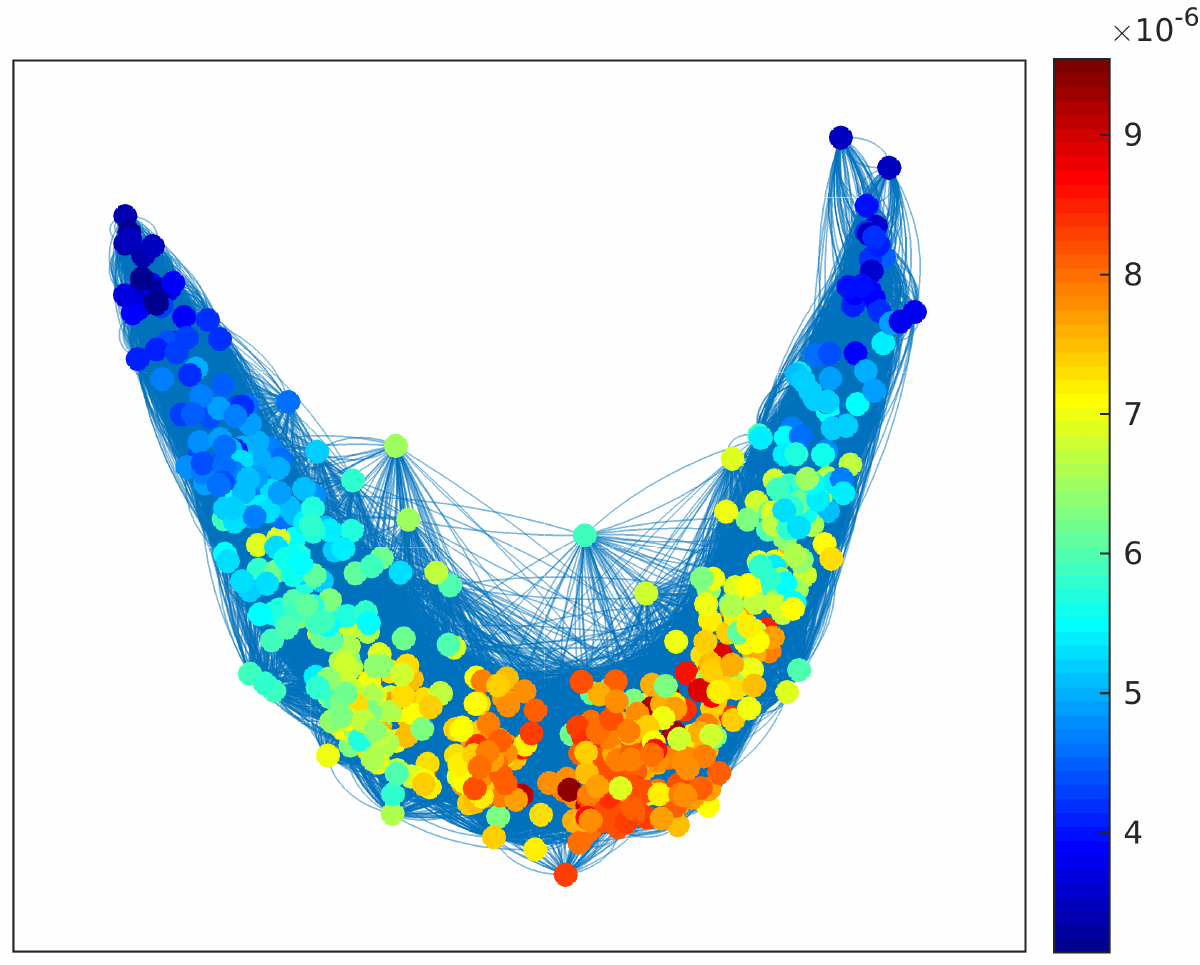} & \tgraph{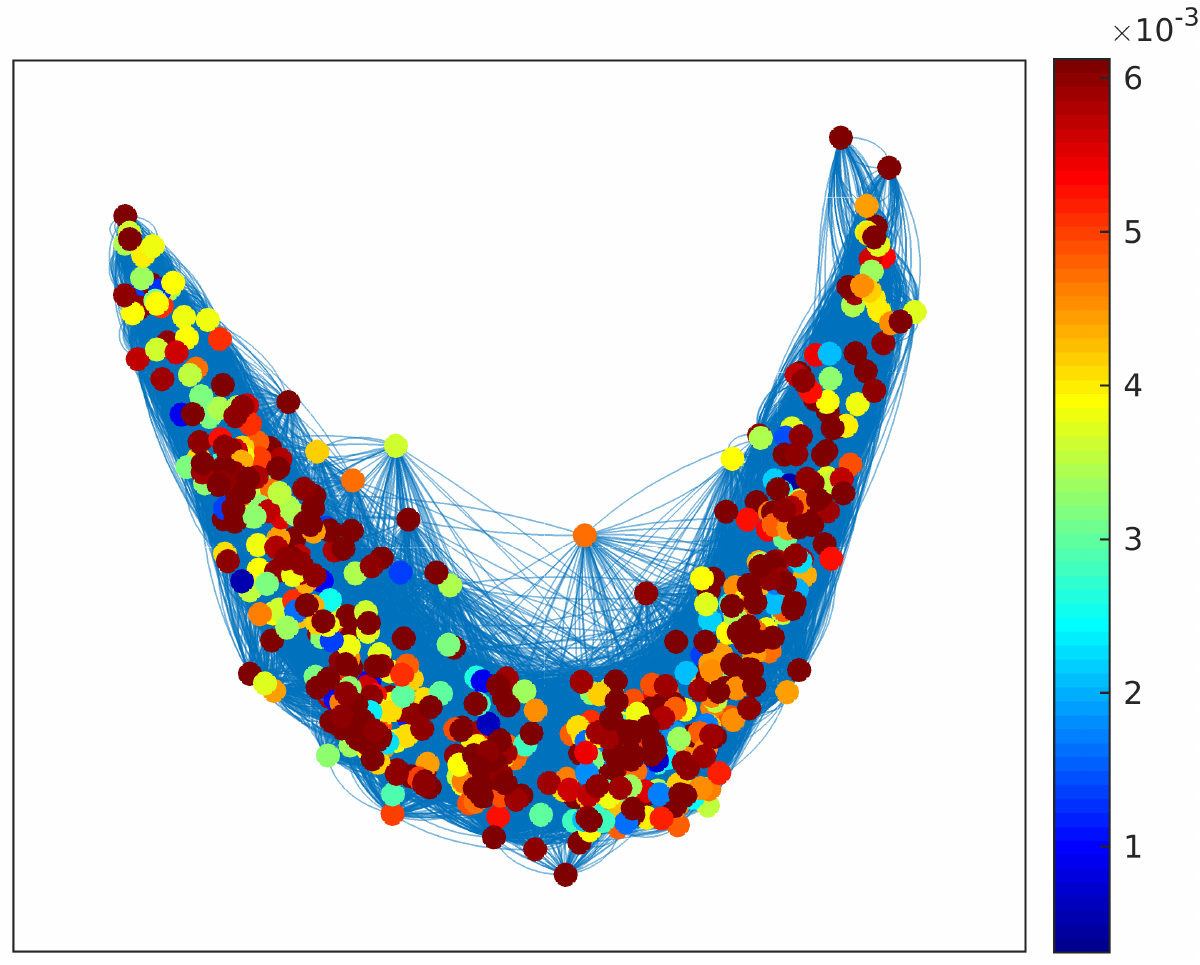} \\ \hline
        \theadb{CVPR'09}                                           & \tgraph{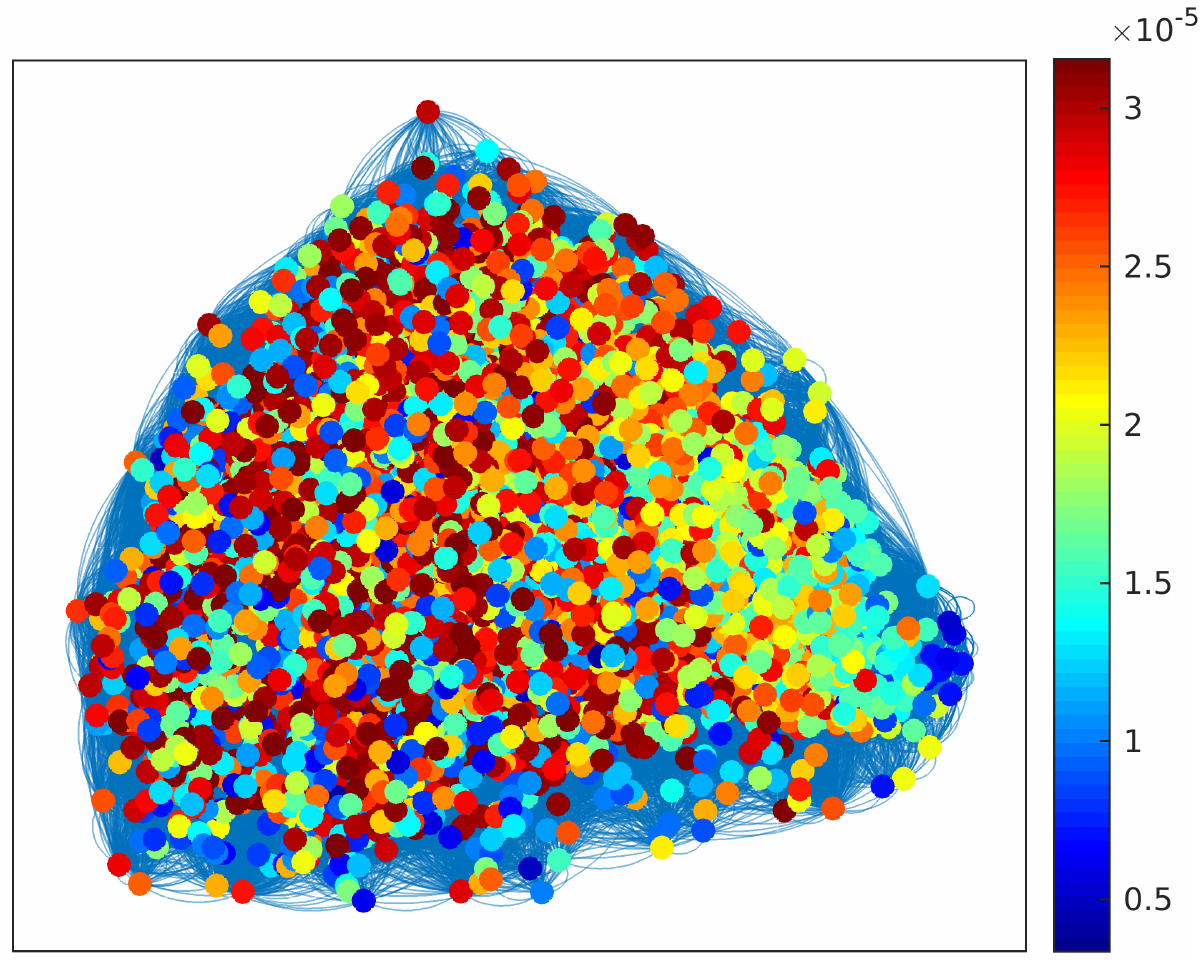}       & \tgraph{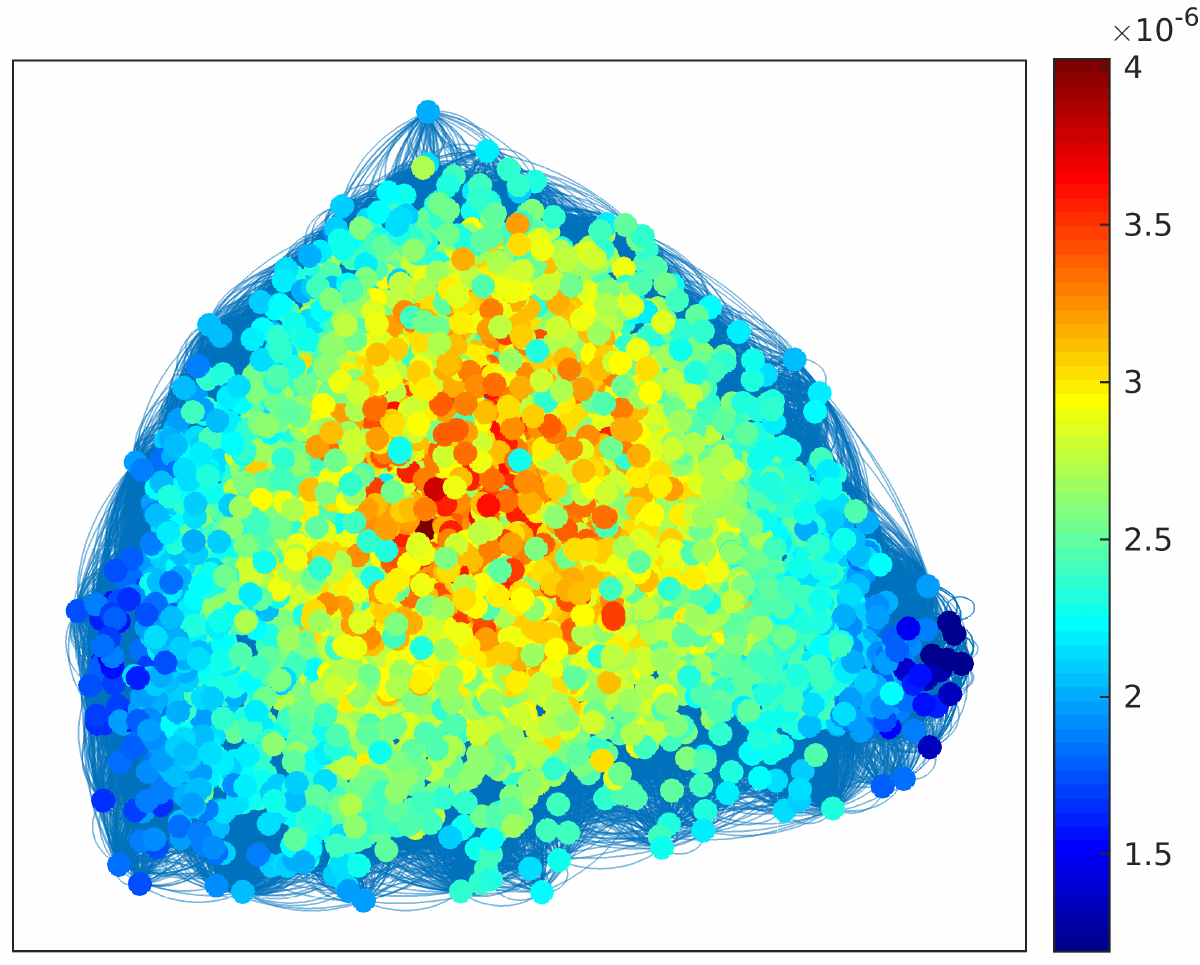}       & \tgraph{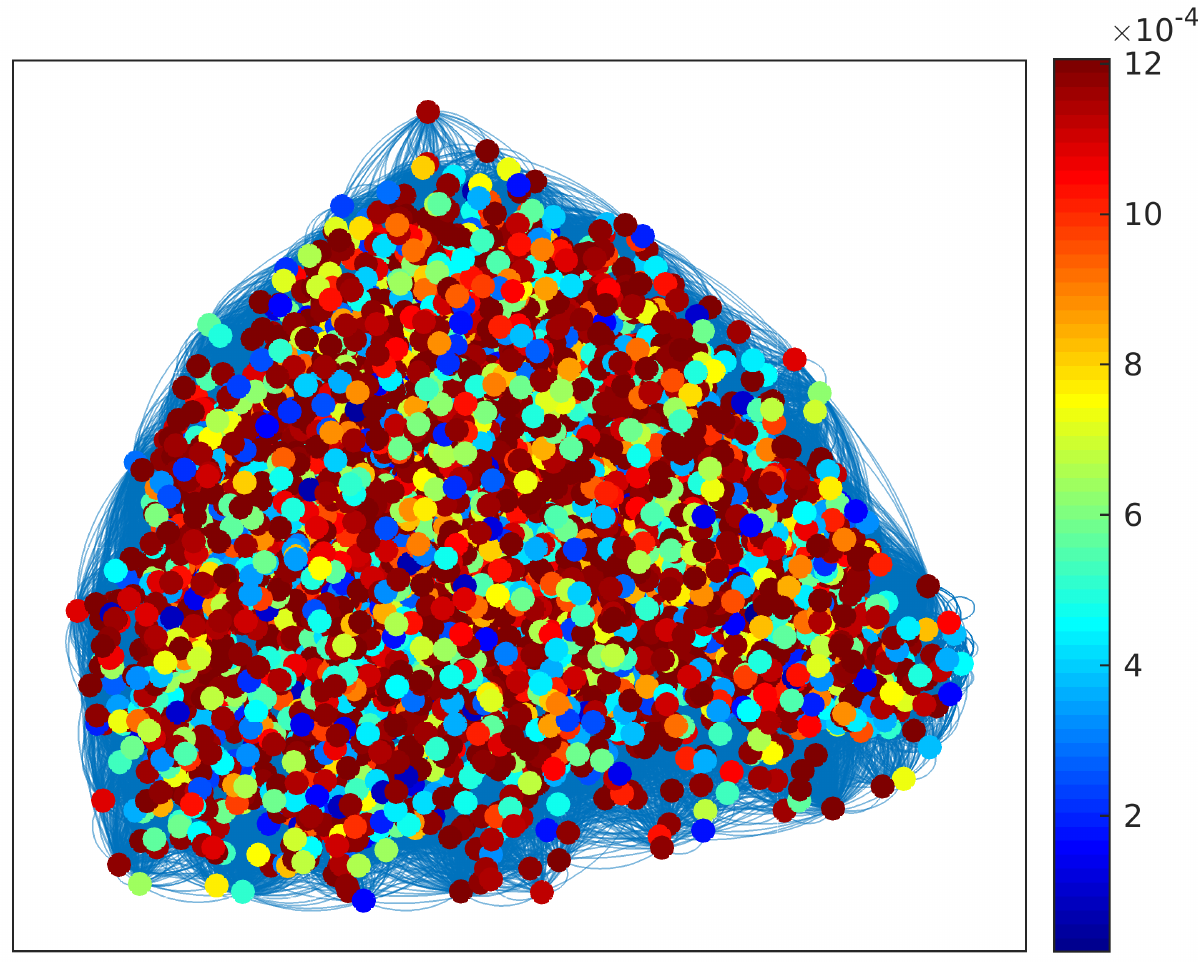}       \\ \hline
    \end{tabular}
    \end{singlespace}
\end{table*}
\begin{table}[!h]
    \caption{Distance comparison between affinity adjustment methods}
    \label{table:affinityadjustment}
    \centering
    \begin{singlespace}\scriptsize
    \begin{tabular}{clrrr}
        \hline\hline
        \textbf{Dataset} & \multicolumn{1}{c}{\textbf{Method}} & \multicolumn{1}{c}{\textbf{\begin{tabular}[c]{@{}c@{}}Max\\intra-cluster\end{tabular}}} & \multicolumn{1}{c}{\textbf{\begin{tabular}[c]{@{}c@{}}Min\\inter-cluster\end{tabular}}} & \multicolumn{1}{c}{\textbf{Mean}} \\\hline
        \multirow{3}{*}{\textbf{\begin{tabular}[c]{@{}c@{}}BCI\\ HaLT\end{tabular}}} & $ b_{ij} $ & $ 1.6005e+01 $ & $ 1.2262e-04 $ & $ 8.6587e-03 $ \\
        & $ c_{ij} $ & $ \mathbf{8.1589e+00} $ & $ \mathbf{2.1238e+00} $ & $ 3.5477e-02 $ \\
        & $ bd_{ij} $ & $ 3.0989e+02 $ & $ 8.9174e-06 $ & $ \mathbf{2.6830e-02} $ \\\hline
        \multirow{3}{*}{\textbf{\begin{tabular}[c]{@{}c@{}}HaSY\_v2\end{tabular}}} & $ b_{ij} $ & $ 9.4172e+00 $ & $ 3.9415e-05 $ & $ 1.6490e-03 $ \\
        & $ c_{ij} $ & $ 1.1414e+01 $ & $ 1.7681e+00 $ & $ 6.1324e-03 $ \\
        & $ bd_{ij} $ & $ \mathbf{4.5049e+00} $ & $ \mathbf{1.4349e+01} $ & $ \mathbf{3.0060e-03} $ \\\hline
        \multirow{3}{*}{\textbf{\begin{tabular}[c]{@{}c@{}}USPS\end{tabular}}} & $ b_{ij} $ & $ \mathbf{7.0751e-01} $ & $ \mathbf{2.2036e-01} $ & $ \mathbf{1.6860e-03} $ \\
        & $ c_{ij} $ & $ 9.4169e-01 $ & $ 8.0737e-02 $ & $ 7.7509e-04 $ \\
        & $ bd_{ij} $ & $ 1.6279e+01 $ & $ 5.5709e-06 $ & $ 2.3803e-03 $ \\\hline
        \multirow{3}{*}{\textbf{\begin{tabular}[c]{@{}c@{}}MNIST\end{tabular}}} & $ b_{ij} $ & $ \mathbf{3.6988e+00} $ & $ \mathbf{1.5879e+00} $ & $ 8.6150e-05 $ \\
        & $ c_{ij} $ & $ 6.0043e+00 $ & $ 4.0365e-01 $ & $ 4.9881e-04 $ \\
        & $ bd_{ij} $ & $ 2.4233e+01 $ & $ 3.8948e-06 $ & $ \mathbf{2.4709e-04} $ \\\hline
        \multirow{3}{*}{\textbf{\begin{tabular}[c]{@{}c@{}}UC Merced\end{tabular}}} & $ b_{ij} $ & $ 1.2515e+02 $ & $ 9.3198e-03 $ & $ 3.2635e-01 $ \\
        & $ c_{ij} $ & $ 1.7931e+02 $ & $ 4.4538e-14 $ & $ 7.8848e-01 $ \\
        & $ bd_{ij} $ & $ \mathbf{3.2659e+01} $ & $ \mathbf{5.0894e-02} $ & $ \mathbf{3.6199e-01} $ \\\hline
        \multirow{3}{*}{\textbf{\begin{tabular}[c]{@{}c@{}}CVPR'09\end{tabular}}} & $ b_{ij} $ & $ 3.3501e+01 $ & $ 1.2778e-15 $ & $ 2.8426e-02 $ \\
        & $ c_{ij} $ & $ \mathbf{2.7084e+01} $ & $ \mathbf{2.2428e-05} $ & $ \mathbf{6.3088e-03} $ \\
        & $ bd_{ij} $ & $ 7.2089e+01 $ & $ 4.7223e-07 $ & $ 4.4607e-03 $ \\\hline
    \end{tabular}
    \end{singlespace}
\end{table}
    The performance of the estimators can be ranked by their eigenvalues, as shown in Fig. \subref{fig_cvpr_ev}. A higher value resulted in better manifold regularization hence, increasing the classification accuracy. Table \ref{table:cvpr} shows the effect of varying $ |N| $ on the model's accuracy, using FPW, local Parzen window estimators, and VPW$ _{C} $. As evident, VPW$ _{C} $ increased the model's mean accuracy by $ \ge 2\% $. It also shows that as the $ |N| $ increases, the underlying model accuracy starts dipping due to cross-category connections.
\subsection{Random Walk based choice of Affinity Adjustment}
    Given the graph, the random walk starts from a vertex $ x_{i} $ and transitions to its neighbor with a probability that is proportional to the affinity between the data points. A random walk starting at a data point is more likely to stay within a group of points with similar labels than travel between dissimilar groups \cite{randomwalk}. This leads to creation of patches of similar densities as there may be regions in the graph with different degrees of unevenness. This propensity of random walk to discover groups can be used to select the best affinity adjustment method. Thus, we create graphs for the same data set using different affinity adjustments and perform random walk. The experiment was performed on all real-world data set classes: brain computer interface, handwritten digit recognition, and scene detection with non-local means, centroid, and Bhattacharyya distance based affinity adjustment methods.
    
    As shown in figures in Table \ref{table:fig_randomwalk}, it was observed that the random walk pattern over the graph constructed using the affinity adjustment methods can be utilized effectively to chose most appropriate adjustment out of three. As evident, for HaLT data, the $ c_{ij} $ gave consistent patches of data points having similar importance followed by $ b_{ij} $ and $ bd_{ij} $. A similar pattern is seen in CVPR'09 and hence, $ c_{ij} $ based affinity adjustment outperformed other two. The data spread and its connectivity in Hasy\_v2 and UC Merced for $ bd_{ij} $ smoothens the patches more effectively as compared to $ b_{ij} $ and $ c_{ij} $ and thus, the former affinity adjustment method performs better than the other two. The non-local means based affinity $ b_{ij} $ in case of both handwritten digit data set USPS and MNIST results in a better connectivity spread than the plain $ e_{ij} $ based affinity and hence, enforces a better function smoothening regularization as compared to $ c_{ij} $ and $ bd_{ij} $. Thus, in order to chose one affinity adjustment of the proposed three, opting for the smooth connectivity spread that can balance the unevenness encountered in plain $ e_{ij} $ metric should lead to an optimal affinity.
    
    \paragraph{\textbf{Affinity agnostic:}} An affinity adjustment that ensures that data with different labels are not likely to be close together would be the affinity choice. Such an affinity adjustment would decrease the inter-cluster similarity and increase the intra-cluster similarity. Minimum intra-cluster and maximum inter-cluster distance are desired for accurate affinity. The small intra-cluster distance ensures that data points belonging to similar class or exhibiting similar properties should remain spatially near and a large inter-cluster distance enforces the discriminative data points to be spatially separated hence, building an optimally connected graph. Additional affinity adjustments are needed when the neighborhood considered is not linear, and the Euclidean distance requires corrections for the non-linearity. Table \ref{table:affinityadjustment} lists the factors to be considered for choosing the best affinity adjustment method. The max intra-cluster column contains the maximum distance between all data points belonging to the same class. Among the three maximum values in each data set, the smallest value identifies the maximum threshold of intra-cluster distance, i.e., the distance between the same class data points will always be less than this value. Thus, keeping similar data points spatially close. Similarly, min inter-cluster distance column lists the minimum distance between data points belonging to different classes. The max value here decides the lower bound of the inter-cluster distance, i.e., no two data points belonging to different classes will have a distance less than this value. A large inter-cluster data points' distances keep them spatially far enough making them easily distinguishable. The last column mean contains the average distance between all the data points, and a mean value among the three leads to optimal results.
    
    A combination of the smallest maximum intra-cluster, largest minimum inter-cluster, and a mean distance would lead to best affinity and optimal point-wise convergence of then obtained graph Laplacian to its respective Laplace-Beltrami operator. As illustrated in the table \ref{table:affinityadjustment}, on BCI HaLT, the $ c_{ij} $ dominated on both intra-cluster and inter-cluster distances. Though it lagged behind $ bd_{ij} $ on mean distance, $ c_{ij} $ based affinity adjustment lead to accurate inferences in classification. Similarly, on HaSY\_v2, $ bd_{ij} $ gave best bounds than $ b_{ij} $ and $ c_{ij} $ on all three parameters hence, it was selected for affinity adjustment. Likewise, for other data set also, the best affinity adjustment method based on the values has been highlighted.
\begin{figure*}[!h]
    \centering
    \subfloat[Toroidal helix]{\includegraphics[width=0.15\linewidth,clip,keepaspectratio]{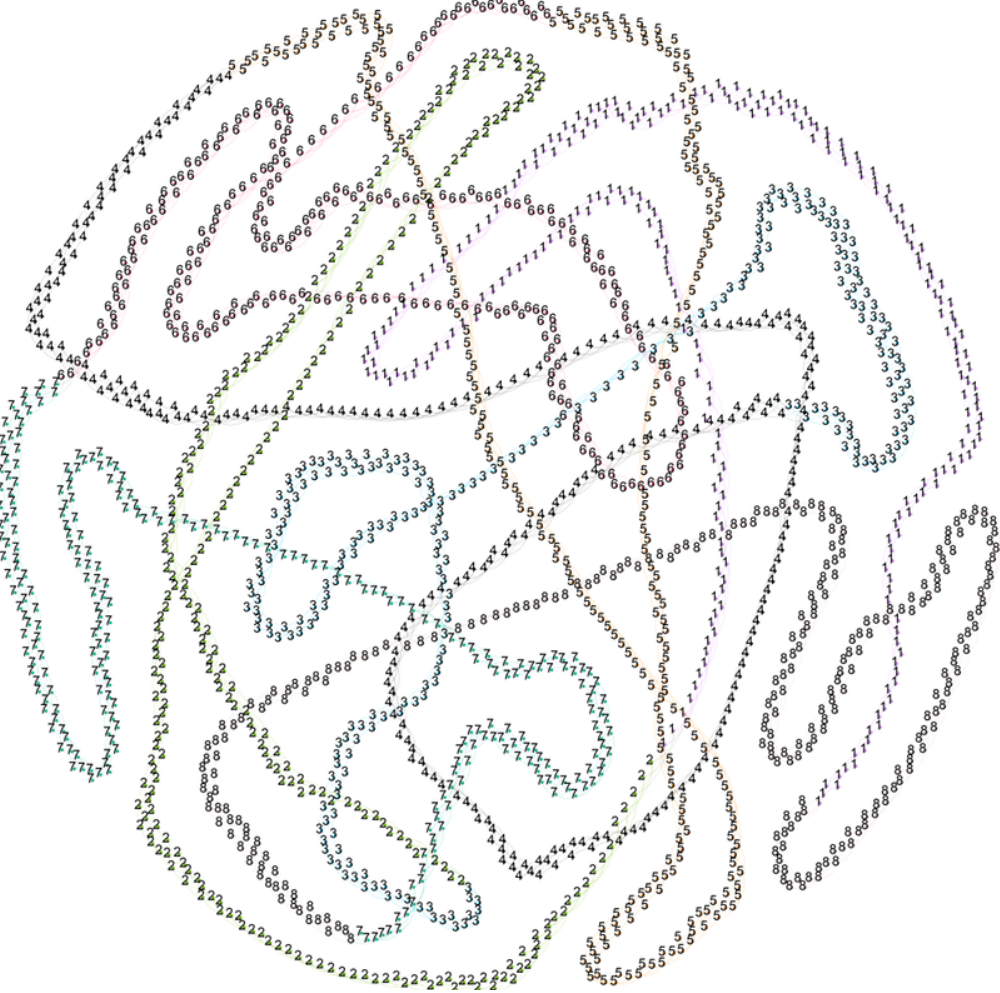}%
        \label{fig_toroidal_g}}
    \hfil
    \subfloat[BCI HaLT]{\includegraphics[width=0.15\linewidth,clip,keepaspectratio]{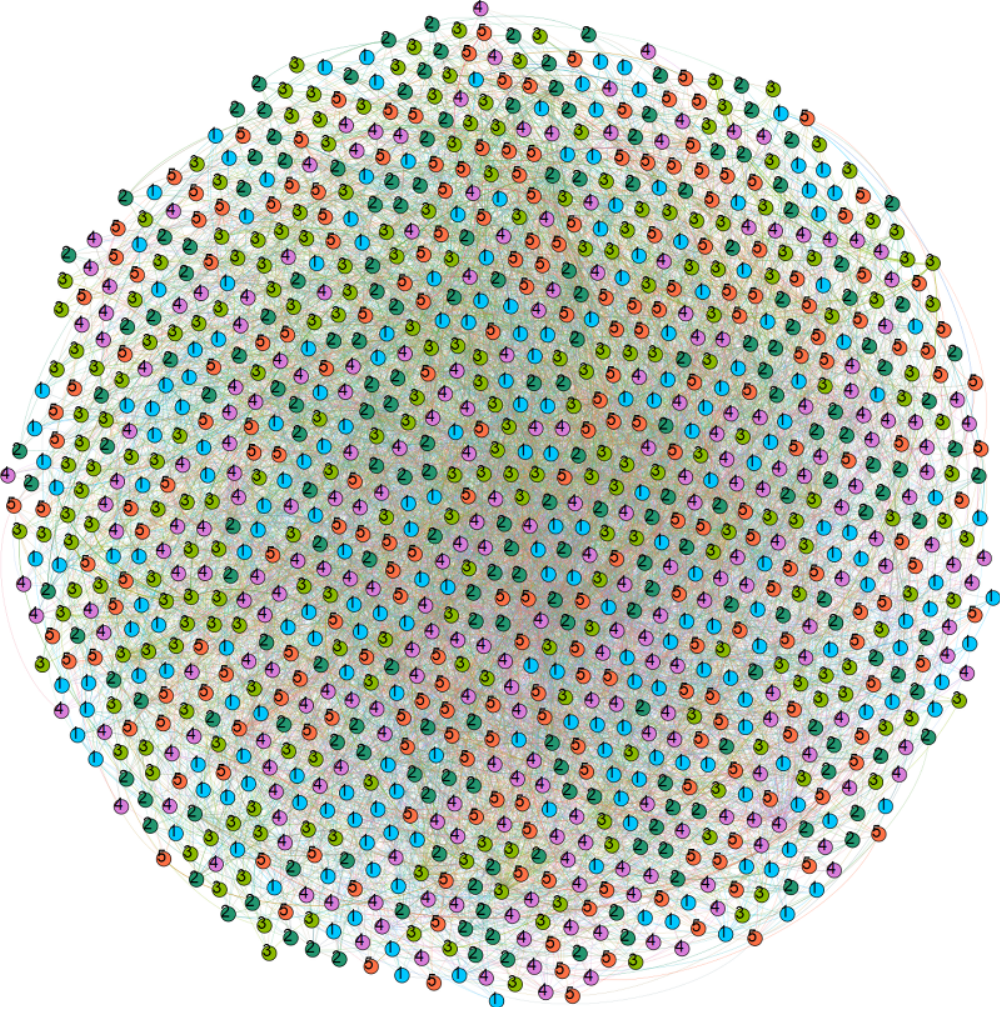}%
        \label{fig_bci_g}}
    \hfil
    \subfloat[Hasy\_v2]{\includegraphics[width=0.15\linewidth,clip,keepaspectratio]{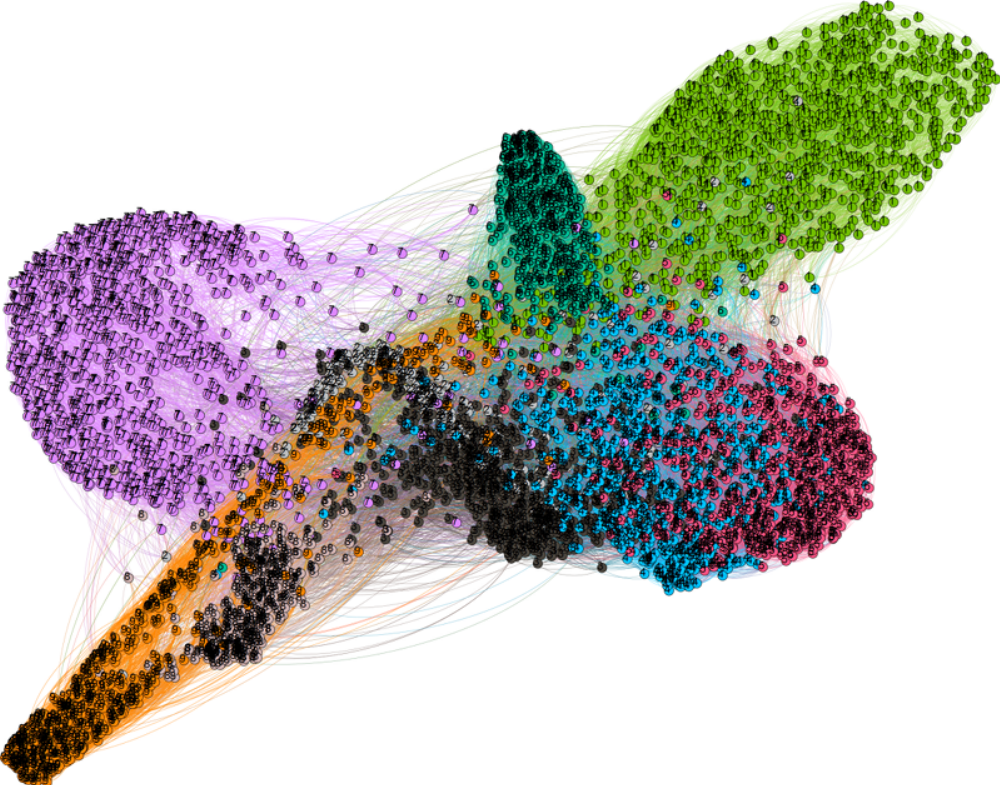}%
        \label{fig_hasy_g}}
    \hfil
    \subfloat[USPS]{\includegraphics[width=0.15\linewidth,clip,keepaspectratio]{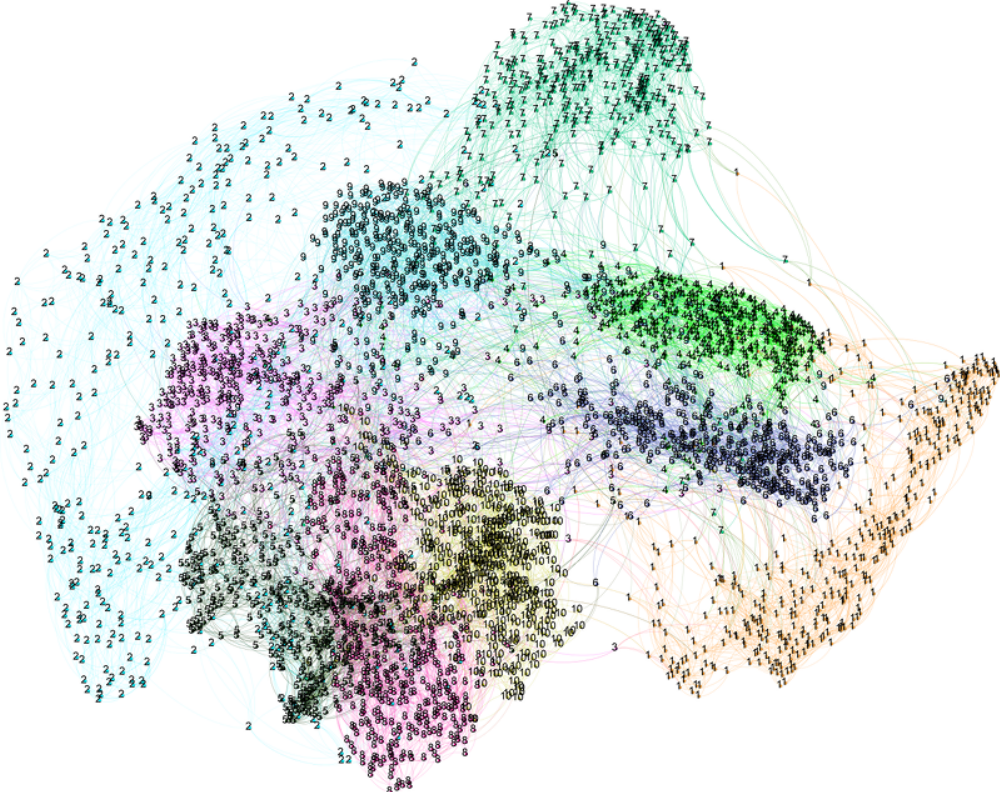}%
        \label{fig_usps_g}}
    \hfil\\
    \subfloat[MNIST]{\includegraphics[width=0.15\linewidth,clip,keepaspectratio]{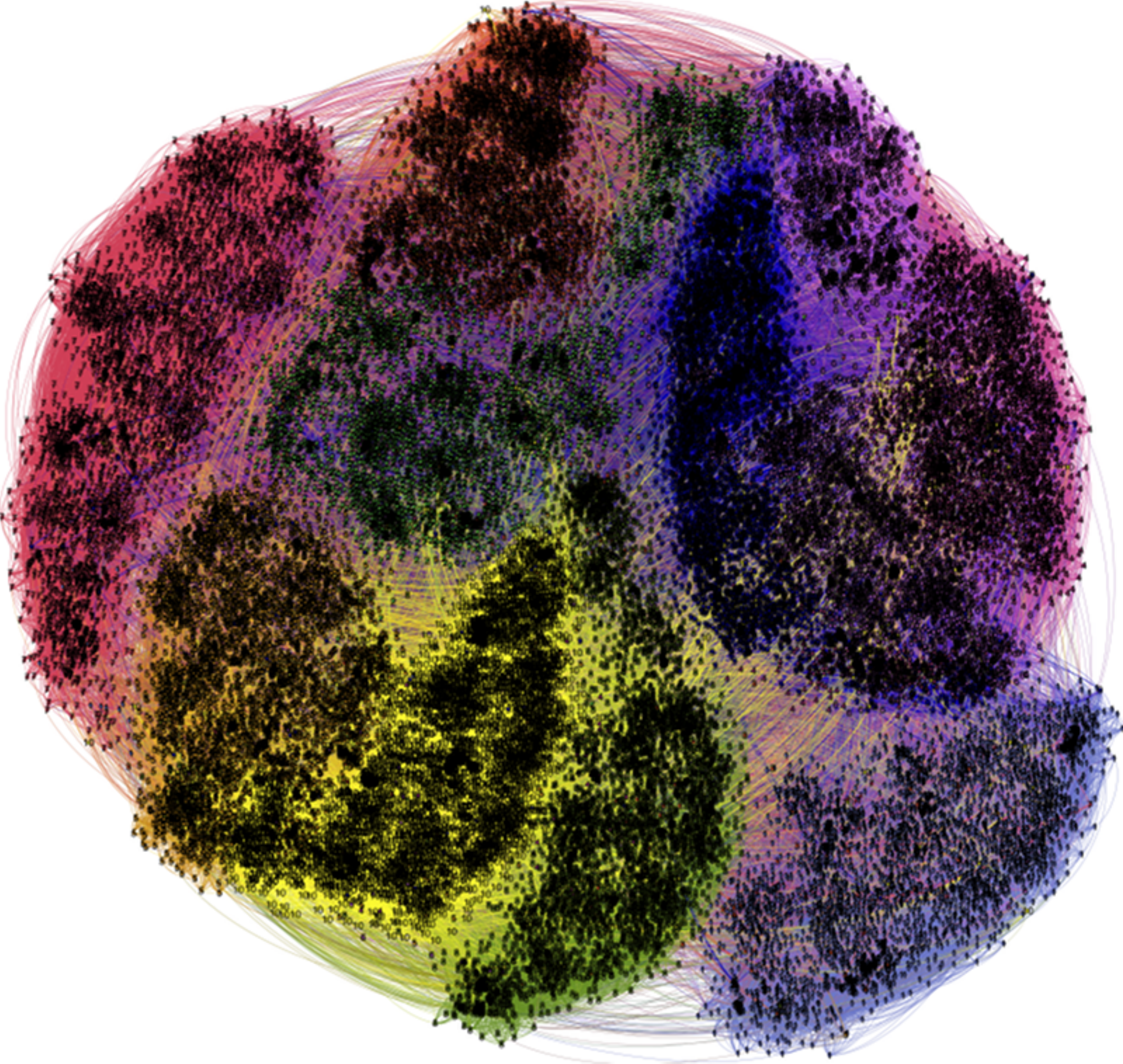}%
        \label{fig_mnist_g}}
    \hfil
    \subfloat[UC Merced]{\includegraphics[width=0.15\linewidth,clip,keepaspectratio]{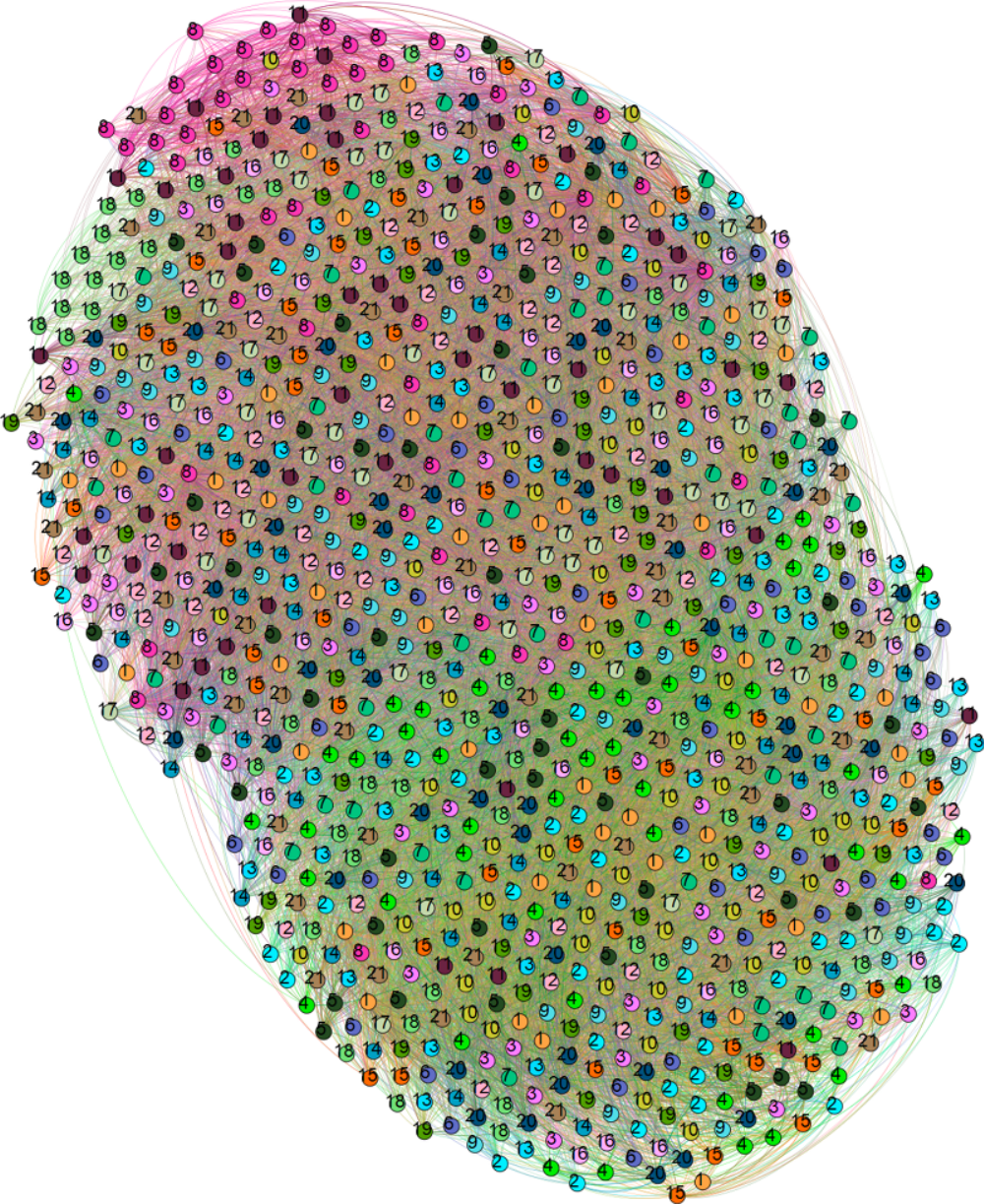}%
        \label{fig_ucmerced_g}}
    \hfil
    \subfloat[CVPR'09]{\includegraphics[width=0.15\linewidth,clip,keepaspectratio]{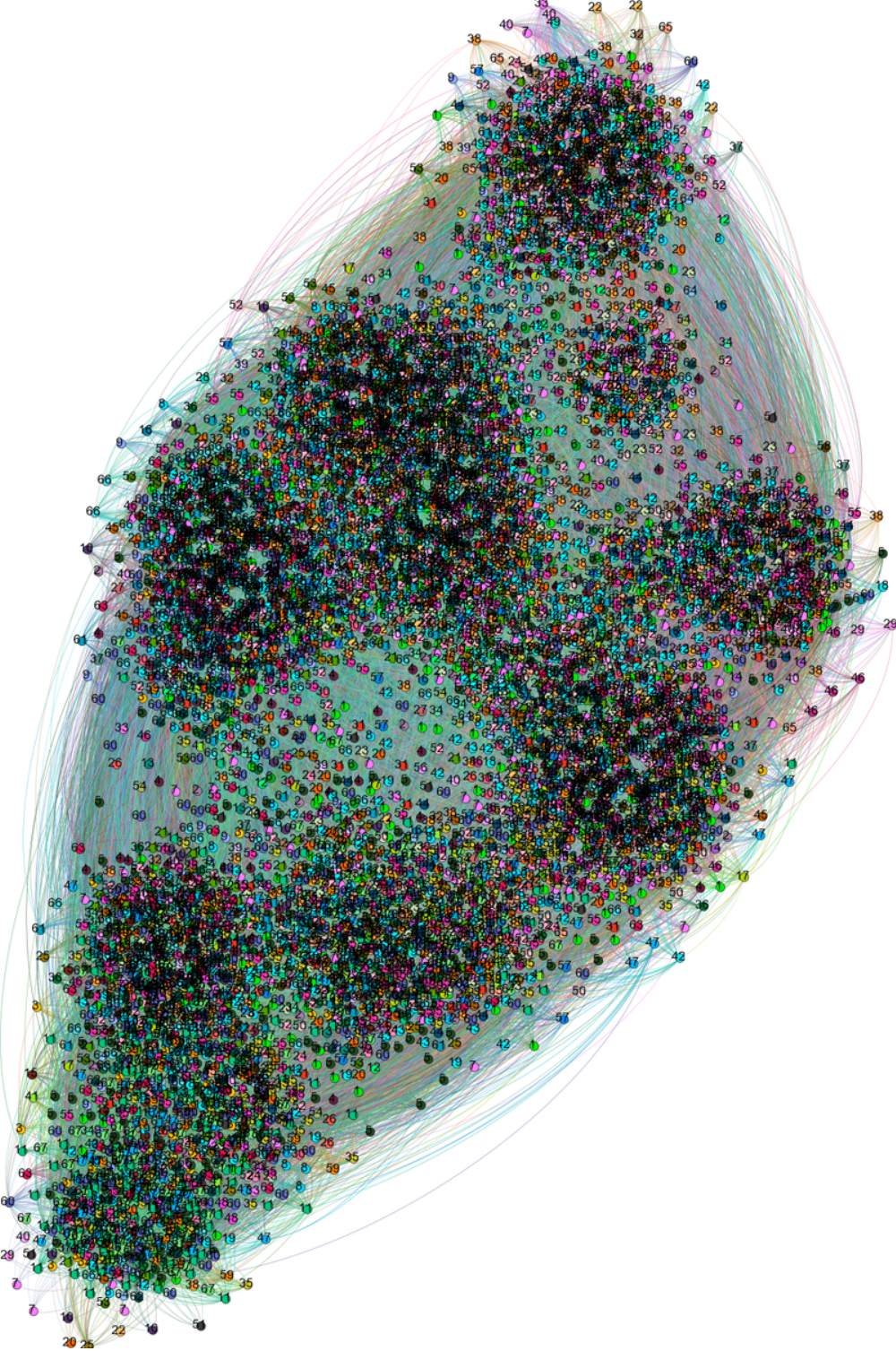}%
        \label{fig_cvpr_g}}
    \hfil
    \caption{Data set graphs}
    \label{fig:graphs}
\end{figure*}
\paragraph{\bfseries Discussion:} Fig. \ref{fig:graphs} shows the nearest neighbor graph created using only training data points of all data sets. In each graph, the data points are marked with their respective true cluster index, and for easy identification, each cluster has been marked with a different color. Based on the density distribution shown in the graphs, they can be broadly categorized into three categories: in the first category, the neighborhood graph of data points exhibit properties on uniformly sampled manifold with strong intra-cluster and minimal inter-cluster connections e.g. Toroidal helix (Fig. \subref{fig_toroidal_g}) and MNIST (Fig. \subref{fig_mnist_g}), in the second category, the graphs contain strong intra-cluster connections around neighborhood mean along with a large number of cross cluster edges e.g. BCI HaLT (Fig. \subref{fig_bci_g}) and CVPR'09 (Fig. \subref{fig_cvpr_g}), and third category graphs exhibit properties of data points with different distributions parameters e.g. Hasy\_v2 (Fig. \subref{fig_hasy_g}) and UC Merced (Fig. \subref{fig_ucmerced_g}). On first category data sets, VPW$ _{B} $ and FPW$ _{\sigma} $ gave approximately similar accuracy on data points for which graph Laplacian was calculated. However, former outperformed later in the test set by avoiding function over-fitting. On the second category, FPW$ _{\sigma} $ shared similar or larger eigenvalues as compared to VPW$ _{C} $, however, giving equal importance to both intra and inter-cluster edges led to under-performance of FPW$ _{\sigma} $ estimator. VPW$ _{C} $ Parzen window estimator corrected the affinity by weighing it with centroid distances which increased the classification model's accuracy. On third category graphs, it became difficult for existing Parzen window estimators to define appropriate value when samples inside the same data set exhibit properties of different distributions. VPW$ _{BD} $ overcame this problem by correcting the affinity based on their neighborhood's true distribution properties and hence, outperformed other estimators. The graph structure of USPS data set differed from all other graphs, as it contained scattered samples. VPW$ _{B} $ accurately adjusted the data spread and hence, gave a more accurately regularized classification model.
\section{Conclusion}
\label{sec:conclusion}
It is known that on an unevenly sampled Riemannian manifold, the globally fixed Parzen window leads to affinity drift towards high-density regions while a local Parzen window approximates the distribution in the neighborhood of a data point which makes it better than global Parzen window methods. It becomes inaccurate when sampling is uneven and neighborhoods are skewed. Variable Parzen window counters uneven sampling by utilizing known local properties to define the appropriate Parzen window between each pair of connected data points. The experimental results confirm that in comparison with existing Parzen window estimators, VPW considers the intrinsic geometrical information more accurately, thus, increasing the underlying model's accuracy. Due to uneven sampling, the respective neighborhoods to two connected data points exhibit different distribution properties individually, which further requires to be corrected through affinity adjustment methods. In general, all these techniques increase the model's accuracy, however, the technique selected based on the size of patches generated by random walk outperformed other techniques. In case of uneven or sparse sampling, the non-local means affinity adjustment gives large random walk patches. Similarly, when the clusters are concentrated around their respective means but also include large inter-cluster connections, affinity weighed using centroid affinity adjustment gives large patches and hence, increases the underlying model's accuracy. In a case when data points are distributed with different mean and variance, Bhattacharyya distance provides accurate affinity adjustment which is also confirmed by the random walk. We can conclude that affinity adjustment is a viable option for increasing classification accuracy on unevenly sampled manifold.
\bibliographystyle{elsarticle-num}
\bibliography{reference}   

\end{document}